\documentclass[10pt,twocolumn,letterpaper]{article}

\usepackage[pagenumbers]{cvpr} 

\usepackage{graphicx}
\usepackage{afterpage}
\usepackage{multicol} 
\usepackage{multirow} 
\usepackage{tabularx} 
\usepackage{subcaption}
\usepackage{stfloats}
\usepackage[accsupp]{axessibility}

\usepackage[dvipsnames]{xcolor}

\usepackage{multirow}
\usepackage{colortbl}

\usepackage{xcolor}
\definecolor{deepgreen}{rgb}{0.0, 0.5, 0.0}
\definecolor{deepblue}{rgb}{0.0, 0.0, 0.5}
\definecolor{deepred}{rgb}{0.5, 0.0, 0.0}
\definecolor{bronze}{rgb}{0.7, 0.2, 0.1}
\definecolor{darkorange}{rgb}{1.0, 0.55, 0.0}
\definecolor{deeppink}{rgb}{1.0, 0.08, 0.58}

\newcolumntype{g}{>{\columncolor{CuGray}}c}
\newcolumntype{z}{>{\columncolor{CuGray}}l}

\renewcommand{\paragraph}[1]{\noindent\textbf{#1.}\,\,}

\usepackage{xspace}

\def\onedot{.\@\xspace}
\def\eg{\emph{e.g}\onedot} 
\def\ie{\emph{i.e}\onedot}

\def\wrt{\emph{w.r.t}\onedot} 
\def\etal{\emph{et al}\onedot}

\newcommand{\Sref}[1]{Sec.~\ref{#1}}
\newcommand{\Eref}[1]{Eq.~(\ref{#1})}
\newcommand{\Fref}[1]{Fig.~\ref{#1}}
\newcommand{\Tref}[1]{Table~\ref{#1}}

\newcommand{\be}{\begin{eqnarray}}
\newcommand{\ee}{\end{eqnarray}}
\newcommand{\bee}{\begin{eqnarray*}}
\newcommand{\eee}{\end{eqnarray*}}

\newcommand{\matrixb}{\left[ \begin{array}}
\newcommand{\matrixe}{\end{array} \right]}

\newcommand{\bestred}[1]{\textcolor{red}{\textbf{#1}}}
\newcommand{\bestblue}[1]{\textcolor{blue}{\underline{#1}}}

\newcommand\blfootnote[1]{%
  \begingroup
  \renewcommand\thefootnote{}\footnote{#1}%
  \addtocounter{footnote}{-1}%
  \endgroup
}

\definecolor{cvprblue}{rgb}{0.21,0.49,0.74}
\usepackage[pagebackref,breaklinks,colorlinks,citecolor=cvprblue]{hyperref}

\def\paperID{14302} 
\def\confName{CVPR}
\def\confYear{2024}

\title{DITTO: Dual and Integrated Latent Topologies for Implicit 3D Reconstruction}

\author{
Jaehyeok Shim and Kyungdon Joo\thanks{Corresponding author.} \\
Artificial Intelligence Graduate School, UNIST \\
{\tt\small \{jh.shim,kyungdon\}@unist.ac.kr} \\
{\small \url{https://vision3d-lab.github.io/ditto}}
}

\newtoggle{issupp}
\togglefalse{issupp}

\begin{document}


\twocolumn[{
\vspace{-8mm}
\maketitle

\vspace{-10mm}
\begin{center}
    \begin{minipage}[b]{0.16\linewidth}
        \centering
        \includegraphics[width=\linewidth]{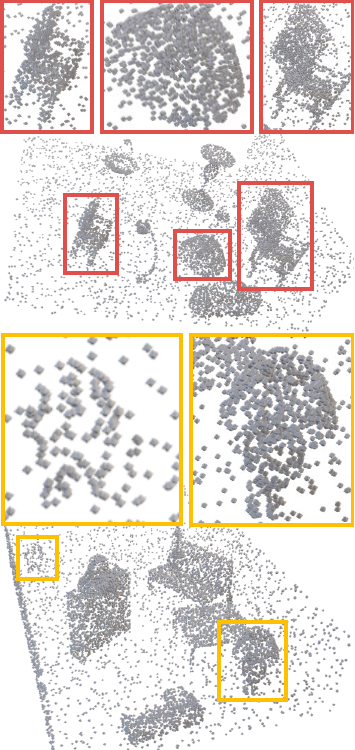}
        \footnotesize{(a) Input points (10K)}
        \label{fig:teasure-b}
    \end{minipage}
    \hspace{2mm}
    \begin{minipage}[b]{0.16\linewidth}
        \centering
        \includegraphics[width=\linewidth]{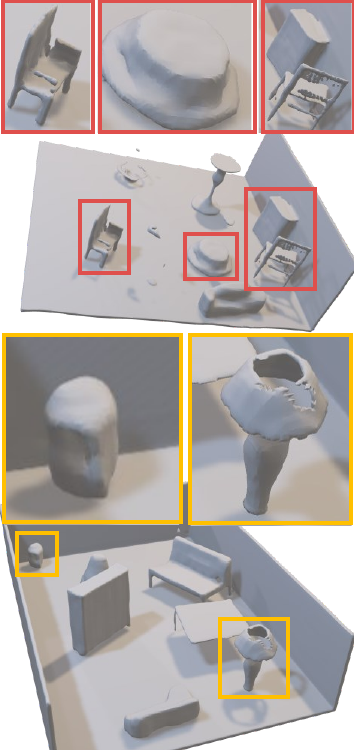}
        \footnotesize{(b) ConvONet~\cite{peng2020convolutional}}
        \label{fig:teasure-b}
    \end{minipage}
    \hspace{2mm}
    \begin{minipage}[b]{0.16\linewidth}
        \centering
        \includegraphics[width=\linewidth]{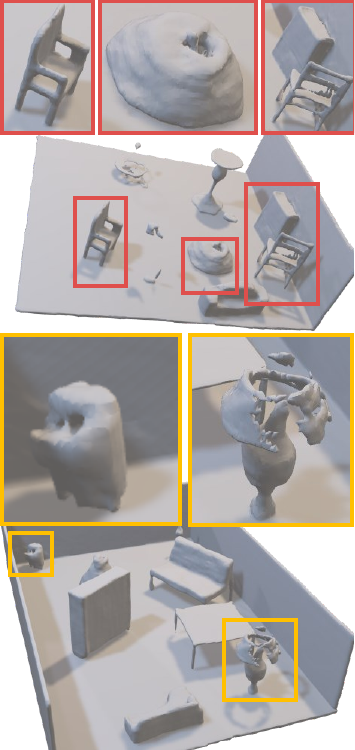}
        \footnotesize{(c) POCO~\cite{boulch2022poco}}
        \label{fig:teasure-c}
    \end{minipage}
    \hspace{2mm}
    \begin{minipage}[b]{0.16\linewidth}
        \centering
        \includegraphics[width=\linewidth]{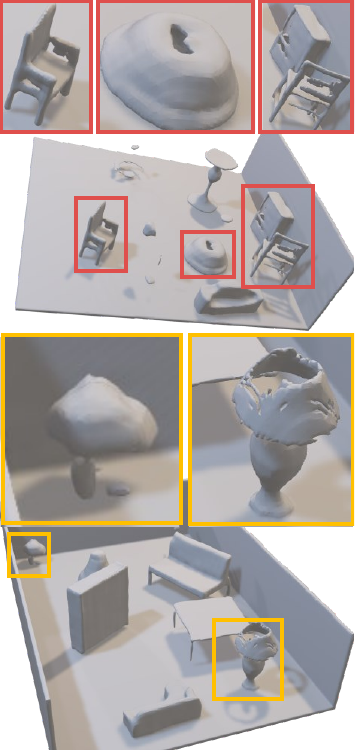}
        \footnotesize{(d) ALTO~\cite{wang2023alto}}
        \label{fig:teasure-d}
    \end{minipage}
    \hspace{2mm}
    \begin{minipage}[b]{0.16\linewidth}
        \centering
        \includegraphics[width=\linewidth]{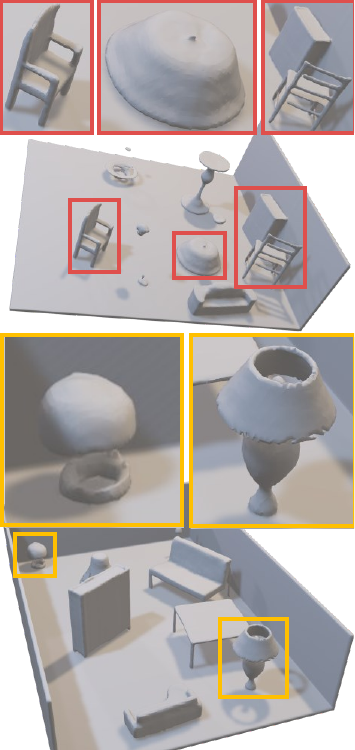}
        \footnotesize{(e) \texttt{DITTO} (ours)}
        \label{fig:teasure-e}
    \end{minipage}
    \vspace{-1mm}
    \captionof{figure}{
        \textbf{Scene-level 3D reconstruction comparison on the Synthetic Rooms dataset~\cite{peng2020convolutional}.}
        \texttt{DITTO} maximizes the benefits of both grid and point latents, thereby improving 3D surface reconstruction performance.
        We particularly focus on refining features based on point latents along with grid latents and integrating them (namely, dual and integrated latent topologies).
        This advancement enhances the ability to restore complex structures precisely, such as thin and intricate geometries, which posed challenges for previous methods.
    }
    \label{fig:teaser}
\end{center}
\vspace{-1mm}
}]

\begin{abstract}
We propose a novel concept of dual and integrated latent topologies (\texttt{DITTO} in short) for implicit 3D reconstruction from noisy and sparse point clouds.
Most existing methods predominantly focus on single latent type, such as point or grid latents.
In contrast, the proposed \texttt{DITTO} leverages both point and grid latents (\ie, dual latent) to enhance their strengths, the stability of grid latents and the detail-rich capability of point latents.
Concretely, \texttt{DITTO} consists of dual latent encoder and integrated implicit decoder.
In the dual latent encoder, a dual latent layer, which is the key module block composing the encoder, refines both latents in parallel, maintaining their distinct shapes and enabling recursive interaction.
Notably, a newly proposed dynamic sparse point transformer within the dual latent layer effectively refines point latents.
Then, the integrated implicit decoder systematically combines these refined latents, achieving high-fidelity 3D reconstruction and surpassing previous state-of-the-art methods on object- and scene-level datasets, especially in thin and detailed structures.
\blfootnote{*Corresponding author.}

\end{abstract}

\begin{figure*}[t]
\centering
\includegraphics[width=.92\linewidth]{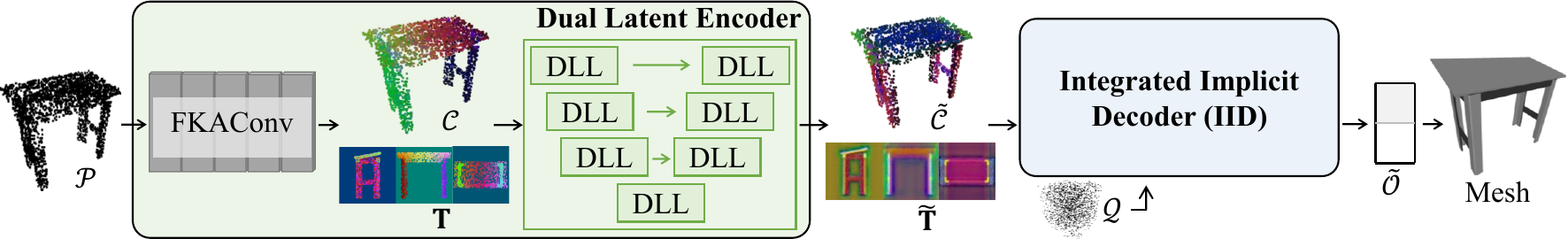}
\vspace{-2mm}
\caption{
    \textbf{Overview of \texttt{DITTO}}.
    \texttt{DITTO} architecture consists of the proposed dual latent encoder and integrated implicit decoder (IID) modules.
    In the encoder, \texttt{DITTO} receives a point cloud $\mathcal{P}$ and generates point and grid latents $\mathcal{C}$ and $\mathbf{T}$, respectively, using shallow FKAConv layers~\cite{boulch2020fkaconv}.
    These latents are refined in a U-shaped network composed with the proposed DLL to produce refined point and grid latents, respectively $\tilde{\mathcal{C}}$ and $\tilde{\mathbf{T}}$.
    Our IID estimates the occupancy of given arbitrary query locations.
    The mesh can be obtained by applying the Marching Cubes algorithm~\cite{lorensen1998marching} to the occupancies estimated as form of a regular grid.
}
\label{fig:overview}
\vspace{-2mm}
\end{figure*}

\section{Introduction}

Implicit 3D reconstruction aims to determine surface boundaries by estimating implicit values, such as occupancy and signed distance fields, based on given query coordinates~\cite{mescheder2019occupancy}.
%
In particular, implicit 3D reconstruction has evolved using geometric primitives like vectors~\cite{mescheder2019occupancy,chen2019learning,saito2019pifu,park2019deepsdf}, grids~\cite{peng2020convolutional,tang2021sa,lionar2021dynamic,peng2021shape}, and point clouds~\cite{boulch2022poco,wang2023aro} as intermediaries, namely, \emph{latent representations}.
Prior studies have focused on selecting appropriate latent representations for 3D reconstruction.
Specifically, early methods~\cite{mescheder2019occupancy,saito2019pifu,park2019deepsdf,chen2019learning} use vectors as their latent representation because of their simplicity.
However, they fall short in handling large-scale scenes due to the absence of a geometric prior (\ie, positional information).
To alleviate this issue, subsequent methods based on grid latent have emerged~\cite{chibane2020implicit,peng2020convolutional,lionar2021dynamic,tang2021sa}.
Grid latents have {similar shapes with occupancy cube}, the target domain of 3D reconstruction. 
Thus, they offer high-fidelity reconstructions at the scene-level but often lack detail because of resolution constraints.
On the other hand, point latent-based approaches~\cite{boulch2022poco,wang2023aro} enable detailed reconstruction because they preserve the details of the input points without information loss (\eg, quantization).
However, they can produce unstable results due to ambiguities, such as holes in thin structures, 
because they can be sensitive to the noise inherent in the input point coordinates. 
In short, each latent has its own pros and cons, so it is crucial to leverage the strengths of each representation properly.

As an attempt to combine the strengths of each latent, Wang~\etal~\cite{wang2023alto} introduces a new alternating latent topology concept, so-called ALTO.
Concretely, ALTO simultaneously utilizes two latent representations by alternatively projecting one latent into another. 
ALTO then decodes the combined features in the form of a grid for 3D reconstruction.
Such an intuitive and alternative approach improves 3D reconstruction performance and is meaningful as a first attempt. 
However, ALTO may overlook the advantages of abundant features extractable from point latents and makes the implicit decoder rely solely on a grid latent-based decoder for convenience.

In this work, we propose a novel concept of dual and integrated latent topologies (\texttt{DITTO}) for implicit 3D reconstruction.
The proposed \texttt{DITTO} aims to systematically integrate the strengths of each latent while maintaining their spatial structure of point and grid latents (\ie, dual latent).
Specifically, we seek to offset the inherent ambiguity of point latents through the stability of grid latents and, conversely, complement the resolution constraints of grid latents through the detailed representation by point latents.

The proposed DITTO employs an encoder-decoder architecture for dual latent (see \Fref{fig:overview}).
From a given point cloud, our encoder, called \emph{dual latent encoder}, constructs point and grid latents and refines this dual latent while preserving their original shapes.
In particular, we propose a new dynamic sparse point transformer (DSPT) for point features, which leverages large receptive fields, enabling effective learning of point-based spatial patterns.
Based on DSPT, we design a dual latent layer (DLL) that iteratively and separately updates dual latent with the correlation between two latents. 
This DLL module allows us to implicitly learn challenging patterns, such as thin objects, that cannot be handled by grid latent alone.
Then, our decoder, called \emph{integrated implicit decoder (IID)}, integrates enhanced dual latent to estimate the implicit value.
Unlike previous methods that utilize only a subset of latents, our decoder considers dual latent; especially, we unify grid-based and point-based implicit decoders together by introducing the concept of integrated latent.
IID helps to restore details by adjusting the relationship between neighbor points and query, adapting to the surface proximity.
Finally, the proposed \texttt{DITTO} improves 3D surface reconstruction performance, outperforming previous approaches and establishing a new state-of-the-art (see \Fref{fig:teaser}).

The main contributions of \texttt{DITTO} are as follows:
\begin{itemize}
    \item {
        \texttt{DITTO} is a new implicit 3D reconstruction method focusing on advanced feature extraction and fusion of grid and point latents, enhancing 3D understanding capabilities.
    }
    \item {
        We design a new dual latent layer module that refines dual latent while preserving their individual strengths. Particularly, we present a dynamic sparse point transformer (DSPT) to emphasize point feature refinement.
    }
    \item {
        We present a novel integrated implicit decoder that uniquely integrates two latents, providing clear surface boundaries, and robust to thin and intricate structures.
    } 
\end{itemize}

\section{Related Work}

3D reconstruction can be explicitly represented using a variety of geometric primitives, such as point, voxel, and mesh, or it can be inherently represented by leveraging such geometric primitives as latent representations~\cite{mescheder2019occupancy,park2019deepsdf,chen2019learning,peng2020convolutional,lionar2021dynamic,venkatesh2021deep,tang2021sa,boulch2022poco,wang2023alto,wang2023aro}. 
We refer to the former as explicit 3D reconstruction and the latter as implicit 3D reconstruction.
The readers refer to \cite{wang2023alto} for explicit representations.
In this section, we discuss the strengths and weaknesses of each latent for implicit 3D reconstructions.

\vspace{1mm} \noindent \textbf{Vector Latent Topologies.} 
Early approaches, such as \cite{mescheder2019occupancy,park2019deepsdf,chen2019learning}, employ an encoder-decoder architecture in a similar manner.
They encode a 3D shape into vector latents by the encoder and then reconstruct the 3D shape by the decoder.
The decoder estimates the implicit values of a given query point at arbitrary locations.
However, since vector latents lack geometric priors, they show decreased detailed reconstruction performance
(\ie, vector latent-based methods are vulnerable in large scenes with complex geometry).

\vspace{1mm} \noindent \textbf{Grid Latent Topologies.} \ 
As an alternative to vector latents, grid latents that encode geometric priors have been proposed~\cite{peng2020convolutional,venkatesh2021deep,tang2021sa,peng2021shape,wang2023alto}. 
Grid latent-based methods quantize the input point cloud into a grid latent during encoding. Then, they extract latent features at the query points using linear interpolation of the adjacent grids during decoding.

According to geometric primitives, we can divide grid latent-based methods into voxel-based and triplane-based methods.
Voxel latents can densely store latent features in the form of 3D grids~\cite{sun2022direct,yan2022shapeformer,mittal2022autosdf,fridovich2022plenoxels,shim2023diffusion}.
ConvONet~\cite{peng2020convolutional} expands the previous ONet~\cite{mescheder2019occupancy} by utilizing voxel latents instead of vector latents. 
ConvONet and ALTO~\cite{wang2023alto} take voxel latents as one of the base representations, which reveal their effectiveness, especially for scene-level reconstructions.
However, voxels require cubic computation and inherently have limited resolutions.
On the other hand, triplanes use 2D planes, which have less resolution constraints, allowing higher resolution than voxels.
This advantage leads to more effective restoration than voxels, particularly for object-level tasks~\cite{chen2022tensorf,chan2022efficient,gao2022get3d,chou2023diffusion,shue20233d,gupta20233dgen,wang2023rodin,dong2023ag3d}, which has slightly lower geometric complexity than scenes.
For this reason, ConvONet and ALTO use triplanes as the primary latent topologies in object-level reconstruction.

Even though grid latent-based approaches show effective reconstruction, but still require point feature quantization.
This process can result in a loss of fine details of the 3D surface, which is a fundamental limitation of grid latents.

\vspace{1mm} \noindent \textbf{Point Latent Topologies.} \ 
There are a few methods~\cite{boulch2022poco,zhang20223dilg,zhang20233dshape2vecset,wang2023aro} that encode features into point latents, typically offering benefits for the preservation of spatial information.
Namely, there is no need for quantization, preventing the loss of details.
POCO~\cite{boulch2022poco} encodes latent features into point latents and further enhances each feature by leveraging the point latents of neighbor points through point convolution~\cite{boulch2020fkaconv} and attention mechanism~\cite{vaswani2017attention}.
ARO-Net~\cite{wang2023aro} improves point features by introducing methods like anchor points and radial observations, enhancing the performance of the point-based implicit decoder.
However, these point latent-based methods may suffer instability from preserving spatial information of the point cloud, including even noise points. 
Concretely, point latents from noise points can affect neighbor points. 
In addition, the query feature extraction process may introduce ambiguity since different queries can share the same neighbor points.
We address these issues of point latents by leveraging the stability of grid latents.

\begin{figure}[t]
\centering
\begin{subfigure}[]{\linewidth}
    \centering
    \includegraphics[width=.95\linewidth]{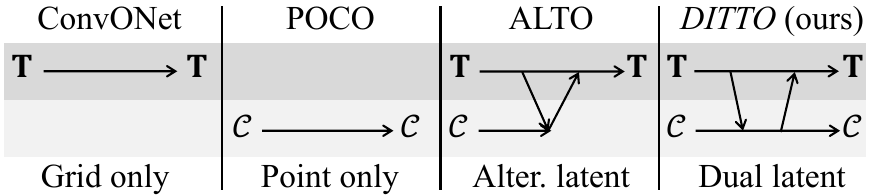}
    \caption{Encoder architectures}
    \label{fig:conceptual-a}
\end{subfigure}
\vspace{1mm}
\begin{subfigure}[b]{\linewidth}
    \centering
    \includegraphics[width=.95\linewidth]{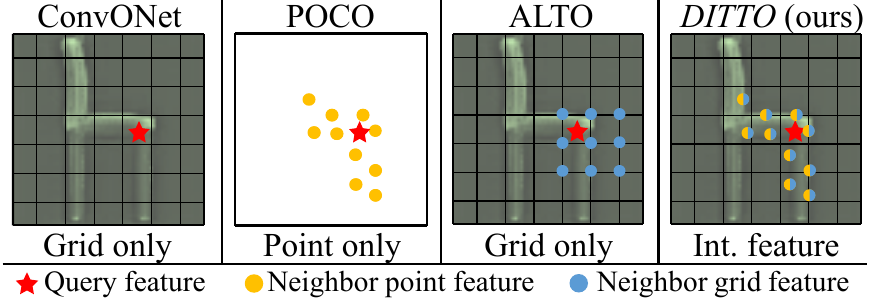}
    \caption{Decoder architectures}
    \vspace{-3mm}
    \label{fig:conceptual-b}
\end{subfigure}
\caption{
    \textbf{Conceptual comparison of {\small\texttt{DITTO}}.}
    We compare the concept of implicit 3D reconstruction methods in terms of latent representations: (a) encoders and (b) decoders.
    In (b), the image of green chairs represents grid features.
}
\label{fig:conceptual}
\end{figure}

\vspace{1mm} \noindent \textbf{Blended Latent Topologies.} \ 
ALTO~\cite{wang2023alto} introduces a new method that leverages multiple types of feature representations, utilizing both grid and point latents. 
This attempt is the first approach to combine the strengths of each latent.
Specifically, ALTO iteratively projects its features from grid to point cloud and vice versa.
By doing this, ALTO aims to preserve the details inherent in the point cloud and enable feature sharing between planes, leading to improved detailed surface reconstruction.
However, ALTO primarily relies on grid latents, with limited feature extraction from point latents, hindering its capacity to fully exploit the potential of point latents.
In particular, its decoder exclusively employs grid latents, making it directly susceptible to the resolution constraints inherent in grid latents.
To address these issues, we propose \texttt{DITTO} designed with advanced 3D geometry understanding capabilities.
\texttt{DITTO} comprises an enhanced point encoder, based on FKAConv~\cite{boulch2020fkaconv}, and advanced module for extracting features from both point and grid latents.
Furthermore, we present an implicit decoder that leverages the fusion of both grid and point latents for improved performance.

\section{Dual and Integrated Latent Topologies}

In this section, we propose a new topological concept, \emph{dual and integrated latent topologies} (\texttt{DITTO}), for implicit 3D reconstruction from a given noisy and sparse point cloud.
\texttt{DITTO} employs dual latent (\ie, point and grid latents) to leverage both the structural stability of grid latents and the preciseness of point latents.
\texttt{DITTO}, composed of encoder-decoder architecture for dual latent, refines and integrates these two latents, overcoming individual limitations and improving overall efficacy.
This strategy leads to high-fidelity surface reconstruction, even for thin, intricate structures.
%


\subsection{Overview} \label{subsec:overview}

Given a noisy and sparse point cloud $\mathcal P = \{\mathbf p_i \in \mathbb R^3 \}^N_{i=1}$ as input, 
the goal of \texttt{DITTO} is to accurately reconstruct 3D surfaces in a form of occupancy $\mathcal O = \{ o_j \in \{0, 1\} \}^M_{j=1}$ 
for query coordinates of arbitrary location $\mathcal Q = \{ \mathbf q_j \in \mathbb R^3 \}^M_{j=1}$, 
where $N$ and $M$ are the number of input points and queries. 

The proposed \texttt{DITTO} comprises two main parts: dual latent encoder and integrated implicit decoder (see \Fref{fig:overview}).
In the dual latent encoder, we first extract the point latents $\mathcal C = \{ \mathbf c_i \in \mathbb R^d \}^N_{i=1}$ for $\mathcal P$ based on FKAConv layers~\cite{boulch2020fkaconv}, where $d$ is the dimension of the point latents.
We then project $\mathcal C$ to grid latents.
Following the convention~\cite{peng2020convolutional,wang2023alto}, we use triplanes $\mathbf T \in \mathbb R^{3 \times R \times R \times d}$ or voxels $\mathbf V \in \mathbb R^{R \times R \times R \times d}$ as grid latents, where $R$ is the resolution of grids.
In this section, we explain the details of \texttt{DITTO} based on triplanes as grid latents, but they can seamlessly be replaced with voxels.
After extracting the initial dual latent ($\mathcal C$ and $\mathbf T$), we refine them using a UNet architecture~\cite{ronneberger2015u}, where each layer consists of the proposed dual latent layer~(DLL).
This UNet estimates refined grid $\tilde{\mathbf T}$ and point $\tilde{\mathcal C}$ latents.
The detailed description of DLL is in \Sref{sec:dll}.
Then, integrated implicit decoder~(IID) estimates occupancy probability $\tilde{\mathcal O} = \{ \tilde{o}_j \in \mathbb R \}_{j=1}^M$ of given query coordinates $\mathcal Q$ by integrating latents $\tilde{\mathbf T}$ and $\tilde{\mathcal C}$.
IID effectively manipulates the distinct characteristics of grid and point latents, facilitating the reconstruction of detailed and thin structures.
The detailed description of IID is in \Sref{sec:iid}.
More detailed network architectures are provided in supplementary materials.


\subsection{Dual Latent Layer} \label{sec:dll}

\begin{figure}[t]
\centering
\includegraphics[width=.85\linewidth]{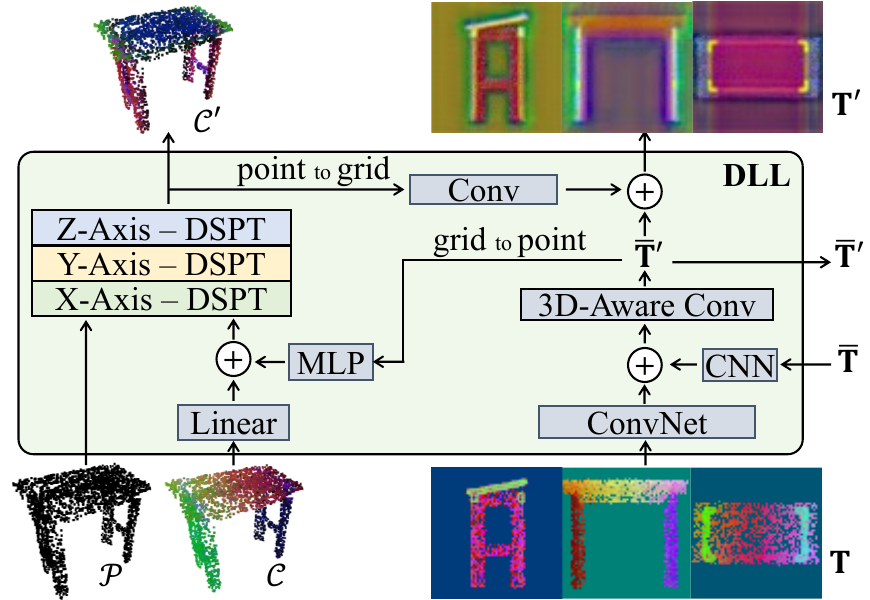}
\vspace{-1mm}
\caption{
    \textbf{Overview of our proposed DLL}.
    The input consists of $\mathcal C$ and $\mathbf T$, while the output comprises $\mathcal C'$ and $\mathbf T'$, representing the grid and point latents of the current layer, respectively.
    The $\bar{\mathbf T}$ and $\bar{\mathbf T}'$ represent the input and output grid latents, respectively, used for establishing a dense skip-connection between the layers.
}
\label{fig:dll}
\end{figure}

We present a dual latent layer~(DLL) that adeptly combines the advantages of both grid and point latents.
Specifically, DLL individually enhances two latents and infuses correlation between them to optimize their combined performance while preserving their own strengths, instead of combining them into a single latent (see \Fref{fig:conceptual-a}).
In particular, we newly present a dynamic sparse point transformer (DSPT) as a point feature extractor.
DSPT boosts point feature refinement by directly employing a transformer to point-based dynamic windows.

\begin{figure}[t]
\centering
\includegraphics[width=.85\linewidth]{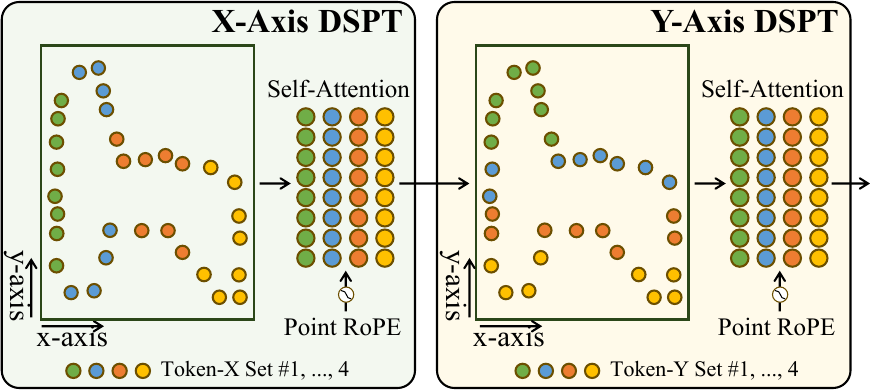}
\vspace{-2mm}
\caption{
    \textbf{Conceptual illustration of DSPT.}
    We visualize DSPT in the 2D domain for better understanding, but DSPT works in the 3D domain by sequentially processing the x-, y-, and z-axes.
}
\label{fig:dspt}
\end{figure}

\vspace{1mm} \noindent \textbf{Dynamic Sparse Point Transformer.} \ 
The concept of DSPT is to apply a transformer directly to points, stimulating local-global interactions among point latents.
Inspired by the concept of windowed attention~\cite{liu2021swin}, we create point-based non-overlapping windows and apply self-attention to each window. 
However, unlike images or voxels having regular grids, points have free form and are unordered. 
To alleviate this issue, we adopt a sorting-based windowing scheme, motivated by DSVT~\cite{wang2023dsvt}, FlatFormer~\cite{liu2023flatformer} and CSwinTransformer~\cite{dong2022cswin}.
Concretely, we sort points for a certain axis and divide them into multiple windows so that each window has an equal number of points. Subsequently, self-attention is applied within each window.
In a DSPT block, we repeat this procedure for each of the x-, y- and z-axes (see 2D illustration of DSPT in \Fref{fig:dspt}). 

Here, we describe the details of DSPT.
For given point features $\mathcal{C}$ and their coordinates $\mathcal{P}$, we first sort $\mathcal{C}$ based on $\mathcal{P}$ along a specific axis. 
We then divide the sorted point features into windows $\{\mathcal{C}^{\text{wnd}}_l\}_{l=1}^L$ so that the number of points belonging to each window is equal:
\vspace{-1mm}
\begin{equation}
\begin{aligned}
    \{\mathcal{C}^{\text{wnd}}_l\}_{l=1}^L = \textit{split}(\textit{sort}_\text{x}(\mathcal{C}, \mathcal{P}), L), \\
    \text{where\ \  }\mathcal{C}^{\text{wnd}}_l = \{\mathbf{c}_{i_\text{sorted}} \in \mathbb{R}^d\}_{i_\text{sorted}=1}^{{N/L}},
\end{aligned}
\vspace{-1mm}
\label{eq:dspt}
\end{equation}
$L$ is the number of windows, $\textit{sort}_\text{x}(\cdot,\cdot)$ denotes a sort function, which sorts $\mathcal C$ \wrt their coordinates $\mathcal P$ of x-axis, $\textit{split}(\cdot,\cdot)$ denotes a splitting function, which divides the given $\mathcal C$ into $L$ windows, and $i_\text{sorted}$ indicates the sorted index.
For convenience, we use the x-axis in \Eref{eq:dspt}, but we can also apply the y- and z-axes.
Before applying self-attention, a rotary positional embedding (RoPE)~\cite{su2021roformer}, modified one for point clouds~\cite{li2022lepard}, is applied to the query and key of the inputs of self-attention.
Subsequently, a shared MLP is applied to each $\mathbf{c}$, and the points are re-sorted in reverse to restore the original point order.
%

\vspace{1mm} \noindent 
\textbf{Overall Architecture of DLL for Dual Latent.} \ 
The proposed DLL takes dual latent, $\mathcal{C}$ and $\mathbf{T}$, from the previous DLL as input and systematically refines each latent feature while maintaining their shape and strengths (see \Fref{fig:dll}).  
Within DLL, we first enhance grid latents $\mathbf T$ using a simple CNN-based architecture. 
Note that when the triplane is used as grid latents, 3D-aware-conv~\cite{wang2023rodin} is additionally applied to induce feature exchange among three planes.
We call this enhanced grid latents as intermediate grid features $\bar{\mathbf{T}}'$, 
which is used to refine $\mathcal{C}$ and form a dense skip-connection between consecutive DLL modules, similar to \cite{wang2023alto}.

In the point latent perspective, we combine $\mathcal{C}$ with the point features projected from $\bar{\mathbf{T}}$, 
inducing information of stable grid latents to point latents.
We then refine these point latents using the proposed DSPT. We denote these refined point latents as $\mathcal{C}'$.
In the grid latent perspective, we project $\mathcal{C}'$ into grid domain and merge with intermediate latent $\bar{\mathbf{T}}'$, forming enhanced grid latents $\mathbf{T}'$, which creates a synergy of dual latent.


\subsection{Integrated Implicit Decoder} \label{sec:iid}

We present an integrated implicit decoder (IID) that estimates implicit values (\ie, occupancy) for a given query location using refined dual latent $\tilde{\mathbf T}$ and $\tilde{\mathcal C}$.
In particular, IID integrates the grid and point latents while reinforcing the strengths and compensating for shortcomings associated with each latent (see \Fref{fig:conceptual-b}).
To this end, we first analyze the pros and cons of each latent in terms of decoding perspective and then introduce IID in detail.

\vspace{1mm} \noindent \textbf{Point Latents \emph{vs.}~Grid Latents in Decoding.} 
Point-based implicit decoding approaches like POCO~\cite{boulch2022poco} utilize K-nearest neighbors (KNN) to define neighbor points for a given query point and use features of neighbor points to calculate query feature.
This point-based decoder is advantageous for detail restoration.
However, such point-based decoders can be fragile when handling thin structures due to inherent ambiguity because different query points may share the same neighbors, resulting in instability.

Grid-based decoding methods (\eg, ConvONet~\cite{peng2020convolutional} and ALTO~\cite{wang2023alto}) estimate occupancy by interpolating features from adjacent grids for a given query to determine the query feature. 
Additionally, ALTO employs an attention mechanism to alleviate the resolution constraints inherent to grid latents, thereby enhancing performance.
Specifically, ALTO compares query feature and adjacent grid features through subtraction-based cross-attention~\cite{zhao2021point}.
This comparison allows ALTO to use not only the query feature but also the varying patterns of adjacent grid features, mitigating the resolution constraints of grid latents.
However, despite this improvement, detailed reconstruction remains limited due to the inherent limitations of relying solely on grid latents.

\begin{figure}[t]
\centering
\includegraphics[width=.85\linewidth]{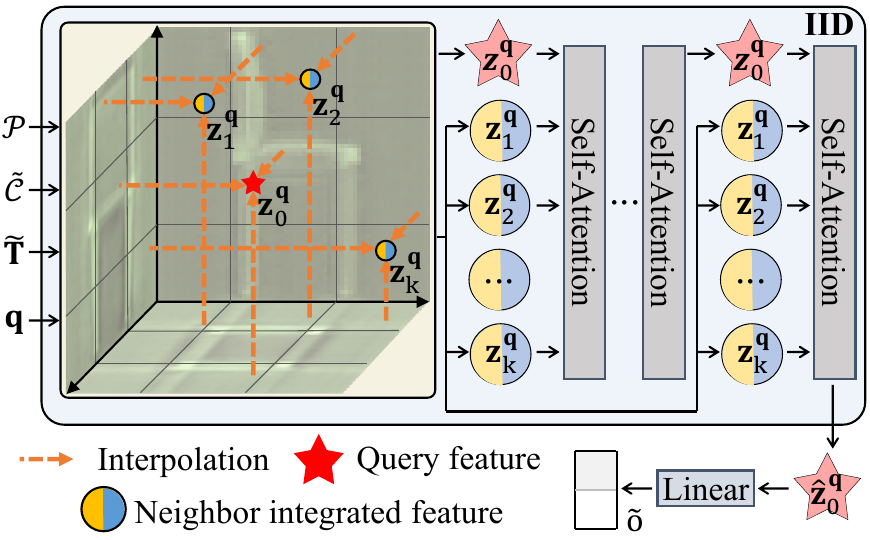}
\vspace{-2mm}
\caption{
    \textbf{Illustration of IID. 
    }
    IID makes query feature $\mathbf{z}^\mathbf{q}_0$ and neighbor integrated features $\{\mathbf z_1^\mathbf q, \dots, \mathbf z_k^\mathbf q\}$.
    Then, IID iteratively applies self-attention to a sequence of $\{\mathbf z_0^\mathbf q, \dots, \mathbf z_k^\mathbf q\}$ to update query feature $\mathbf z_0^\mathbf q$.
    Finally, we estimate the query feature $\mathbf{ \hat{z}}_0^\mathbf q$ and apply linear layer to estimate occupancy $\tilde{o}$.
}
\label{fig:iid}
\end{figure}

\vspace{1mm} \noindent \textbf{Integrated Decoder for Dual Latent.} \
The proposed IID selectively integrates the advantages of each latent decoding method (\ie, a hybrid approach between point and grid latent decoding).
We basically adopt KNN-based neighbors of point latents to determine neighbor point features since it can recover the detailed reconstruction free from the resolution limit.
Instead, we handle the inherent limitation of point features by using grid-based decoding.
That is, we combine point features at neighbor locations with adjacent grid features via interpolation.
This integration enables IID to effectively consider both latents.
We call these combined features as neighbor integrated features. 
Subsequently, we estimate implicit value through self-attention between neighbor integrated features and query feature obtained from grid latents.
Note that since neighbor points are located around the surface, the region of interest (\ie, size of the neighbor integrated features) is adaptively defined \wrt query-neighbor distances.
This adaptive mechanism facilitates the reconstruction of clear surface boundaries of detailed and intricate structures.

\begin{table*}[h]
\centering
\caption{
    \textbf{Object-level quantitative comparison on ShapeNet}.
    From left to right, the input point clouds become sparse (variations in density); they have 3K, 1K, and 300 points, respectively.
    The best scores are \bestred{red} in bold, and the secondary ones are \bestblue{blue} with underline.
}
\vspace{-2mm}
\resizebox{.9\linewidth}{!}{%
\begin{tabular}{c|cccc|cccc|cccc} \toprule \label{exp:shapenet}
\multirow{2}{*}{Method} & \multicolumn{4}{c|}{normal (3K points \& noise level 0.005)} & \multicolumn{4}{c|}{sparse (1K points \& noise level 0.005)} & \multicolumn{4}{c}{sparse (300 points \& noise level 0.005)} \\
 & IoU $\uparrow$ & Chamfer-$L_1$ $\downarrow$ & NC $\uparrow$ & F-score $\uparrow$ & IoU $\uparrow$ & Chamfer-$L_1$ $\downarrow$ & NC $\uparrow$ & F-score $\uparrow$ & IoU $\uparrow$ & Chamfer-$L_1$ $\downarrow$ & NC $\uparrow$ & F-score $\uparrow$ \\
\midrule[\heavyrulewidth]
ONet~\cite{mescheder2019occupancy} & 0.761 & 0.87 & 0.891 & 0.785 & 0.772 & 0.81 & 0.894 & 0.801 & 0.778 & 0.80 & 0.895 & 0.806 \\
ConvONet~\cite{peng2020convolutional} & 0.884 & 0.44 & 0.938 & 0.942 & 0.859 & 0.50 & 0.929 & 0.918 & 0.821 & 0.59 & 0.907 & 0.883 \\
POCO~\cite{boulch2022poco} & 0.926 & \bestblue{0.30} & 0.950 & \bestblue{0.984} & 0.884 & 0.40 & 0.928 & 0.950 & 0.808 & 0.61 & 0.892 & 0.869 \\
ALTO~\cite{wang2023alto} & \bestblue{0.930} & \bestblue{0.30} & \bestblue{0.952} & 0.980 &\bestblue{ 0.905} & \bestblue{0.35} & \bestblue{0.940} & \bestblue{0.964} & \bestblue{0.863} & \bestblue{0.47} & \bestblue{0.922} & \bestblue{0.924} \\
\hline
\texttt{DITTO} (ours) & \bestred{0.949} & \bestred{0.27} & \bestred{0.957} & \bestred{0.988} & \bestred{0.926} & \bestred{0.32} & \bestred{0.949} & \bestred{0.975} & \bestred{0.882} & \bestred{0.43} & \bestred{0.931} & \bestred{0.940} \\
\bottomrule
\end{tabular}
}
\end{table*}

\begin{figure*}[h]
\centering

\begin{subfigure}{0.12\linewidth}
    \includegraphics[width=\linewidth]{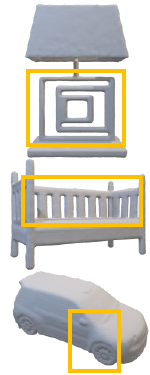}
    \caption{GT Mesh}
\end{subfigure}
\hspace{1mm}
\begin{subfigure}{0.12\linewidth}
    \includegraphics[width=\linewidth]{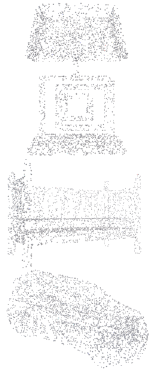}
    \caption{Input points}
\end{subfigure}
\hspace{1mm}
\begin{subfigure}{0.12\linewidth}
    \includegraphics[width=\linewidth]{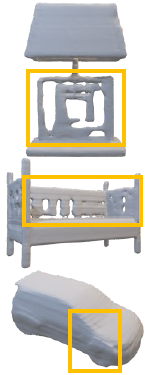}
    \caption{ConvONet~\cite{peng2020convolutional}}
\end{subfigure}
\hspace{1mm}
\begin{subfigure}{0.12\linewidth}
    \includegraphics[width=\linewidth]{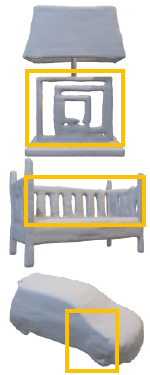}
    \caption{POCO~\cite{boulch2022poco}}
\end{subfigure}
\hspace{1mm}
\begin{subfigure}{0.12\linewidth}
    \includegraphics[width=\linewidth]{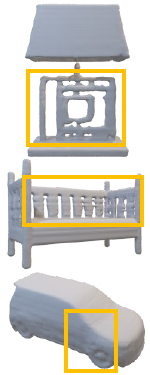}
    \caption{ALTO~\cite{wang2023alto}}
\end{subfigure}
\hspace{1mm}
\begin{subfigure}{0.12\linewidth}
    \includegraphics[width=\linewidth]{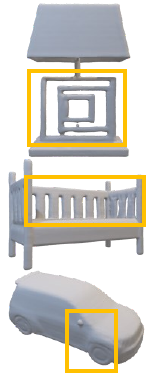}
    \caption{\texttt{DITTO} (ours)}
\end{subfigure}
\vspace{-2mm}
\caption{
    \textbf{Object-level 3D reconstruction comparison on ShapeNet~\cite{chang2015shapenet} with 3K input points}.
    \texttt{DITTO} distinctively excels in reconstructing thin structures, evidenced by the intricate details of lamps and benches. 
    Uniquely, \texttt{DITTO} is the only method that accurately captures the side mirror and the fine details of car wheels.  
}
\vspace{-2mm}
\label{fig:shapenet_qualitative}
\end{figure*}

The detailed IID is visualized in \Fref{fig:iid}. 
First, we estimate query feature $\mathbf{z}^\mathbf{q}_0$ by interpolating the grid features as in \cite{peng2020convolutional,wang2023alto} and applying linear layer so that the query feature has $2d$ dimensions:
\vspace{-1mm}
\begin{equation}
    \mathbf{z}^\mathbf{q}_0 = \textit{interpolate}(\tilde{\mathbf T}, \mathbf{q})\mathbf{W}, \quad \mathbf{z}^\mathbf{q}_0 \in \mathbb{R}^{2d},
    \vspace{-1mm}
\end{equation}
where $\textit{interpolate}(\cdot,\cdot)$ computes a grid feature of a given location using linear interpolation for adjacent grid latents, and $\mathbf{W} \in \mathbb{R}^{d \times 2d}$ is a weight matrix of linear layer.
Subsequently, we find neighbor points $\mathcal{P}^\mathbf{q}$ and neighbor point features $\tilde{\mathcal{C}}^\mathbf{q}$ of the query point $\mathbf q$ by using KNN:\vspace{-1mm}
\begin{equation}
    (\mathcal{P}^\mathbf{q}, \tilde{\mathcal{C}}^\mathbf{q}) = \textit{KNN}(\mathbf{q}, \mathcal{P}, \tilde{\mathcal{C}}, K),
    \vspace{-1mm}
\end{equation}
where $\textit{KNN}(\cdot,\cdot,\cdot,\cdot)$ returns given $K$ number of neighbor point coordinates $\mathcal{P}^\mathbf{q} = \{\mathbf{p}^\mathbf{q}_k \in \mathbb{R}^3\}_{k=1}^K$ and their features $\tilde{\mathcal{C}}^\mathbf{q} = \{\tilde{\mathbf{c}}^\mathbf{q}_k \in \mathbb{R}^d\}_{k=1}^K$. 
%
After that, we compute neighbor grid features $\mathbf{T}^\mathbf{q}$ by interpolating grid latents for every $\mathbf{p}^\mathbf{q}_k$:
\vspace{-1mm}
\begin{equation}
    \tilde{\mathbf{T}}^\mathbf{q} = \{ \textit{interpolate}(\tilde{\mathbf{T}}, \mathbf{p}^\mathbf{q}_k) \}^K_{k=1}.
    \vspace{-1mm}
\end{equation}
Then, we construct the neighbor integrated features $\mathcal{Z}^\mathbf{q}$ by concatenating $\tilde{\mathcal{C}}^\mathbf{q}$ and $\tilde{\mathbf{T}}^\mathbf{q}$ in channel direction:
\vspace{-1mm}
\begin{equation}
    \mathcal{Z}^\mathbf{q} = \textit{concat}(\tilde{\mathcal{C}}^\mathbf{q}, \tilde{\mathbf{T}}^\mathbf{q})
    = \{\mathbf z^\mathbf q_k \in \mathbb R^{2d}\}^K_{k=1},
    \vspace{-1mm}
\end{equation}
where $\textit{concat}(\cdot,\cdot)$ is a concatenation function.
Then, we refine $\mathbf{z}^\mathbf{q}_0$ by applying self-attention multiple times on a sequence $\{ \mathbf{z}^\mathbf{q}_0, \dots, \mathbf{z}^\mathbf{q}_K \}$ including both $\mathbf{z}^\mathbf{q}_0$ and $\mathcal{Z}^\mathbf{q}$.
We denote $\hat{\mathbf{z}}^\mathbf{q}_0$ as the refined query feature.
While applying self-attention, we use point-based RoPE~\cite{su2021roformer} as mentioned in \Sref{sec:dll}.
Note that during the self-attention, we update only $\mathbf{z}^\mathbf{q}_0$ and the other elements of the sequence $\{ \mathbf{z}^\mathbf{q}_1, \dots, \mathbf{z}^\mathbf{q}_K \}$ remain unchanged.
Finally, we estimate occupancy $\tilde{o}$ by applying linear layer to $\hat{\mathbf{z}}^\mathbf{q}_0$:
\vspace{-1mm}
\begin{equation}
    \tilde{o} = \hat{\mathbf{z}}^\mathbf{q}_0 \mathbf{W}_\text{out},
    \vspace{-1mm}
\end{equation}
where $\mathbf{W}_\text{out} \in \mathbb R^{2d}$ is a weight matrix of linear layer.

\subsection{Training Objectives}
%
We use binary cross entropy objective between $\tilde{\mathcal{O}}$ and $\mathcal{O}$:
\begin{equation}
    \vspace{-2mm}
    \mathcal{L}(\tilde{\mathcal{O}}, \mathcal{O}) {=} {-} \frac{1}{M} \sum^M_{j = 1} \left[
        o_j \log (\tilde{o}_j) {+} (1 {-} o_j) \log (1 {-} \tilde{o}_j)
    \right].
    \vspace{-3mm}
    \label{eq:bce_loss}
\end{equation}

\begin{figure*}[t]
\centering

\begin{subfigure}{0.16\textwidth}
    \includegraphics[width=\linewidth]{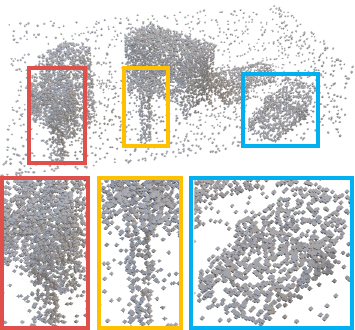}
    \caption{Input Points (10K)}
\end{subfigure}
\hspace{2mm}
\begin{subfigure}{0.16\textwidth}
    \includegraphics[width=\linewidth]{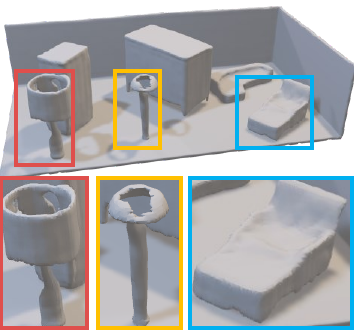}
    \caption{ConvONet~\cite{peng2020convolutional}}
\end{subfigure}
\hspace{2mm}
\begin{subfigure}{0.16\textwidth}
    \includegraphics[width=\linewidth]{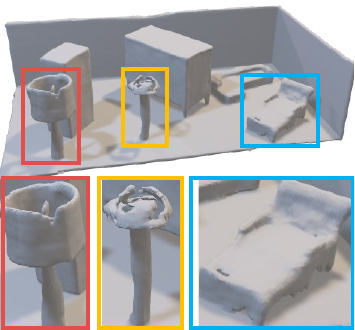}
    \caption{POCO~\cite{boulch2022poco}}
\end{subfigure}
\hspace{2mm}
\begin{subfigure}{0.16\textwidth}
    \includegraphics[width=\linewidth]{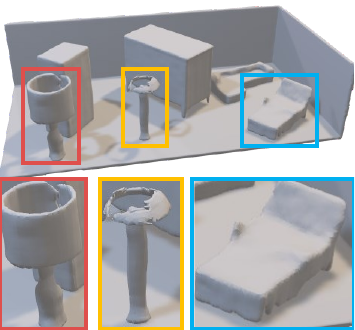}
    \caption{ALTO~\cite{wang2023alto}}
\end{subfigure}
\hspace{2mm}
\begin{subfigure}{0.16\textwidth}
    \includegraphics[width=\linewidth]{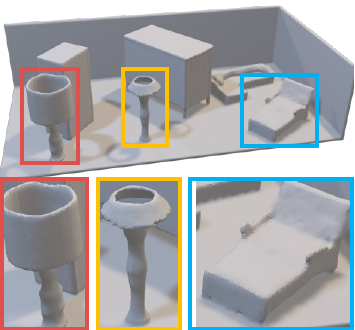}
    \caption{\texttt{DITTO} (ours)}
\end{subfigure}
\vspace{-2mm}
\caption{
    \textbf{Qualitative comparison of scene-level 3D surface reconstruction on the Synthetic Rooms dataset~\cite{peng2020convolutional}}.
}
\vspace{-2mm}
\label{fig:synthetic_rooms_qualitative}
\end{figure*}

\section{Experiments}

We evaluate \texttt{DITTO} against SoTA methods.
Details of implementation are provided in \Sref{sec:exp1}.
Qualitative and quantitative comparisons for object-level and scene-level are in \Sref{sec:exp2} and \Sref{sec:scene-level}, respectively.
In addition, we validate the generality in \Sref{sec:exp4}.
Additional experiment results are available in the supplementary materials.

\subsection{Baselines, Datasets, Metrics} \label{sec:exp1}

\vspace{1mm} \noindent \textbf{Implementation Details.}  
We implement \texttt{DITTO} in PyTorch~\cite{paszke2019pytorch}, utilizing xFormers~\cite{xFormers2022} and mixed-precision~\cite{micikevicius2017mixed}.
We train \texttt{DITTO} with Adam optimizer~\cite{kingma2014adam} and cosine annealing learning rate scheduler~\cite{loshchilov2016sgdr}.
For a fair comparison, we set the resolution of triplanes as $R{=}64$ and $R{=}128$ for object- and scene-level tasks, respectively.
For voxels, we set $R{=}64$ in scene-level tasks.
In DSPT, we use $L{=}25$ for $\{3\text{K}, 1\text{K}, 0.3\text{K}\}$ input points and $L{=}20$ for $10$K input points.
More detailed hyperparameters are described in the supplementary materials.

\vspace{1mm} \noindent \textbf{Comparison Methods.} \ 
To assess the 3D reconstruction performance of \texttt{DITTO}, we compare it with various baseline methods.
These methods include a non-learning-based method \cite{kazhdan2013screened}, as well as implicit methods that utilize diverse latent topologies, such as vector \cite{mescheder2019occupancy}, grid \cite{peng2020convolutional,lionar2021dynamic}, point \cite{boulch2022poco} and blended \cite{wang2023alto} latents.
Our evaluation procedure primarily follows the previous SoTA method, ALTO, including several additional experiments.

\vspace{1mm} \noindent \textbf{Datasets.} \ 
For the evaluation of object-level surface reconstruction, we use ShapeNet~\cite{chang2015shapenet}, which contains 13 categories of object watertight meshes.
For assessment of scene-level surface reconstruction, we use the Synthetic Rooms dataset~\cite{peng2020convolutional}, which has 5K synthetically created rooms utilizing objects from ShapeNet.
We follow the same train/val/test splits in both datasets with convention~\cite{peng2020convolutional,boulch2022poco,wang2023alto} for fair comparison.
The points are randomly sampled, and Gaussian noise is applied.
In addition, we adopt ScanNet-v2~\cite{dai2017scannet}, which contains 1,513 scene scans, for generality evaluation.


\begin{table}[t]
\centering
\caption{
    \textbf{Scene-level quantitative comparison on the Synthetic Rooms dataset~\cite{peng2020convolutional}}.
    We train each method for 10K input points with 0.005 noise level.
    The \textit{triplane comparison} involves assessing the impact of triplane in methods using grid latents.
}
\vspace{-2mm}
\resizebox{.9\linewidth}{!}{%
\begin{tabular}{c|cccc} \toprule \label{exp.synthetic_rooms}
Method & IoU $\uparrow$ & Chamfer-$L_1$ $\downarrow$ & NC $\uparrow$ & F-score $\uparrow$ \\
\midrule[\heavyrulewidth]
ONet~\cite{mescheder2019occupancy} & 0.475 & 2.03 & 0.783 & 0.541 \\
SPSR~\cite{kazhdan2013screened} & - & 2.23 & 0.866 & 0.810 \\
SPSR trimmed~\cite{kazhdan2013screened} & - & 0.69 & 0.890 & 0.892 \\
ConvONet~\cite{peng2020convolutional} & 0.849 & 0.42 & 0.915 & 0.964 \\
DP-ConvONet~\cite{lionar2021dynamic} & 0.800 & 0.42 & 0.912 & 0.960 \\
POCO~\cite{boulch2022poco} & 0.884 & 0.36 & 0.919 & 0.980 \\
ALTO~\cite{wang2023alto} & \bestblue{0.914} & \bestblue{0.35} & \bestblue{0.921} & \bestblue{0.981} \\
\midrule[\heavyrulewidth]
Ours & \bestred{0.928} & \bestred{0.34} & \bestred{0.930} & \bestred{0.984} \\
\hline \hline
\multicolumn{5}{c}{\textit{Triplane comparison}} \\
\midrule[\heavyrulewidth]
ConvONet~\cite{peng2020convolutional} & 0.805 & 0.44 & 0.903 & 0.948 \\
ALTO~\cite{wang2023alto} & \bestblue{0.895} & \bestblue{0.37} & \bestblue{0.910} & \bestblue{0.974} \\
\midrule[\heavyrulewidth]
Ours & \bestred{0.931} & \bestred{0.33} & \bestred{0.931} & \bestred{0.984} \\
\bottomrule
\end{tabular}
}
\end{table}

\vspace{1mm} \noindent \textbf{Evaluation Metrics.} \
We measure the reconstruction performance using standard quantitative metrics, such as IoU, Chamfer-$L_1$ distance, normal consistency (NC), and F-score~\cite{tatarchenko2019single}, following baseline methods.
For Chamfer-$L_1$ distance, we multiply 100 for convenience and use the threshold value as 1\% for F-score.

\subsection{Object-Level 3D Surface Reconstruction} \label{sec:exp2}

\vspace{1mm} \noindent \textbf{Quantitative Evaluation.} \
In \Tref{exp:shapenet}, we quantitatively compare the object-level surface reconstruction performance on ShapeNet~\cite{chang2015shapenet}.
\texttt{DITTO} exhibits superior performance across all metrics. 
Notably, \texttt{DITTO} demonstrates a substantial lead in the IoU metric; a four times larger gap compared to the previous SoTA~\cite{wang2023alto}.
In addition, evaluation on various input point densities implies that \texttt{DITTO} shows robust scores regardless of the number of points, even though we use point latents.
This result demonstrates the effectiveness and robustness of \texttt{DITTO}.

\vspace{1mm} \noindent \textbf{Qualitative Evaluation.} \ 
Figure~\ref{fig:shapenet_qualitative} shows the qualitative results.
\texttt{DITTO} shows clear surface boundaries, especially for thin and intricate structures.
Note that \texttt{DITTO} is the only method that successfully reconstructs the complex structures, such as the intricate pedestal of the lamp (first row), back of the bench (second row), rearview mirror and pattern on wheels of the car (third row).
In particular, reconstructing shapes with repeated thin structures is a challenging problem.
Grid latents struggle due to resolution constraints, while point latents often fail to create clear boundaries due to their inherent ambiguity.
In this challenging case, \texttt{DITTO} successfully creates a clear surface boundary.

\begin{table}[t]
\centering
\caption{
    \textbf{Scene-level quantitative comparison on the Synthetic Rooms dataset~\cite{peng2020convolutional} with sparse and noisy input data}.
    We train each method for sparse (3K points with 0.005 noise level) and noisy (10K points with 0.025 noise level) input point clouds.
}
\vspace{-2mm}
\label{exp:synthetic_rooms_noisy}
\resizebox{.8\linewidth}{!}{%
\begin{tabular}{c|cccc} \toprule \label{exp.synthetic_rooms-noisy}
Method & IoU $\uparrow$ & Chamfer-$L_1$ $\downarrow$ & NC $\uparrow$ & F-score $\uparrow$ \\
\midrule[\heavyrulewidth]

\multicolumn{5}{c}{\textit{Sparse input points (3K input points)}} \\
\midrule[\heavyrulewidth]

ConvONet~\cite{peng2020convolutional} & 0.818 & 0.46 & 0.906 & 0.943 \\
POCO~\cite{boulch2022poco} & 0.801 & 0.57 & 0.904 & 0.812 \\
ALTO~\cite{wang2023alto} & \bestblue{0.882} & \bestblue{0.39} & \bestblue{0.911} & \bestblue{0.969} \\
\hline
\texttt{DITTO} (ours) & \bestred{0.900} & \bestred{0.37} & \bestred{0.919} & \bestred{0.975} \\

\hline
\hline

\multicolumn{5}{c}{\textit{Noisy input points (0.025 noise level)}} \\
\midrule[\heavyrulewidth]

ConvONet~\cite{peng2020convolutional} & 0.777 & 0.57 & 0.872 & 0.885 \\
POCO~\cite{boulch2022poco} & 0.701 & 0.64 & 0.848 & 0.857 \\
ALTO~\cite{wang2023alto} & \bestblue{0.804} & \bestred{0.55} & \bestred{0.877} & \bestred{0.898} \\
\hline
\texttt{DITTO} (ours) & \bestred{0.811} & \bestred{0.55} & \bestblue{0.875} & \bestred{0.898} \\

\bottomrule
\end{tabular}
}
\end{table}

\subsection{Scene-Level 3D Surface Reconstruction} \label{sec:scene-level}

\vspace{1mm} \noindent \textbf{Quantitative Evaluation.}
We assess the scene-level reconstruction performance on the Synthetic Rooms dataset~\cite{peng2020convolutional}.
The quantitative results are in \Tref{exp.synthetic_rooms}.
\texttt{DITTO} surpasses previous methods in most of the metrics.
Regarding grid latent representation, \texttt{DITTO} marks a turning point.
Most grid-based methods~\cite{peng2020convolutional,wang2023alto} with triplane representations show decreased performance in complex scene-level reconstructions.
In contrast, \texttt{DITTO} maintains consistent performance, even with triplane representations.
In addition, quantitative results focusing on sparse and noisy inputs are in \Tref{exp:synthetic_rooms_noisy}.
\texttt{DITTO} shows outstanding performance robust to sparse and noisy input point clouds.
These results contrast the method solely based on point latents, which is vulnerable to contamination of the input point clouds.
%

\begin{figure*}[t]
\centering

\begin{subfigure}{0.16\textwidth}
    \includegraphics[width=\linewidth]{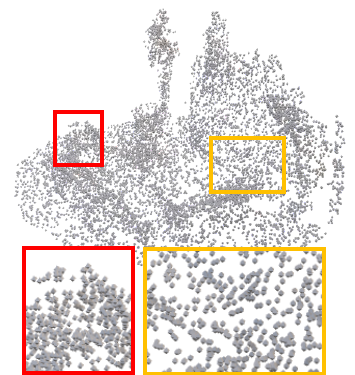}
    \caption{Input Points (10K)}
\end{subfigure}
\hspace{1mm}
\begin{subfigure}{0.16\textwidth}
    \includegraphics[width=\linewidth]{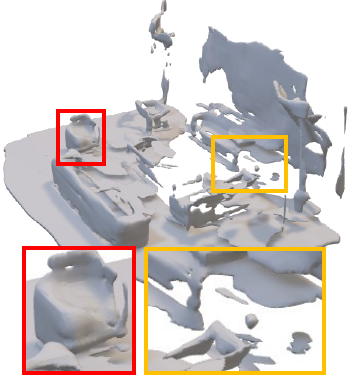}
    \caption{ConvONet~\cite{peng2020convolutional}}
\end{subfigure}
\hspace{1mm}
\begin{subfigure}{0.16\textwidth}
    \includegraphics[width=\linewidth]{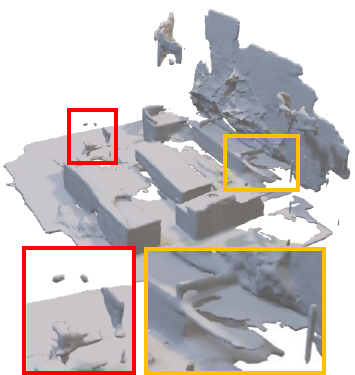}
    \caption{POCO~\cite{boulch2022poco}}
\end{subfigure}
\hspace{1mm}
\begin{subfigure}{0.16\textwidth}
    \includegraphics[width=\linewidth]{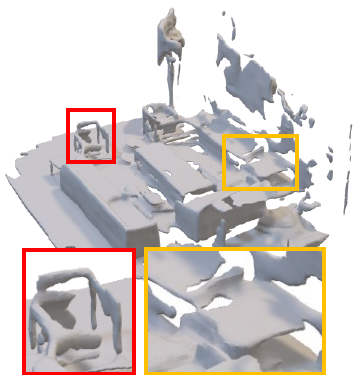}
    \caption{ALTO~\cite{wang2023alto}}
\end{subfigure}
\hspace{1mm}
\begin{subfigure}{0.16\textwidth}
    \includegraphics[width=\linewidth]{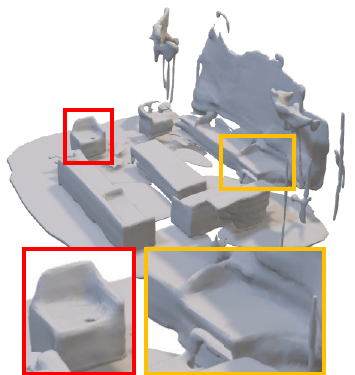}
    \caption{\texttt{DITTO} (ours)}
\end{subfigure}
\vspace{-2mm}
\caption{
    \textbf{Qualitative comparison of ScanNet-v2~\cite{peng2020convolutional}}.
}
\vspace{-4mm}
\label{fig:scannet}
\end{figure*}

\vspace{1mm} \noindent \textbf{Qualitative Evaluation.} \ 
We visualize the qualitative comparisons in \Fref{fig:teaser} and \Fref{fig:synthetic_rooms_qualitative}.
The results of ConvONet~\cite{peng2020convolutional}, a grid-based method, are relatively stable, with fewer holes in thin structure, but they lack detail (see lamps in red boxes).
On the other hand, POCO~\cite{boulch2022poco}, a point-based method, shows better detail but is less stable, often resulting in holes and artifacts (see lamps and chairs in yellow boxes).
ALTO~\cite{wang2023alto}, employing both latents, tends to offer better detail compared to ConvONet and better stability than POCO.
However, ALTO displays less stability than ConvONet and less detail than POCO.
Unlike the previous approaches, \texttt{DITTO} validates both superior stability and detail.
This result demonstrates that \texttt{DITTO} achieves synergy from the integration of two latents.

\subsection{Ablation Study}  \label{sec:exp4}

\begin{table}[t]
\centering
\caption{
    \textbf{Ablation study on the proposed modules.}
    We train networks on the Synthetic Rooms dataset by switching modules one by one from the baseline to the suggested method.
}
\vspace{-2mm}
\resizebox{\linewidth}{!}{%
\begin{tabular}{l|cccc} \toprule \label{exp.ablation.modules}
Method & IoU $\uparrow$ & Chamfer-$L_1$ $\downarrow$ & NC $\uparrow$ & F-score $\uparrow$ \\
\midrule[\heavyrulewidth]
ALTO~\cite{wang2023alto} (triplane; baseline) & 0.895 & 0.35 & 0.921 & 0.981 \\
\midrule[\heavyrulewidth]
+ DLL (DSPT backbone) & 0.921 & 0.35 & 0.925 & 0.981 \\
+ DLL (FKAConv~\cite{boulch2020fkaconv} backbone) & 0.917 & 0.35 & 0.922 & 0.979 \\
+ DLL (PointTransformer~\cite{zhao2021point} backbone) & 0.911 & 0.36 & 0.917 & 0.976 \\
+ IID & 0.907 & 0.35 & 0.913 & 0.976 \\
+ PointEncoder (FKAConv~\cite{boulch2020fkaconv}) & 0.912 & 0.36 & 0.918 & 0.978 \\
+ DLL + IID & 0.925 & 0.34 & 0.927 & 0.983 \\
+ DLL + PointEncoder & 0.926 & 0.34 & 0.928 & 0.981 \\
+ IID + PointEncoder & 0.918 & 0.35 & 0.920 & 0.981 \\
\midrule[\heavyrulewidth]
\texttt{DITTO} (ours) & \bestred{0.931} & \bestred{0.33} & \bestred{0.931} & \bestred{0.984} \\
\bottomrule
\end{tabular}
}
\end{table}

We evaluate the impact of each proposed module by incrementally incorporating them into baseline ALTO (see \Tref{exp.ablation.modules}).
DLL demonstrates significant performance improvements regardless of the type of point backbones.
This result reveals that the point feature extraction module that learns spatial patterns of point latents is more crucial than the simple MLP of ALTO.
Additionally, while NC and F-score slightly decrease with IID alone, other metrics improve.
However, combining DLL with IID significantly boosts performance across all metrics.
We deduce that the variation in performance is due to the latent points without DLL not being optimized to learn spatial patterns.

\subsection{Real-World 3D Surface Reconstruction}

We conduct an additional experiment to assess generality (see \Tref{exp.scannet} and \Fref{fig:scannet}).
\texttt{DITTO} demonstrates superior performance than previous methods.
\texttt{DITTO} successfully restores the sofas (see red and yellow boxes), where other methods encounter difficulties.
We also evaluate the performance using triplanes as grid latents (see \textit{Triplane comparison for grid latents}).
Unlike the scene-level results in \Sref{sec:scene-level}, triplanes generally show poorer performance on ScanNet-v2.
We attribute this performance issue to a lack of geometric inductive bias in triplanes.
Since each plane in a triplane misses information in a direction, triplane latents are required to learn spatial rules that the voxels naturally have.
This limitation poses challenges for triplanes when reconstructing out-of-distribution data.
Despite these challenges, \texttt{DITTO} shows better performance than other grid-based methods with triplanes.

\begin{table}[t]
\centering
\caption{
    \textbf{Quantitative comparison on ScanNet-v2~\cite{dai2017scannet}.}
    We test models pre-trained with the Synthetic Rooms dataset~\cite{peng2020convolutional} on ScanNet-v2, which is not used during training.
    Since the dataset has no ground truth occupancy, we only measure Chamfer distance and F-score, following \cite{wang2023alto}. 
}
\vspace{-2mm}
\resizebox{.8\linewidth}{!}{%
\begin{tabular}{c|cc|cc} \toprule \label{exp.scannet}
\multirow{2}{*}{Method} & \multicolumn{2}{c|}{$N_\text{train},N_\text{test}=10\text{K},10\text{K}$} & \multicolumn{2}{c}{$N_\text{train},N_\text{test}=10\text{K},3\text{K}$} \\
& Chamfer-$L_1$ & F-score & Chamfer-$L_1$ & F-score \\
\midrule[\heavyrulewidth]
ConvONet~\cite{peng2020convolutional} & 1.02 & 0.694 & 1.01 & 0.719 \\
POCO~\cite{boulch2022poco} & 0.87 & 0.757 & 0.93 & 0.737 \\
ALTO~\cite{wang2023alto} & \bestblue{0.79} & \bestblue{0.779} & \bestblue{0.87} & \bestblue{0.746} \\
\midrule[\heavyrulewidth]
\texttt{DITTO} (ours) & \bestred{0.70} & \bestred{0.808} & \bestred{0.78} & \bestred{0.773} \\
\hline \hline
\multicolumn{5}{c}{\large\emph{Triplane comparison}} \\
\midrule[\heavyrulewidth]
ConvONet~\cite{peng2020convolutional} & 1.45 & 0.636 & 1.55 & 0.614 \\
ALTO~\cite{wang2023alto} & \bestblue{1.43} & \bestblue{0.640} & \bestblue{1.44} & \bestblue{0.601} \\
\midrule[\heavyrulewidth]
\texttt{DITTO} (ours) & \bestred{1.26} & \bestred{0.677} & \bestred{1.26} & \bestred{0.660} \\
\bottomrule
\end{tabular}
}
\end{table}

\section{Conclusion}

We have proposed \texttt{DITTO}, a novel concept of dual and integrated latent topologies for implicit 3D reconstruction from noisy and sparse point clouds.
Specifically, we have studied the use of grid and point latents together as dual latent to integrate their own strengths. 
To this end, we have proposed the DLL architecture with the DSPT module for enhancing dual latent while maintaining their original shape at the encoder level. 
Then, we explored how to utilize both refined latents in the proposed integrated implicit decoder.
\texttt{DITTO} outperforms previous state-of-the-art implicit 3D reconstruction methods, especially \texttt{DITTO} facilitates the reconstruction of thin structures and intricate shape details.

\iftoggle{cvprpagenumbers}{}{
\vspace{1mm} \noindent \textbf{Acknowledgments.}
This work was supported by Institute of Information \& communications Technology Planning \& Evaluation~(IITP) grant funded by the Korea government~(MSIT)
(No. 2022-0-00612, Geometric and Physical Commonsense Reasoning based Behavior Intelligence for Embodied AI, No.2022-0-00907, Development of AI Bots Collaboration Platform and Self-organizing AI and No.2020-0-01336, Artificial Intelligence Graduate School Program~(UNIST)).
}

{
    \small
    \bibliographystyle{ieeenat_fullname}
    \bibliography{main}
}

\iftoggle{cvprpagenumbers}{
\clearpage
\onecolumn
\setcounter{page}{1}
\setcounter{section}{0}
\setcounter{figure}{0}
\setcounter{table}{0}
\maketitlesupplementary
\section*{Overview}

In this supplementary material, we provide detailed descriptions of \texttt{DITTO} that could not be handled in the main paper due to space constraints.
In \Sref{sec:supp:implementation}, we describe more details necessary for implementing our work, such as detailed network architecture and hyperparameters.
Additional ablation studies related to \texttt{DITTO} are available in \Sref{sec:supp:ablation}.
We provide additional experiment results in \Sref{supp:sec:exp}.
\iftoggle{issupp}{\blfootnote{*Corresponding author.}}{}

\section{Implementation Details} \label{sec:supp:implementation}

Additional details of the \texttt{DITTO} network architecture are available in \Sref{sec:supp:network}, and hyperparameters that are used for training \texttt{DITTO} are available in \Sref{sec:supp:hyperparameters}.

\subsection{Network Architecture Details} \label{sec:supp:network}

In this section, we provide additional network architecture details of our work.
\texttt{DITTO} mainly consists of the dual latent encoder and the integrated implicit decoder (IID).
The dual latent encoder can be subdivided into a point encoder and an UNet with dual latent layers (DLLs).
An illustration of the dual latent encoder is available in \Fref{supp:fig:dual_latent_encoder}.
We describe details of each of the modules below.
In addition, we provide layer-level details in \Tref{tab:supp:arch:point_encoder}.

\begin{figure}[h]
\centering
\vspace{-2mm}
\resizebox{.4\linewidth}{!}{
\includegraphics{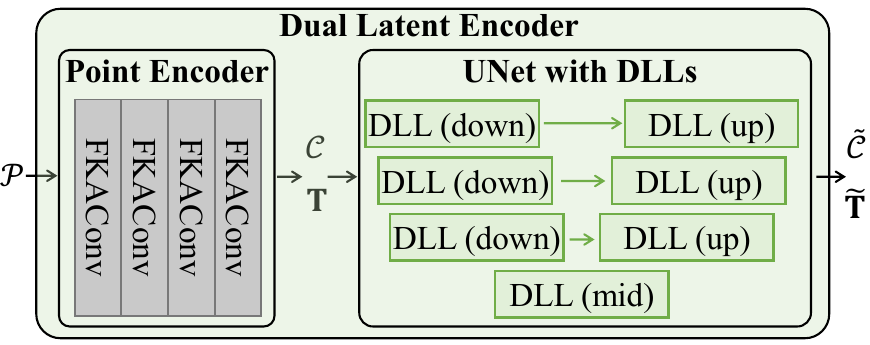}
}
\caption{
    \textbf{Detailed illustration of our dual latent encoder}.
}
\label{supp:fig:dual_latent_encoder}
\end{figure}

\vspace{1mm} \noindent \textbf{Point Encoder}.
The point encoder receives input point cloud $\mathcal P$, and generates point latents $\mathcal C$.
This module then produces grid latents $\mathbf T$ by projecting $\mathcal C$ onto either triplanes or voxels.
While extracting $\mathcal C$, we employ a stack of four FKAConv layers~\cite{boulch2020fkaconv} instead of the conventional architecture based on PointNet~\cite{liu2019point}, called as local pooled PointNet.
Local pooled PointNet generates point features by directly encoding the point coordinates, but this layer processes each point independently, without considering the relationship between points.
In contrast, simply employing a FKAConv-based encoder similar to POCO~\cite{boulch2022poco} gives additional performance gains.
This improvement is further reinforced by DLL and IID due to their emphasis on point latents.

\vspace{1mm} \noindent \textbf{Dual Latent Layer}.
Our DLL focuses on refining point latents.
To analyze what DLL learns, we visualize point features in \Fref{fig:supp:feature_visualization}.
The features of \texttt{DITTO} exhibit clear boundaries between different parts of the object.
For instance, the body of the airplane has a distinct color compared to its wings, and similarly, the bottom and side parts of the chair and the gun have different colors.
In contrast, most of the point features of ALTO~\cite{wang2023alto} have similar colors.
From this difference, we infer that \texttt{DITTO} appears to implicitly learn semantic information, such as planes and their direction or curvature.
We expect that these point features can help provide a clear surface boundary between two surfaces that are adjacent yet not in contact.
We would like to note that \texttt{DITTO} considers point-level geometry with a point feature extractor, such as the proposed DSPT, while ALTO handles point latents using MLPs that account for each point independently.

We provide a visual comparison of our DLL, along with the architecture design of ALTO (see \Fref{fig:supp:visual_comparison_of_dll}).
Our design of DLL mainly differs in two aspects: the DSPT layers for point features refinement and the skip-connection from point latents to grid latents.
ALTO outputs refined grid latents $\mathbf T'$ that are directly projected from the point latents.
These grid latents have many empty cells, hindering feature extraction of grid latents.
To address this problem, we find that simply creating a residual connection between grid latents and those projected from point latents can enhance performance (see our additional ablation study in \Sref{sec:supp:ablation:dll}).
Additionally, we incorporate a convolutional layer to reduce sparseness of grid latents projected from point latents.

\begin{figure}[h]
\centering

\begin{subfigure}{.4\linewidth}
    \includegraphics[width=\linewidth]{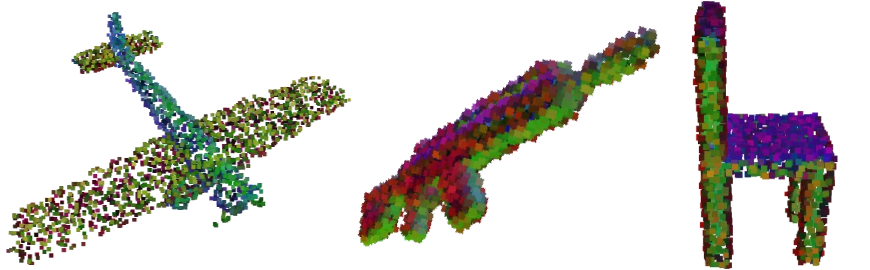}
    \caption{Point features of \texttt{DITTO}}
\end{subfigure}
\hspace{3mm}
\begin{subfigure}{.4\linewidth}
    \includegraphics[width=\linewidth]{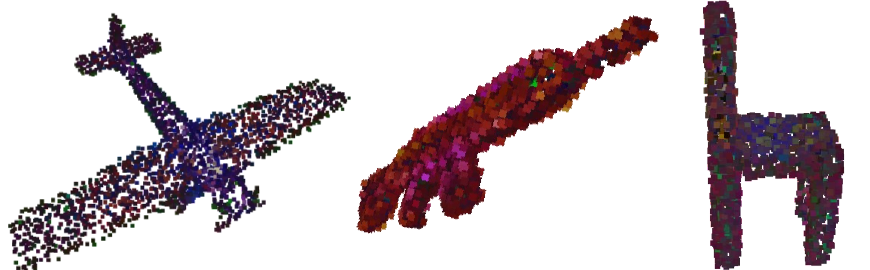}
    \caption{Point features of ALTO~\cite{wang2023alto}}
\end{subfigure}
\vspace{-2mm}
\caption{
    \textbf{Visualization of point features}.
    We visualize the of the refined point features $\tilde{\mathcal C}$.
    These features are colored by reducing feature dimension into three channels by using principal component analysis (PCA).
}
\vspace{-2mm}
\label{fig:supp:feature_visualization}
\end{figure}

\begin{figure}[h]
\centering

\begin{subfigure}{.44\linewidth}
    \includegraphics[width=\linewidth]{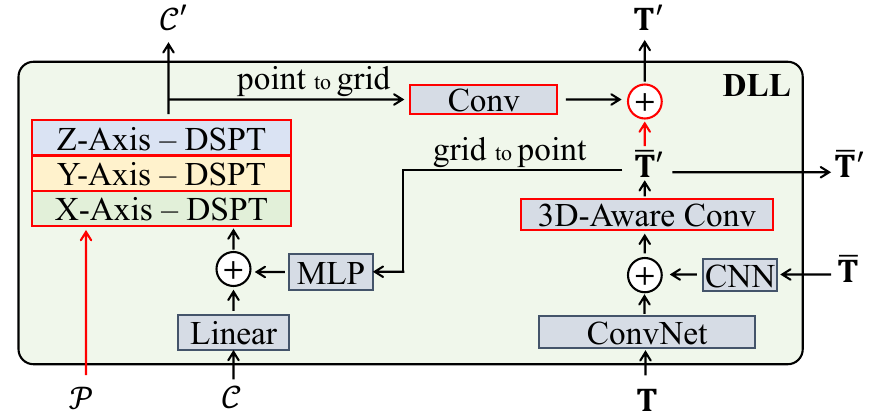}
    \caption{\texttt{DITTO} (ours)}
\end{subfigure}
\hspace{3mm}
\begin{subfigure}{.44\linewidth}
    \includegraphics[width=\linewidth]{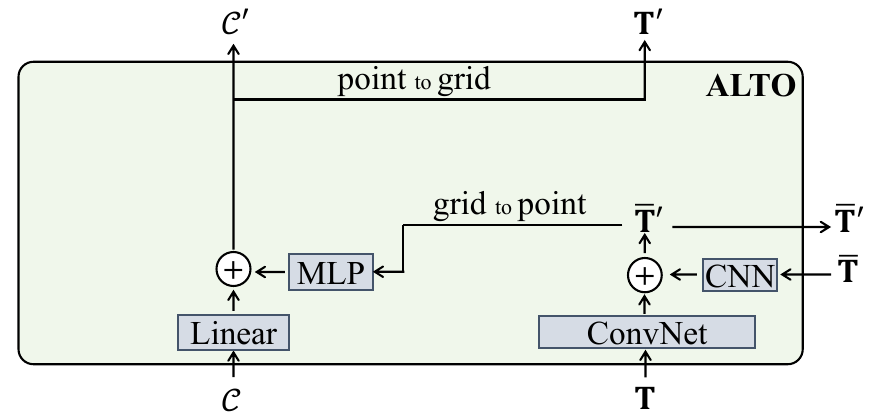}
    \caption{ALTO~\cite{wang2023alto}}
\end{subfigure}
\vspace{-2mm}
\caption{
    \textbf{Visual comparison of DLL and a layer of ALTO}.
    Key differences are highlighted with red outlines and red arrows.
}
\vspace{-2mm}
\label{fig:supp:visual_comparison_of_dll}
\end{figure}

\vspace{1mm} \noindent \textbf{Dynamic Sparse Transformer}.
IID receives point coordinates $\mathcal P$ and their features $\mathcal C$, and enhances the point features.
When dividing the point features into windows $\{\mathcal{C}^{\text{wnd}}_l\}_{l=1}^L$, we recycle the sorting indices in each DSPT layer to reduce computation load.
Concretely, since both the number and the coordinates of points remain constant throughout all processes of \texttt{DITTO}, we initially calculate the sorted indices of point coordinates and reuse them in every DSPT layer.
This method effectively reduces the computation needed for recalculating sorted indices.

\vspace{1mm} \noindent \textbf{UNet with Dual Latent Layers}.
Our UNet architecture is similar to traditional UNet~\cite{ronneberger2015u,cciccek20163d}, but each layer is replaced with the proposed DLL module.
Our UNet has three down DLLs, a mid DLL, and three up DLLs (see \Fref{supp:fig:dual_latent_encoder}).
The down DLL can optionally downsample the grid latents using MaxPooling, while the up DLL can upsample them using transposed convolution.
Specifically, within our UNet, the second and third down DLLs downsample the grid latents, whereas the first and second up DLLs upsample them.

\vspace{1mm} \noindent \textbf{Integrated Implicit Decoder}.
IID receives the refined latents and estimates occupancy by comprehensive consideration of these latents.
Note that, while \Tref{tab:supp:arch:point_encoder} describes IID with a single query point $\mathbf q$ for convenience, IID actually processes multiple query points $\mathcal Q$ in parallel.

\subsection{Hyperparameters} \label{sec:supp:hyperparameters}
The detailed hyperparameters can be found in \Tref{tab:supp:hyperparameter}.

\begin{table}[h]
\centering
\caption{
    \textbf{Detailed hyperparameters of \texttt{DITTO}.}
}
\vspace{-2mm}
\resizebox{.9\linewidth}{!}{%
\begin{tabular}{cc|ccc|cccc} \toprule \label{tab:supp:hyperparameter}
\multirow{2}{*}{Notation} & \multirow{2}{*}{Meaning} & Object (3K) & Object (1K) & Object(0.3K) &  \multicolumn{2}{c}{Scene (10K)} & \multicolumn{2}{c}{Scene (3K)} \\
&& Triplane & Triplane & Triplane & Triplane & Voxel & Triplane & Voxel \\
\midrule[\heavyrulewidth]
& Epoch & 1,000 & 1,000 & 1,000 & 2,500 & 2,500 & 2,500 & 2,500 \\
& Learning rate & 1e{-}4 & 1e{-}4 & 1e{-}4 & 1e{-}4 & 1e{-}4 & 1e{-}4 & 1e{-}4 \\
& Batch size & 32 & 32 & 32 & 32 & 16 & 32 & 16 \\
$R$ & Feature resolution & 64 & 64 & 64 & 128 & 64 & 128 & 64 \\
$d$ & Channel size & 32 & 32 & 32 & 32 & 32 & 32 & 32 \\
$L$ & \# of windows in DSPT & 25 & 25 & 25 & 20 & 20 & 25 & 25 \\
$K$ & \# of neighbor points in IID & 32 & 32 & 32 & 32 & 32 & 32 & 32 \\
$M$ & \# of query points per training iteration & 2,048 & 2,048 & 2,048 & 2,048 & 2,048 & 2,048 & 2,048 \\
\bottomrule
\end{tabular}
}
\end{table}

\section{Ablation Studies} \label{sec:supp:ablation}

We conduct ablation studies to demonstrate effectiveness of each module of \texttt{DITTO}.
Specifically, we perform ablation studies on the point encoder, a residual connection, and the number of windows for DSPT in \Sref{sec:supp:ablation:dll}, and \Sref{sec:supp:ablation:dspt}, respectively.

\subsection{Ablation Study on Residual Connection} \label{sec:supp:ablation:dll}

To demonstrate the impact of residual connections between grid latents and those projected from point latents, we conduct an ablation study comparing ALTO with and without this residual connection (see \Tref{tab:supp:ablation:dll}).
The results suggest that only a simple addition of residual connection can significantly enhance performance.
Moreover, a convolutional layer can improve performance by reducing the sparsity of grid latents projected from point latents.

\begin{table}[h]
\centering
\caption{
    \textbf{Ablation study on residual connection}.
    Training and inference are conducted on the Synthetic Rooms dataset with 10K input points and 0.005 noise level.
}
\vspace{-2mm}
\resizebox{.6\linewidth}{!}{%
\begin{tabular}{l|cccc} \toprule \label{tab:supp:ablation:dll}
\# of windows & IoU $\uparrow$ & Chamfer-$L_1$ $\downarrow$ & NC $\uparrow$ & F-score $\uparrow$ \\
\midrule[\heavyrulewidth]
ALTO (triplane) & 0.895 & 0.37 & 0.910 & 0.974 \\
ALTO + residual connection & 0.904 & \bestred{0.36} & \bestred{0.915} & 0.976 \\
ALTO + residual connection + conv layer & \bestred{0.907} & \bestred{0.36} & \bestred{0.915} & \bestred{0.977} \\
\bottomrule
\end{tabular}
}
\end{table}

\subsection{Ablation Study on DSPT} \label{sec:supp:ablation:dspt}

We conduct an ablation study to determine the appropriate number of windows for DSPT.
The results can be found in \Tref{tab:supp:ablation:dspt}.
The results demonstrate robustness across various window numbers.
However, the current window number ($L=20$) shows slightly improved performance.

\begin{table}[h]
\centering
\caption{
    \textbf{Ablation study on the number of windows in our DSPT}.
    The training and inference are conducted on the Synthetic Rooms dataset with 10K input points and 0.005 noise level.
}
\vspace{-2mm}
\resizebox{.4\linewidth}{!}{%
\begin{tabular}{l|cccc} \toprule \label{tab:supp:ablation:dspt}
\# of windows & IoU $\uparrow$ & Chamfer-$L_1$ $\downarrow$ & NC $\uparrow$ & F-score $\uparrow$ \\
\midrule[\heavyrulewidth]
40 & 0.929 & 0.34 & 0.930 & \bestred{0.984} \\
25 & 0.930 & 0.34 & 0.930 & 0.983 \\
20 (\texttt{DITTO)} & \bestred{0.931} & \bestred{0.33} & \bestred{0.931} & \bestred{0.984} \\
10 & 0.929 & 0.34 & 0.930 & \bestred{0.984} \\
\bottomrule
\end{tabular}
}
\end{table}


\section{Additional Experiment Results} \label{supp:sec:exp}

In this section, we provide additional qualitative results in Sections \ref{supp:sec:exp_shapenet}, \ref{supp:sec:exp_synthetic_rooms}, \ref{supp:sec:exp_scannet}.

\subsection{Additional Results on ShapeNet} \label{supp:sec:exp_shapenet}

\vspace{1mm} \noindent \textbf{Quantitative Results}.
We provide additional object-level 3D surface reconstruction results on ShapeNet~\cite{chang2015shapenet}.
Detailed per-category quantitative results are presented with different input point densities: 
3K input points in \Tref{supp:tab:shapenet1}, 
1K input points in \Tref{supp:tab:shapenet2}, 
and 0.3K input points in \Tref{supp:tab:shapenet3}, 
all at a consistent noise level of 0.005.
Each of these tables is a per-category extension to Table 1 in the main paper.
\texttt{DITTO} outperforms previous methods in most categories.
Note that, while ALTO~\cite{wang2023alto} outperforms POCO~\cite{boulch2022poco} in most of metrics, POCO shows higher F-score than ALTO when dealing with 3K input points.
In contrast, \texttt{DITTO} demonstrates superior performance in most metrics and categories.

\vspace{1mm} \noindent \textbf{Qualitative Results}.
We present additional object-level 3D reconstruction results on ShapeNet.
The result meshes are visualized in \Fref{fig:supp:shapenet1} for 3K input points at 0.005 noise level.
\texttt{DITTO} shows high-fidelity reconstruction especially thin and intricate structures such as the legs of the chairs and the tables (second and third rows).
Enhanced details are also notable, such as the boundary interface between two parts of the chair (second row).

To demonstrate robustness for sparsity, we visualize qualitative results in sparse case: 1K input points in \Fref{fig:supp:shapenet2}, 0.3K input points in \Fref{fig:supp:shapenet3} with consistent noise level.
Even with sparse point clouds, \texttt{DITTO} shows superior reconstruction quality in intricate shapes, such as the bookshelf (first row in \Fref{fig:supp:shapenet2}) and underside of the car (third row in \Fref{fig:supp:shapenet2}).
Moreover, the results of \texttt{DITTO} generate more clear shape details, such as the chair (second row in \Fref{fig:supp:shapenet2}).

\subsection{Additional Results on Synthetic Rooms} \label{supp:sec:exp_synthetic_rooms}

We provide additional qualitative results for scene-level 3D surface reconstruction on the Synthetic Rooms dataset~\cite{peng2020convolutional}.
We visualize the results in \Fref{fig:supp:synthetic_rooms} for 10K input points with 0.005 noise level.
Result meshes of \texttt{DITTO} exhibit clear surface boundaries in intricate cases, such as the bookshelf (box in the left scene), and the lamps (boxes in the middle and right scenes).
In addition, due to the precise detail reconstruction capability of \texttt{DITTO}, it can successfully reconstruct fine details of chairs (boxes in the middle and right scenes) and lamps (boxes in the middle scene).

To demonstrate the performance with sparse input point clouds, we visualize the results for 3K input points in \Fref{fig:supp:synthetic_rooms_sparse}.
Even with sparse point clouds, \texttt{DITTO} outperforms previous methods.
\texttt{DITTO} is the only method that successfully reconstructs the chairs (boxes in the left and middle scenes).
In addition, \texttt{DITTO} is the most successful method in reconstructing the bookshelves (boxes in the right scene).

\subsection{Additional Results on ScanNet-V2} \label{supp:sec:exp_scannet}

We present additional qualitative results on ScanNet-v2~\cite{dai2017scannet} to demonstrate generalization performance.
Consistent with the main paper, we test on this dataset using models pre-trained with the Synthetic Rooms dataset (see \Fref{fig:supp:scannet}).
\texttt{DITTO} successfully reconstructs the tables and sofas (boxes in the left and middle scenes).
For the right scene, our method successfully reconstructs the tables and chairs, even though every previous method fails to generate the scene accurately.
These results demonstrate the robustness of \texttt{DITTO} in handling complex geometries and details, even in the dataset that is not used during the training phase.


\begin{table}[p]
\centering
\caption{
    \textbf{Layer-level network module architecture designs of \texttt{DITTO}}.
}
\vspace{-2mm}
\resizebox{.9\linewidth}{!}{%
\begin{tabular}{lll} \toprule \label{tab:supp:arch:point_encoder}
Layer Name & Input & Output \\
\midrule[\heavyrulewidth]

\textbf{Point Encoder} \\
Input / Output & $\mathcal P\ (N \times 3)$ & $\mathcal C\ (N \times d), \mathbf T\ (3 \times R \times R \times d)$ \\
\hline
FKAConv layer & $\mathcal P\ (N \times 3)$ & $\mathcal C\ (N \times d)$ \\
FKAConv layer & $\mathcal P\ (N \times 3), \mathcal C\ (N \times d)$ & $\mathcal C\ (N \times d)$ \\
FKAConv layer & $\mathcal P\ (N \times 3), \mathcal C\ (N \times d)$ & $\mathcal C\ (N \times d)$ \\
FKAConv layer & $\mathcal P\ (N \times 3), \mathcal C\ (N \times d)$ & $\mathcal C\ (N \times d)$ \\
Quantization & $\mathcal P\ (N \times 3), \mathcal C\ (N \times d)$ & $\mathbf T\ (3 \times R \times R \times d)$ \\
\midrule[\heavyrulewidth] \midrule[\heavyrulewidth]

\textbf{DLL} \\
Input / Output
    & $\mathbf T\ (3 \times R \times R \times d_\text{in}), \mathcal P\ (N \times 3), \mathcal C\ (N \times d_\text{in})$
    & $\mathbf T'\ (3 \times R/2 \times R/2 \times d_\text{out}), \mathcal C'\ (N \times d_\text{out})$ \\
\hline
ConvNet & $\mathbf T\ (3 \times R \times R \times d_\text{in})$ & $\mathbf T\ (3 \times R \times R \times d_\text{out})$ \\
Conv2d & $\bar{\mathbf T}\ (3 \times R \times R \times d_\text{in})$ & $\bar{\mathbf T}\ (3 \times R \times R \times d_\text{out})$ \\
Sum & $\mathbf T\ (3 \times R \times R \times d_\text{out}), \bar{\mathbf T}\ (3 \times R \times R \times d_\text{out})$ & $\bar{\mathbf T}\ (3 \times R \times R \times d_\text{out})$ \\
3D-Aware-Conv & $\bar{\mathbf T}\ (3 \times R \times R \times d_\text{out})$ & $\bar{\mathbf T}'\ (3 \times R \times R \times d_\text{out})$ \\
Linear & $\mathcal C\ (N \times d_\text{in})$ & $\mathcal C\ (N \times d_\text{out})$ \\
Grid-to-Point & $\bar{\mathbf T}'\ (3 \times R \times R \times d_\text{out}), \mathcal P\ (N \times 3)$ & $(N \times d_\text{out})$ \\
MLP & $(N \times d_\text{out})$ & $(N \times d_\text{out})$ \\
Sum & $(N \times d_\text{out}), \mathcal C\ (N \times d_\text{out})$ & $\mathcal C\ (N \times d_\text{out})$ \\
X-Axis DSPT & $\mathcal P(N \times 3), \mathcal C\ (N \times d_\text{out})$ & $\mathcal C\ (N \times d_\text{out})$ \\
Y-Axis DSPT & $\mathcal P(N \times 3), \mathcal C\ (N \times d_\text{out})$ & $\mathcal C\ (N \times d_\text{out})$ \\
Z-Axis DSPT & $\mathcal P(N \times 3), \mathcal C\ (N \times d_\text{out})$ & $\mathcal C'\ (N \times d_\text{out})$ \\
Point-to-Grid & $\mathcal P(N \times 3), \mathcal C'\ (N \times d_\text{out})$ & $(3 \times R \times R \times d_\text{out})$ \\
Conv2d & $(3 \times R \times R \times d_\text{out})$ & $(3 \times R \times R \times d_\text{out})$ \\
Sum & $(3 \times R \times R \times d_\text{out}), \bar{\mathbf T}'\ (3 \times R \times R \times d_\text{out})$ & $\mathbf T'\ (3 \times R \times R \times d_\text{out})$ \\
Pooling & $\mathbf T'\ (3 \times R \times R \times d_\text{out})$ & $\mathbf T'\ (3 \times R/2 \times R/2 \times d_\text{out})$ \\
\midrule[\heavyrulewidth] \midrule[\heavyrulewidth]

\textbf{UNet with DLLs}  \\
Input / Output
    & $\mathcal P\ (N \times 3), \mathcal C\ (N \times d), \mathbf T\ (3 \times R \times R \times d)$
    & $\mathbf T'\ (3 \times R \times R \times d), \tilde{\mathcal C}\ (N \times d)\textbf{}$ \\
\hline
DLL (down) 
    & $\mathcal P\ (N \times 3), \mathcal C\ (N \times d), \mathbf T\ (3 \times R \times R \times d)$ 
    & $\mathcal C'\ (N \times d), \mathbf T'\ (3 \times R \times R \times d)$ \\
DLL (down) 
    & $\mathcal P\ (N \times 3), \mathcal C\ (N \times d), \mathbf T\ (3 \times R \times R \times d)$ 
    & $\mathcal C'\ (N \times 2d), \mathbf T'\ (3 \times R/2 \times R/2 \times 2d)$ \\
DLL (down) 
    & $\mathcal P\ (N \times 3), \mathcal C\ (N \times 2d), \mathbf T\ (3 \times R/2 \times R/2 \times 2d)$ 
    & $\mathcal C'\ (N \times 4d), \mathbf T'\ (3 \times R/4 \times R/4 \times 4d)$ \\
DLL (mid) 
    & $\mathcal P\ (N \times 3), \mathcal C\ (N \times 4d), \mathbf T\ (3 \times R/4 \times R/4 \times 4d)$ 
    & $\mathcal C'\ (N \times 8d), \mathbf T'\ (3 \times R/4 \times R/4 \times 8d)$ \\
DLL (up) 
    & $\mathcal P\ (N \times 3), \mathcal C\ (N \times 8d), \mathbf T\ (3 \times R/4 \times R/4 \times 8d)$ 
    & $\mathcal C'\ (N \times 4d), \mathbf T'\ (3 \times R/2 \times R/2 \times 4d)$ \\
DLL (up) 
    & $\mathcal P\ (N \times 3), \mathcal C\ (N \times 4d), \mathbf T\ (3 \times R/2 \times R/2 \times 4d)$ 
    & $\mathcal C'\ (N \times 2d), \mathbf T'\ (3 \times R \times R \times 2d)$ \\
DLL (up) 
    & $\mathcal P\ (N \times 3), \mathcal C\ (N \times 2d), \mathbf T\ (3 \times R \times R \times 2d)$ 
    & $\mathcal C'\ (N \times d), \mathbf T'\ (3 \times R \times R \times d)$ \\
\midrule[\heavyrulewidth] \midrule[\heavyrulewidth]

\textbf{DSPT} \\
Input / Output & $\mathcal P\ (N \times 3), \mathcal C\ (N \times d)$ & $\mathcal C\ (N \times d)$ \\
\hline
\textit{sort} 
    & $\mathcal P\ (N \times 3), \mathcal C\ (N \times d)$ 
    & $(N \times d)$ \\
\textit{split}
    & $(N \times d)$
    & $\{\mathcal{C}^{\text{wnd}}_l\}_{l=1}^L$ \\
Self-attention
    & $\{\mathcal{C}^{\text{wnd}}_l\}_{l=1}^L$
    & $\{\mathcal{C}^{\text{wnd}}_l\}_{l=1}^L$ \\
\textit{split}$^{{-}1}$ 
    & $\{\mathcal{C}^{\text{wnd}}_l\}_{l=1}^L$
    & $(N \times d)$ \\
\textit{sort}$^{{-}1}$
    & $(N \times d)$
    & $\mathcal C\ (N \times d)$ \\
Layer norm & $\mathcal C\ (N \times d)$ & $\mathcal C\ (N \times d)$ \\
Linear & $\mathcal C\ (N \times d)$ & $\mathcal C\ (N \times 4d)$ \\
ReLU & $\mathcal C\ (N \times 4d)$ & $\mathcal C\ (N \times 4d)$ \\
Linear & $\mathcal C\ (N \times 4d)$ & $\mathcal C\ (N \times d)$ \\
\midrule[\heavyrulewidth] \midrule[\heavyrulewidth]

\textbf{IID} \\
Input / Output 
    & $\tilde{\mathbf T}\ (3 \times R \times R \times d), \mathcal P\ (N \times 3), \tilde{\mathcal C}\ (N \times d), \mathbf q\ (3)$ 
    & $\tilde{o}\ (1)$ \\
\hline
Interpolation & $\tilde{\mathbf T}\ (3 \times R \times R \times d), \mathbf q\ (3)$ & $(d)$ \\
Linear & $(d)$ & $\mathbf z_0\ (2d)$ \\

\textit{KNN }
    & $\mathbf q\ (3), \mathcal P\ (N \times 3), \tilde{\mathcal C}\ (N \times d)$ 
    & $\mathcal P^\mathbf q\ (K \times 3), \tilde{\mathcal C}^\mathbf q\ (K \times d)$ \\
\textit{interpolation} 
    & $\tilde{\mathbf T}\ (3 \times R \times R \times d), \mathcal P^\mathbf q\ (K \times 3)$ 
    & $\tilde{\mathbf T}^\mathbf q\ (K \times d)$ \\
\textit{concat}
    & $\tilde{\mathcal C}^\mathbf q\ (K \times d), \tilde{\mathbf T}^\mathbf q\ (K \times d)$
    & $\mathcal Z^\mathbf q\ (K \times 2d)$ \\
Self-attention
    & $\{\mathbf z_0^\mathbf q, \dots, \mathbf z_K^\mathbf q\}\ ((K{+}1) \times 2d), \{\mathbf q, \mathbf p_1^\mathbf q, \dots, \mathbf p_K^\mathbf q\}\ ((K{+}1) \times 3)$
    & $\mathbf z_0^\mathbf q\ (2d)$ \\
Self-attention
    & $\{\mathbf z_0^\mathbf q, \dots, \mathbf z_K^\mathbf q\}\ ((K{+}1) \times 2d), \{\mathbf q, \mathbf p_1^\mathbf q, \dots, \mathbf p_K^\mathbf q\}\ ((K{+}1) \times 3)$
    & $\mathbf z_0^\mathbf q\ (2d)$ \\
Self-attention
    & $\{\mathbf z_0^\mathbf q, \dots, \mathbf z_K^\mathbf q\}\ ((K{+}1) \times 2d), \{\mathbf q, \mathbf p_1^\mathbf q, \dots, \mathbf p_K^\mathbf q\}\ ((K{+}1) \times 3)$
    & $\mathbf z_0^\mathbf q\ (2d)$ \\
Self-attention
    & $\{\mathbf z_0^\mathbf q, \dots, \mathbf z_K^\mathbf q\}\ ((K{+}1) \times 2d), \{\mathbf q, \mathbf p_1^\mathbf q, \dots, \mathbf p_K^\mathbf q\}\ ((K{+}1) \times 3)$
    & $\hat{\mathbf z}_0^\mathbf q\ (2d)$ \\
Linear & $\hat{\mathbf z}_0^\mathbf q\ (2d)$ & $\tilde{o}\ (1)$ \\
\bottomrule
\end{tabular}
}
\end{table}

\begin{table}[p]
\centering
\caption{
    \textbf{Object-level quantitative comparison on ShapeNet with 3K input points with noise level 0.005}.
}
\vspace{-2mm}
\resizebox{\linewidth}{!}{%
\begin{tabular}{c|ccccc|ccccc} \toprule \label{supp:tab:shapenet1}
\multirow{3}{*}{Method} & \multicolumn{5}{c|}{IoU $\uparrow$} & \multicolumn{5}{c}{Chamfer-$L_1\ \downarrow$} \\
& ONet~\cite{mescheder2019occupancy} & ConvONet~\cite{peng2020convolutional} & POCO~\cite{boulch2022poco} & ALTO~\cite{wang2023alto} & \texttt{DITTO} (ours) & ONet~\cite{mescheder2019occupancy} & ConvONet~\cite{peng2020convolutional} & POCO~\cite{boulch2022poco} & ALTO~\cite{wang2023alto} & \texttt{DITTO} (ours) \\
\midrule[\heavyrulewidth]

Airplane & 0.734 & 0.849 & 0.902 & \bestblue{0.908} & \bestred{0.935} & 0.64 & 0.34 & 0.23 & \bestblue{0.22} & \bestred{0.19} \\
Bench & 0.682 & 0.830 & 0.865 & \bestblue{0.890} & \bestred{0.919} & 0.67 & 0.35 & 0.28 & \bestblue{0.26} & \bestred{0.23} \\
Cabinet & 0.855 & 0.940 & 0.960 & \bestblue{0.965} & \bestred{0.976} & 0.82 & 0.46 & 0.37 & \bestblue{0.34} & \bestred{0.31} \\
Car & 0.830 & 0.886 & 0.921 & \bestblue{0.924} & \bestred{0.943} & 1.04 & 0.75 & \bestblue{0.41} & 0.43 & \bestred{0.36} \\
Chair & 0.720 & 0.871 & 0.919 & \bestblue{0.925} & \bestred{0.948} & 0.95 & 0.46 & 0.33 & \bestblue{0.32} & \bestred{0.29} \\
Display & 0.799 & 0.927 & 0.956 & \bestblue{0.962} & \bestred{0.973} & 0.82 & 0.36 & 0.28 & \bestblue{0.27} & \bestred{0.25} \\
Lamp & 0.546 & 0.785 & \bestblue{0.877} & 0.868 & \bestred{0.914} & 1.59 & 0.59 & \bestblue{0.33} & 0.34 & \bestred{0.28} \\
Loudspeaker & 0.826 & 0.918 & \bestblue{0.957} & 0.953 & \bestred{0.970} & 1.18 & 0.64 & \bestblue{0.41} & \bestblue{0.41} & \bestred{0.35} \\
Rifle & 0.668 & 0.846 & 0.897 & \bestblue{0.898} & \bestred{0.925} & 0.66 & 0.28 & \bestblue{0.19} & \bestblue{0.19} & \bestred{0.16} \\
Sofa & 0.865 & 0.936 & 0.963 & \bestblue{0.966} & \bestred{0.976} & 0.73 & 0.42 & 0.30 & \bestblue{0.29} & \bestred{0.26} \\
Table & 0.739 & 0.888 & 0.924 & \bestblue{0.937} & \bestred{0.956} & 0.76 & 0.38 & 0.31 & \bestblue{0.29} & \bestred{0.27} \\
Telephone & 0.896 & 0.955 & 0.968 & \bestblue{0.977} & \bestred{0.982} & 0.46 & 0.27 & 0.22 & \bestblue{0.21} & \bestred{0.20} \\
Vessel & 0.729 & 0.865 & \bestblue{0.927} & 0.924 & \bestred{0.948} & 0.94 & 0.43 & \bestblue{0.25} & 0.26 & \bestred{0.22} \\
mean & 0.761 & 0.884 & 0.926 & \bestblue{0.931} & \bestred{0.949} & 0.87 & 0.44 & \bestblue{0.30} & \bestblue{0.30} & \bestred{0.27} \\

\hline \hline

\multirow{3}{*}{Method} & \multicolumn{5}{c|}{NC $\uparrow$} & \multicolumn{5}{c}{F-Score $\uparrow$} \\
& ONet~\cite{mescheder2019occupancy} & ConvONet~\cite{peng2020convolutional} & POCO~\cite{boulch2022poco} & ALTO~\cite{wang2023alto} & \texttt{DITTO} (ours) & ONet~\cite{mescheder2019occupancy} & ConvONet~\cite{peng2020convolutional} & POCO~\cite{boulch2022poco} & ALTO~\cite{wang2023alto} & \texttt{DITTO} (ours) \\
\midrule[\heavyrulewidth]

Airplane & 0.886 & 0.931 & 0.944 & \bestblue{0.949} & \bestred{0.958} & 0.829 & 0.965 & \bestblue{0.994} & 0.992 & \bestred{0.997} \\
Bench & 0.871 & 0.921 & 0.928 & \bestblue{0.941} & \bestred{0.950} & 0.827 & 0.964 & 0.988 & \bestblue{0.991} & \bestred{0.996} \\
Cabinet & 0.913 & 0.956 & 0.961 & \bestblue{0.967} & \bestred{0.970} & 0.833 & 0.956 & 0.979 & \bestblue{0.982} & \bestred{0.989} \\
Car & 0.874 & 0.893 & 0.894 & \bestred{0.917} & \bestblue{0.914} & 0.747 & 0.849 & \bestblue{0.946} & 0.940 & \bestred{0.963} \\
Chair & 0.886 & 0.943 & 0.956 & \bestblue{0.959} & \bestred{0.968} & 0.730 & 0.939 & \bestblue{0.985} & \bestblue{0.985} & \bestred{0.994} \\
Display & 0.926 & 0.968 & 0.975 & \bestblue{0.976} & \bestred{0.981} & 0.795 & 0.971 & \bestblue{0.994} & 0.993 & \bestred{0.997} \\
Lamp & 0.809 & 0.900 & \bestblue{0.929} & 0.924 & \bestred{0.942} & 0.581 & 0.892 & \bestblue{0.975} & 0.962 & \bestred{0.984} \\
Loudspeaker & 0.903 & 0.939 & \bestblue{0.952} & 0.951 & \bestred{0.961} & 0.727 & 0.892 & \bestblue{0.964} & 0.955 & \bestred{0.976} \\
Rifle & 0.849 & 0.929 & \bestblue{0.949} & \bestblue{0.949} & \bestred{0.960} & 0.818 & 0.980 & \bestblue{0.998} & 0.996 & \bestred{0.999} \\
Sofa & 0.928 & 0.958 & 0.967 & \bestblue{0.971} & \bestred{0.975} & 0.832 & 0.953 & \bestblue{0.989} & 0.987 & \bestred{0.994} \\
Table & 0.917 & 0.959 & 0.966 & \bestblue{0.968} & \bestred{0.975} & 0.824 & 0.967 & \bestblue{0.991} & 0.990 & \bestred{0.996} \\
Telephone & 0.970 & 0.983 & 0.985 & \bestblue{0.987} & \bestred{0.988} & 0.930 & 0.989 & \bestblue{0.998} & \bestblue{0.998} & \bestred{0.999} \\
Vessel & 0.857 & 0.919 & \bestblue{0.940} & \bestblue{0.940} & \bestred{0.952} & 0.734 & 0.931 & \bestblue{0.989} & 0.982 & \bestred{0.992} \\
mean & 0.891 & 0.938 & 0.950 & \bestblue{0.954} & \bestred{0.957} & 0.785 & 0.942 & \bestblue{0.984} & 0.981 & \bestred{0.988} \\

\midrule[\heavyrulewidth]
\bottomrule
\end{tabular}
}
\end{table}

\begin{table}[p]
\centering
\caption{
    \textbf{Object-level quantitative comparison on ShapeNet with 1K input points with noise level 0.005}.
}
\vspace{-2mm}
\resizebox{\linewidth}{!}{%
\begin{tabular}{c|ccccc|ccccc} \toprule \label{supp:tab:shapenet2}
\multirow{3}{*}{Method} & \multicolumn{5}{c|}{IoU $\uparrow$} & \multicolumn{5}{c}{Chamfer-$L_1\ \downarrow$} \\
& ONet~\cite{mescheder2019occupancy} & ConvONet~\cite{peng2020convolutional} & POCO~\cite{boulch2022poco} & ALTO~\cite{wang2023alto} & \texttt{DITTO} (ours) & ONet~\cite{mescheder2019occupancy} & ConvONet~\cite{peng2020convolutional} & POCO~\cite{boulch2022poco} & ALTO~\cite{wang2023alto} & \texttt{DITTO} (ours) \\
\midrule[\heavyrulewidth]

Airplane & 0.748 & 0.825 & 0.850 & \bestblue{0.872} & \bestred{0.908} & 0.59 & 0.39 & 0.32 & \bestblue{0.29} & \bestred{0.23} \\
Bench & 0.702 & 0.798 & 0.804 & \bestblue{0.856} & \bestred{0.891} & 0.62 & 0.40 & 0.38 & \bestblue{0.30} & \bestred{0.26} \\
Cabinet & 0.862 & 0.926 & 0.936 & \bestblue{0.953} & \bestred{0.964} & 0.76 & 0.50 & 0.46 & \bestblue{0.37} & \bestred{0.35} \\
Car & 0.837 & 0.867 & 0.878 & \bestblue{0.901} & \bestred{0.921} & 0.99 & 0.83 & 0.60 & \bestblue{0.50} & \bestred{0.45} \\
Chair & 0.736 & 0.837 & 0.867 & \bestblue{0.894} & \bestred{0.922} & 0.89 & 0.55 & 0.44 & \bestblue{0.39} & \bestred{0.33} \\
Display & 0.812 & 0.911 & 0.930 & \bestblue{0.946} & \bestred{0.960} & 0.78 & 0.41 & 0.34 & \bestblue{0.31} & \bestred{0.28} \\
Lamp & 0.567 & 0.741 & 0.807 & \bestblue{0.820} & \bestred{0.877} & 1.44 & 0.68 & \bestblue{0.50} & \bestblue{0.50} & \bestred{0.35} \\
Loudspeaker & 0.831 & 0.899 & 0.923 & \bestblue{0.933} & \bestred{0.951} & 1.14 & 0.72 & 0.54 & \bestblue{0.48} & \bestred{0.42} \\
Rifle & 0.680 & 0.801 & 0.850 & \bestblue{0.862} & \bestred{0.892} & 0.63 & 0.36 & 0.27 & \bestblue{0.25} & \bestred{0.20} \\
Sofa & 0.873 & 0.921 & 0.937 & \bestblue{0.952} & \bestred{0.964} & 0.69 & 0.47 & 0.38 & \bestblue{0.33} & \bestred{0.30} \\
Table & 0.757 & 0.858 & 0.880 & \bestblue{0.913} & \bestred{0.937} & 0.70 & 0.44 & 0.38 & \bestblue{0.33} & \bestred{0.30} \\
Telephone & 0.897 & 0.946 & 0.953 & \bestblue{0.968} & \bestred{0.975} & 0.46 & 0.29 & 0.26 & \bestblue{0.23} & \bestred{0.21} \\
Vessel & 0.736 & 0.840 & 0.880 & \bestblue{0.893} & \bestred{0.923} & 0.91 & 0.51 & 0.37 & \bestblue{0.33} & \bestred{0.27} \\
mean & 0.772 & 0.859 & 0.884 & \bestblue{0.905} & \bestred{0.926} & 0.82 & 0.50 & 0.40 & \bestblue{0.35} & \bestred{0.32} \\

\hline \hline

\multirow{3}{*}{Method} & \multicolumn{5}{c|}{NC $\uparrow$} & \multicolumn{5}{c}{F-Score $\uparrow$} \\
& ONet~\cite{mescheder2019occupancy} & ConvONet~\cite{peng2020convolutional} & POCO~\cite{boulch2022poco} & ALTO~\cite{wang2023alto} & \texttt{DITTO} (ours) & ONet~\cite{mescheder2019occupancy} & ConvONet~\cite{peng2020convolutional} & POCO~\cite{boulch2022poco} & ALTO~\cite{wang2023alto} & \texttt{DITTO} (ours) \\
\midrule[\heavyrulewidth]

Airplane & 0.894 & 0.922 & 0.920 & \bestblue{0.933} & \bestred{0.949} & 0.850 & 0.946 & 0.970 & \bestblue{0.976} & \bestred{0.990} \\
Bench & 0.882 & 0.911 & 0.902 & \bestblue{0.925} & \bestred{0.940} & 0.849 & 0.943 & 0.956 & \bestblue{0.979} & \bestred{0.990} \\
Cabinet & 0.925 & 0.949 & 0.945 & \bestblue{0.957} & \bestred{0.964} & 0.852 & 0.939 & 0.951 & \bestblue{0.972} & \bestred{0.978} \\
Car & \bestred{0.904} & 0.885 & 0.867 & 0.889 & \bestred{0.904} & 0.763 & 0.819 & 0.868 & \bestblue{0.912} & \bestred{0.934} \\
Chair & 0.893 & 0.931 & 0.930 & \bestblue{0.946} & \bestred{0.960} & 0.753 & 0.902 & 0.943 & \bestblue{0.965} & \bestred{0.982} \\
Display & 0.930 & 0.961 & 0.962 & \bestblue{0.970} & \bestred{0.976} & 0.805 & 0.956 & 0.976 & \bestblue{0.984} & \bestred{0.991} \\
Lamp & 0.820 & 0.885 & 0.895 & \bestblue{0.905} & \bestred{0.929} & 0.606 & 0.845 & 0.924 & \bestblue{0.926} & \bestred{0.964} \\
Loudspeaker & 0.914 & 0.929 & 0.928 & \bestblue{0.936} & \bestred{0.950} & 0.740 & 0.863 & 0.908 & \bestblue{0.926} & \bestred{0.951} \\
Rifle & 0.859 & 0.916 & 0.928 & \bestblue{0.936} & \bestred{0.949} & 0.828 & 0.957 & 0.984 & \bestblue{0.987} & \bestred{0.994} \\
Sofa & 0.937 & 0.950 & 0.950 & \bestblue{0.960} & \bestred{0.969} & 0.846 & 0.932 & 0.961 & \bestblue{0.974} & \bestred{0.985} \\
Table & 0.918 & 0.950 & 0.949 & \bestblue{0.961} & \bestred{0.970} & 0.842 & 0.947 & 0.964 & \bestblue{0.979} & \bestred{0.989} \\
Telephone & 0.972 & 0.980 & 0.979 & \bestblue{0.984} & \bestred{0.986} & 0.940 & 0.983 & 0.990 & \bestblue{0.994} & \bestred{0.996} \\
Vessel & 0.866 & 0.906 & 0.913 & \bestblue{0.923} & \bestred{0.940} & 0.740 & 0.899 & 0.952 & \bestblue{0.961} & \bestred{0.979} \\
mean & 0.901 & 0.929 & 0.928 & \bestblue{0.940} & \bestred{0.949} & 0.801 & 0.918 & 0.950 & \bestblue{0.964} & \bestred{0.975} \\

\midrule[\heavyrulewidth]
\bottomrule
\end{tabular}
}
\end{table}

\begin{table}[p]
\centering
\caption{
    \textbf{Object-level quantitative comparison on ShapeNet with 0.3K input points with noise level 0.005}.
}
\vspace{-2mm}
\resizebox{\linewidth}{!}{%
\begin{tabular}{c|ccccc|ccccc} \toprule \label{supp:tab:shapenet3}
\multirow{3}{*}{Method} & \multicolumn{5}{c|}{IoU $\uparrow$} & \multicolumn{5}{c}{Chamfer-$L_1\ \downarrow$} \\
& ONet~\cite{mescheder2019occupancy} & ConvONet~\cite{peng2020convolutional} & POCO~\cite{boulch2022poco} & ALTO~\cite{wang2023alto} & \texttt{DITTO} (ours) & ONet~\cite{mescheder2019occupancy} & ConvONet~\cite{peng2020convolutional} & POCO~\cite{boulch2022poco} & ALTO~\cite{wang2023alto} & \texttt{DITTO} (ours) \\
\midrule[\heavyrulewidth]

Airplane & 0.760 & 0.782 & 0.744 & \bestblue{0.825} & \bestred{0.857} & 0.57 & 0.48 & 0.57 & \bestblue{0.39} & \bestred{0.32} \\
Bench & 0.716 & 0.743 & 0.707 & \bestblue{0.801} & \bestred{0.835} & 0.60 & 0.50 & 0.56 & \bestblue{0.39} & \bestred{0.34} \\
Cabinet & 0.867 & 0.900 & 0.889 & \bestblue{0.927} & \bestred{0.941} & 0.73 & 0.52 & 0.58 & \bestblue{0.46} & \bestred{0.43} \\
Car & 0.834 & 0.843 & 0.817 & \bestblue{0.867} & \bestred{0.885} & 0.99 & 0.76 & 0.83 & \bestblue{0.67} & \bestred{0.61} \\
Chair & 0.736 & 0.787 & 0.776 & \bestblue{0.840} & \bestred{0.871} & 0.89 & 0.67 & 0.71 & \bestblue{0.52} & \bestred{0.45} \\
Display & 0.817 & 0.885 & 0.878 & \bestblue{0.917} & \bestred{0.931} & 0.76 & 0.47 & 0.49 & \bestblue{0.38} & \bestred{0.35} \\
Lamp & 0.567 & 0.663 & 0.681 & \bestblue{0.747} & \bestred{0.808} & 1.38 & 1.02 & 0.93 & \bestblue{0.76} & \bestred{0.61} \\
Loudspeaker & 0.827 & 0.870 & 0.867 & \bestblue{0.901} & \bestred{0.916} & 1.16 & 0.78 & 0.79 & \bestblue{0.64} & \bestred{0.59} \\
Rifle & 0.691 & 0.757 & 0.742 & \bestblue{0.801} & \bestred{0.832} & 0.61 & 0.43 & 0.45 & \bestblue{0.35} & \bestred{0.30} \\
Sofa & 0.872 & 0.898 & 0.893 & \bestblue{0.926} & \bestred{0.938} & 0.69 & 0.52 & 0.53 & \bestblue{0.42} & \bestred{0.38} \\
Table & 0.758 & 0.813 & 0.794 & \bestblue{0.868} & \bestred{0.894} & 0.72 & 0.52 & 0.57 & \bestblue{0.42} & \bestred{0.37} \\
Telephone & 0.916 & 0.939 & 0.927 & \bestblue{0.952} & \bestred{0.960} & 0.41 & 0.31 & 0.33 & \bestblue{0.27} & \bestred{0.25} \\
Vessel & 0.748 & 0.797 & 0.795 & \bestblue{0.846} & \bestred{0.872} & 0.85 & 0.63 & 0.60 & \bestblue{0.47} & \bestred{0.40} \\
mean & 0.778 & 0.821 & 0.808 & \bestblue{0.863} & \bestred{0.882} & 0.80 & 0.59 & 0.61 & \bestblue{0.47} & \bestred{0.43} \\

\hline \hline

\multirow{3}{*}{Method} & \multicolumn{5}{c|}{NC $\uparrow$} & \multicolumn{5}{c}{F-Score $\uparrow$} \\
& ONet~\cite{mescheder2019occupancy} & ConvONet~\cite{peng2020convolutional} & POCO~\cite{boulch2022poco} & ALTO~\cite{wang2023alto} & \texttt{DITTO} (ours) & ONet~\cite{mescheder2019occupancy} & ConvONet~\cite{peng2020convolutional} & POCO~\cite{boulch2022poco} & ALTO~\cite{wang2023alto} & \texttt{DITTO} (ours) \\
\midrule[\heavyrulewidth]

Airplane & 0.897 & 0.901 & 0.867 & \bestblue{0.914} & \bestred{0.931} & 0.864 & 0.902 & 0.867 & \bestblue{0.938} & \bestred{0.962} \\
Bench & 0.878 & 0.886 & 0.864 & \bestblue{0.906} & \bestred{0.920} & 0.860 & 0.912 & 0.882 & \bestblue{0.947} & \bestred{0.966} \\
Cabinet & 0.916 & 0.931 & 0.917 & \bestblue{0.943} & \bestred{0.953} & 0.856 & 0.916 & 0.896 & \bestblue{0.943} & \bestred{0.957} \\
Car & \bestblue{0.875} & 0.864 & 0.835 & 0.873 & \bestred{0.887} & 0.757 & 0.810 & 0.766 & \bestblue{0.850} & \bestred{0.879} \\
Chair & 0.889 & 0.905 & 0.885 & \bestblue{0.923} & \bestred{0.940} & 0.754 & 0.850 & 0.833 & \bestblue{0.910} & \bestred{0.941} \\
Display & 0.926 & 0.947 & 0.938 & \bestblue{0.956} & \bestred{0.964} & 0.813 & 0.926 & 0.916 & \bestblue{0.957} & \bestred{0.967} \\
Lamp & 0.813 & 0.853 & 0.834 & \bestblue{0.875} & \bestred{0.902} & 0.618 & 0.771 & 0.781 & \bestblue{0.857} & \bestred{0.908} \\
Loudspeaker & 0.897 & 0.911 & 0.897 & \bestblue{0.916} & \bestred{0.932} & 0.737 & 0.832 & 0.819 & \bestblue{0.871} & \bestred{0.899} \\
Rifle & 0.863 & 0.890 & 0.883 & \bestblue{0.909} & \bestred{0.925} & 0.838 & 0.919 & 0.918 & \bestblue{0.952} & \bestred{0.968} \\
Sofa & 0.928 & 0.935 & 0.924 & \bestblue{0.946} & \bestred{0.956} & 0.846 & 0.906 & 0.899 & \bestblue{0.941} & \bestred{0.956} \\
Table & 0.917 & 0.933 & 0.917 & \bestblue{0.945} & \bestred{0.957} & 0.839 & 0.913 & 0.894 & \bestblue{0.947} & \bestred{0.966} \\
Telephone & 0.970 & 0.975 & 0.970 & \bestblue{0.978} & \bestred{0.982} & 0.942 & 0.975 & 0.971 & \bestblue{0.984} & \bestred{0.988} \\
Vessel & 0.860 & 0.879 & 0.867 & \bestblue{0.898} & \bestred{0.914} & 0.758 & 0.850 & 0.851 & \bestblue{0.909} & \bestred{0.935} \\
mean & 0.895 & 0.908 & 0.892 & \bestblue{0.922} & \bestred{0.931} & 0.806 & 0.883 & 0.869 & \bestblue{0.924} & \bestred{0.940} \\

\midrule[\heavyrulewidth]
\bottomrule
\end{tabular}
}
\end{table}


\begin{figure}[p]
\centering

\begin{subfigure}{.30\linewidth}
    \includegraphics[width=\linewidth]{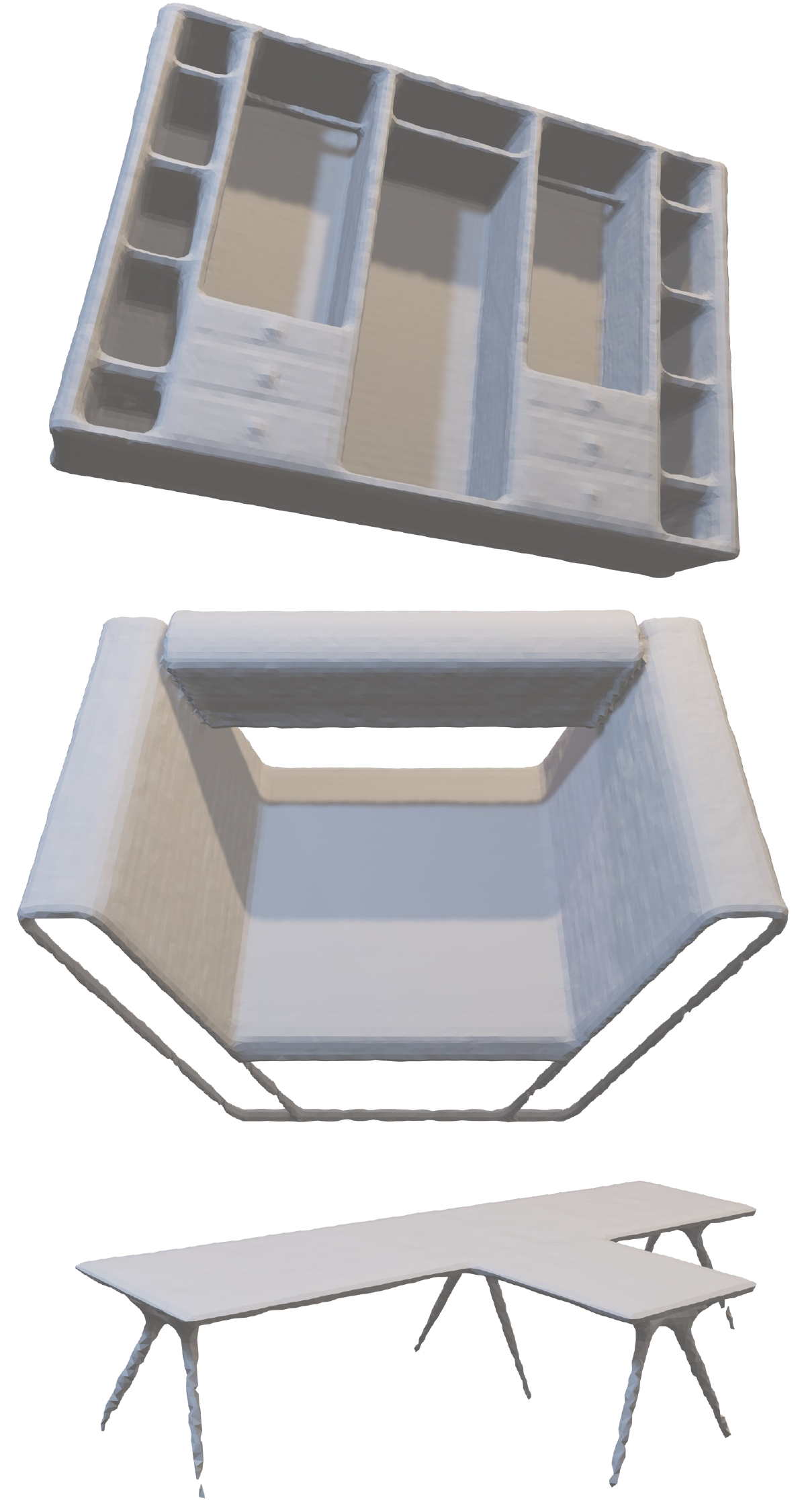}
    \caption{GT Mesh}
\end{subfigure}
\hspace{1mm}
\hfill
\begin{subfigure}{.30\linewidth}
    \includegraphics[width=\linewidth]{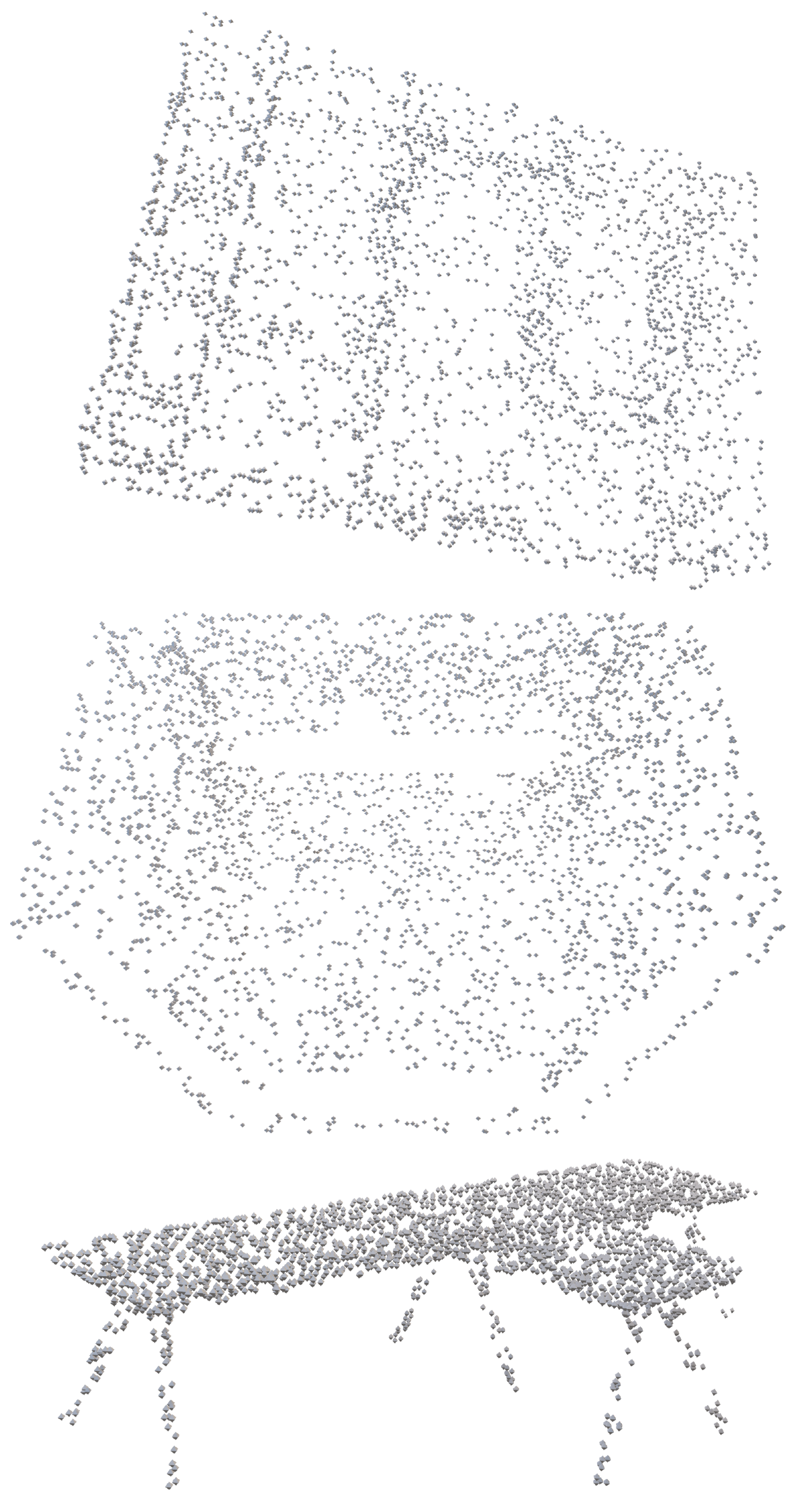}
    \caption{Input points}
\end{subfigure}
\hspace{1mm}
\hfill
\begin{subfigure}{.30\linewidth}
    \includegraphics[width=\linewidth]{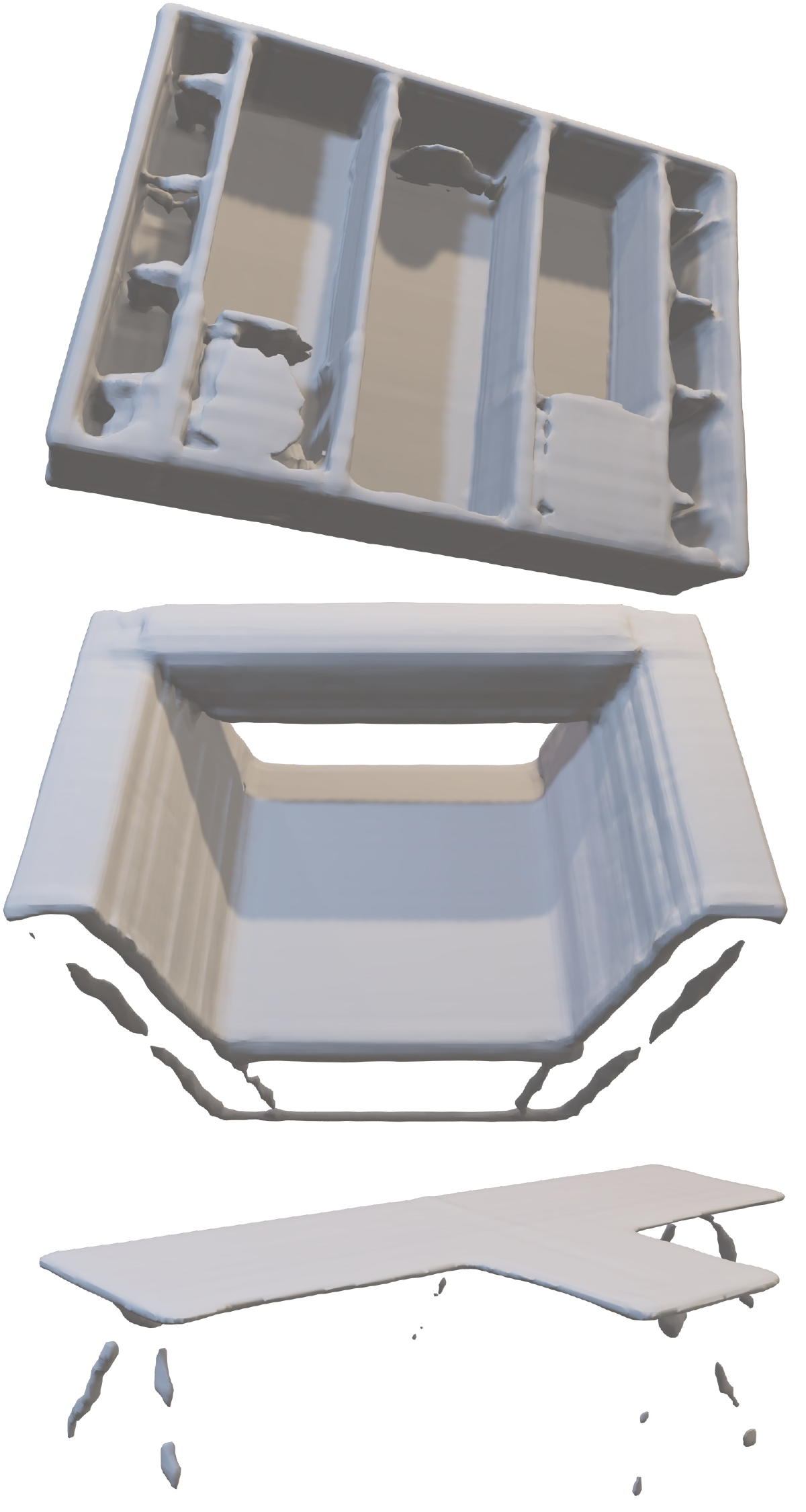}
    \caption{ConvONet~\cite{peng2020convolutional}}
\end{subfigure}
\hspace{1mm}
\hfill
\begin{subfigure}{.30\linewidth}
    \includegraphics[width=\linewidth]{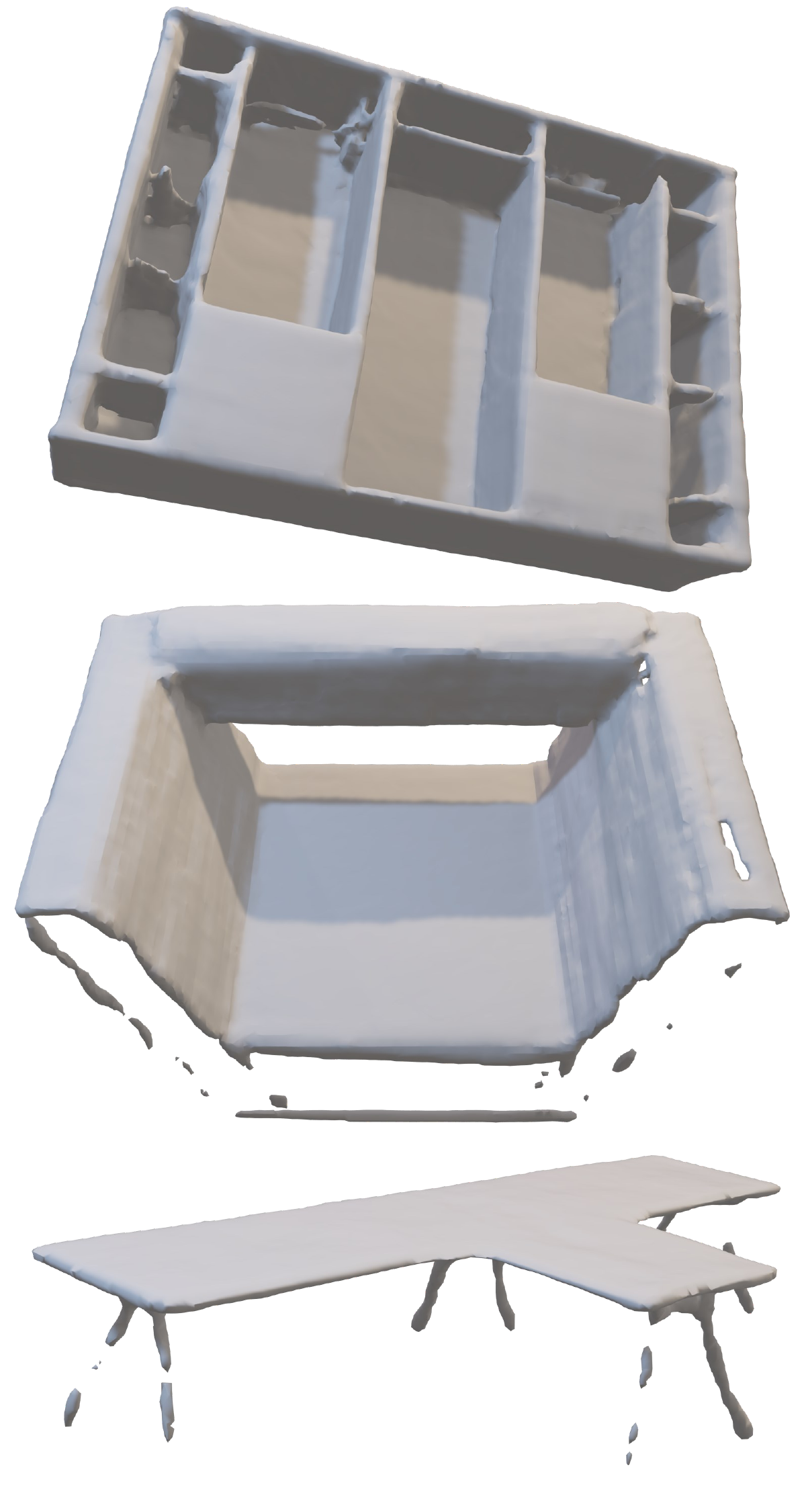}
    \caption{POCO~\cite{boulch2022poco}}
\end{subfigure}
\hspace{1mm}
\hfill
\begin{subfigure}{.30\linewidth}
    \includegraphics[width=\linewidth]{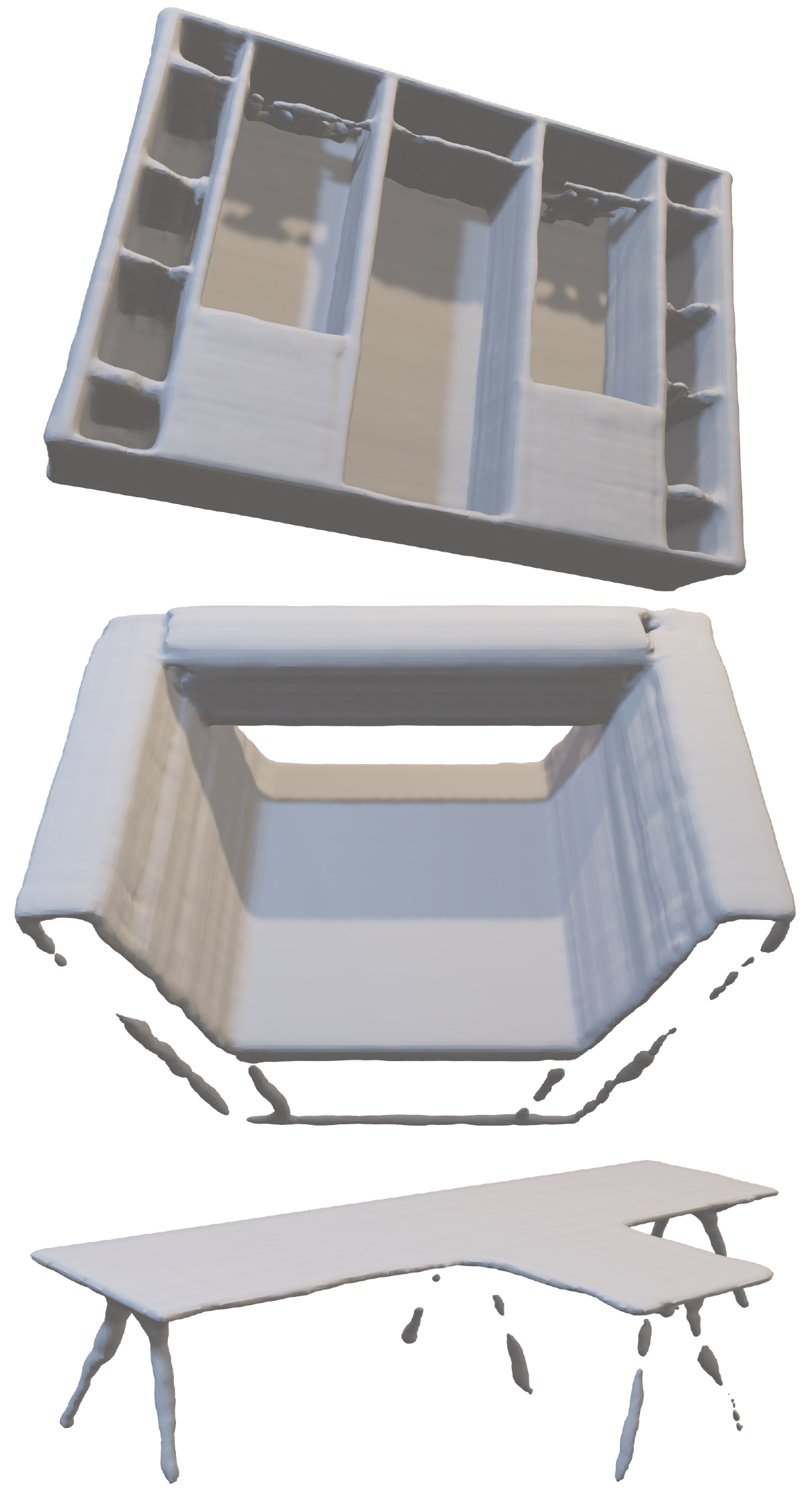}
    \caption{ALTO~\cite{wang2023alto}}
\end{subfigure}
\hspace{1mm}
\hfill
\begin{subfigure}{.30\linewidth}
    \includegraphics[width=\linewidth]{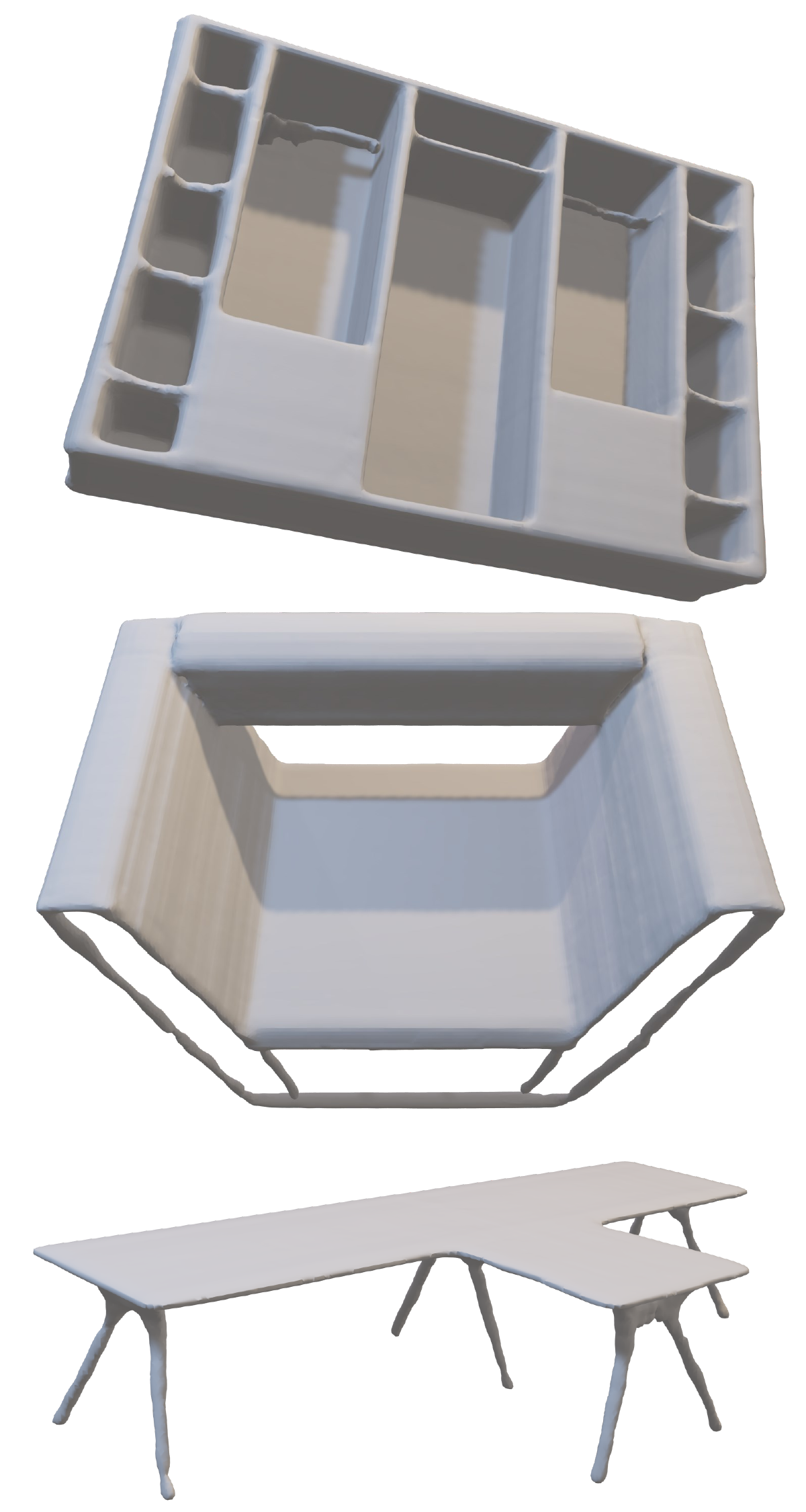}
    \caption{\texttt{DITTO} (ours)}
\end{subfigure}
\vspace{-2mm}
\caption{
    \textbf{Object-level 3D reconstruction comparison on ShapeNet~\cite{chang2015shapenet} with 3K input points and noise level 0.005}.
}
\vspace{-2mm}
\label{fig:supp:shapenet1}
\end{figure}

\begin{figure}[p]
\centering

\begin{subfigure}{.30\linewidth}
    \includegraphics[width=\linewidth]{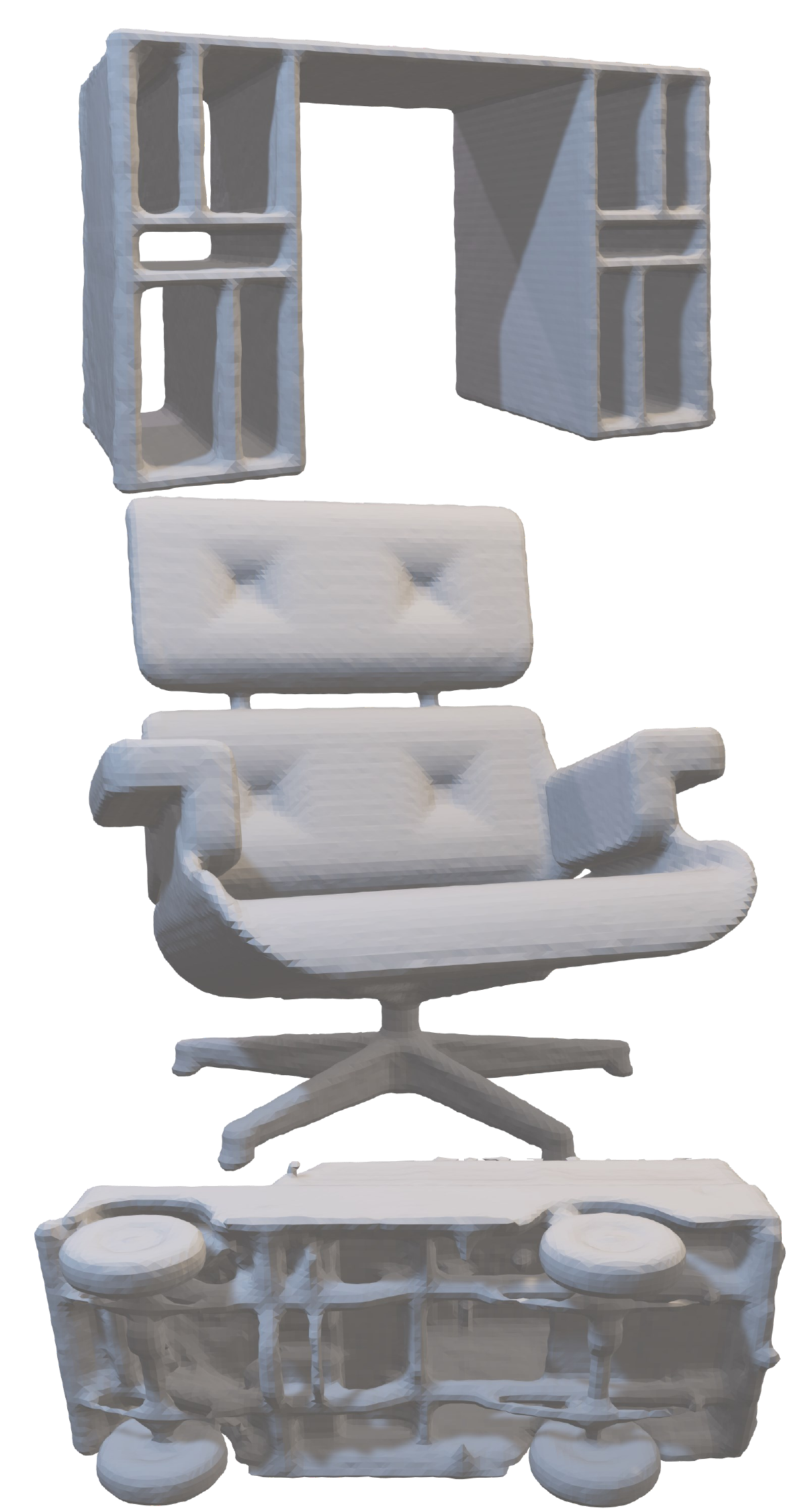}
    \caption{GT Mesh}
\end{subfigure}
\hspace{1mm}
\hfill
\begin{subfigure}{.30\linewidth}
    \includegraphics[width=\linewidth]{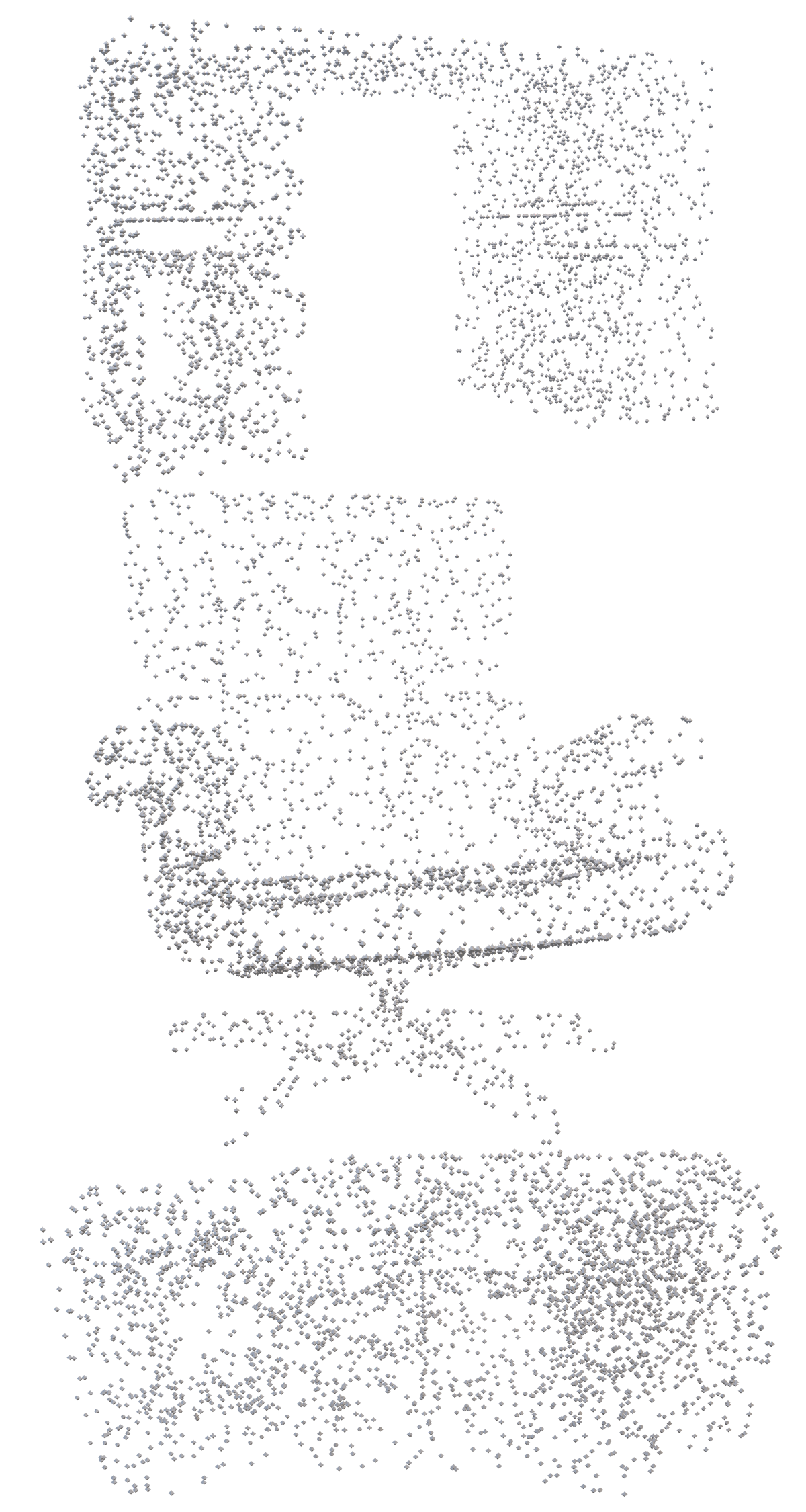}
    \caption{Input points}
\end{subfigure}
\hspace{1mm}
\hfill
\begin{subfigure}{.30\linewidth}
    \includegraphics[width=\linewidth]{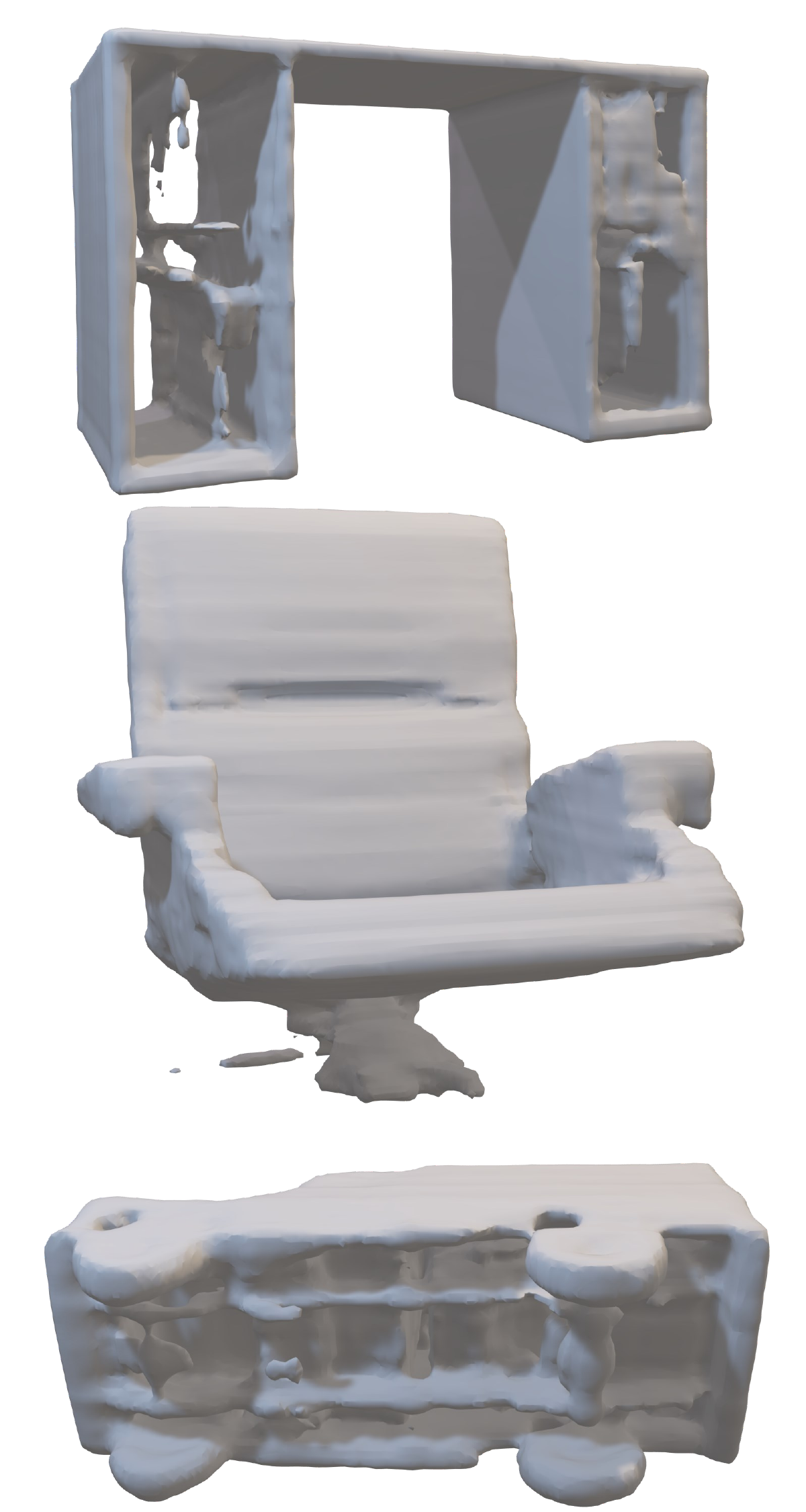}
    \caption{ConvONet~\cite{peng2020convolutional}}
\end{subfigure}
\hspace{1mm}
\hfill
\begin{subfigure}{.30\linewidth}
    \includegraphics[width=\linewidth]{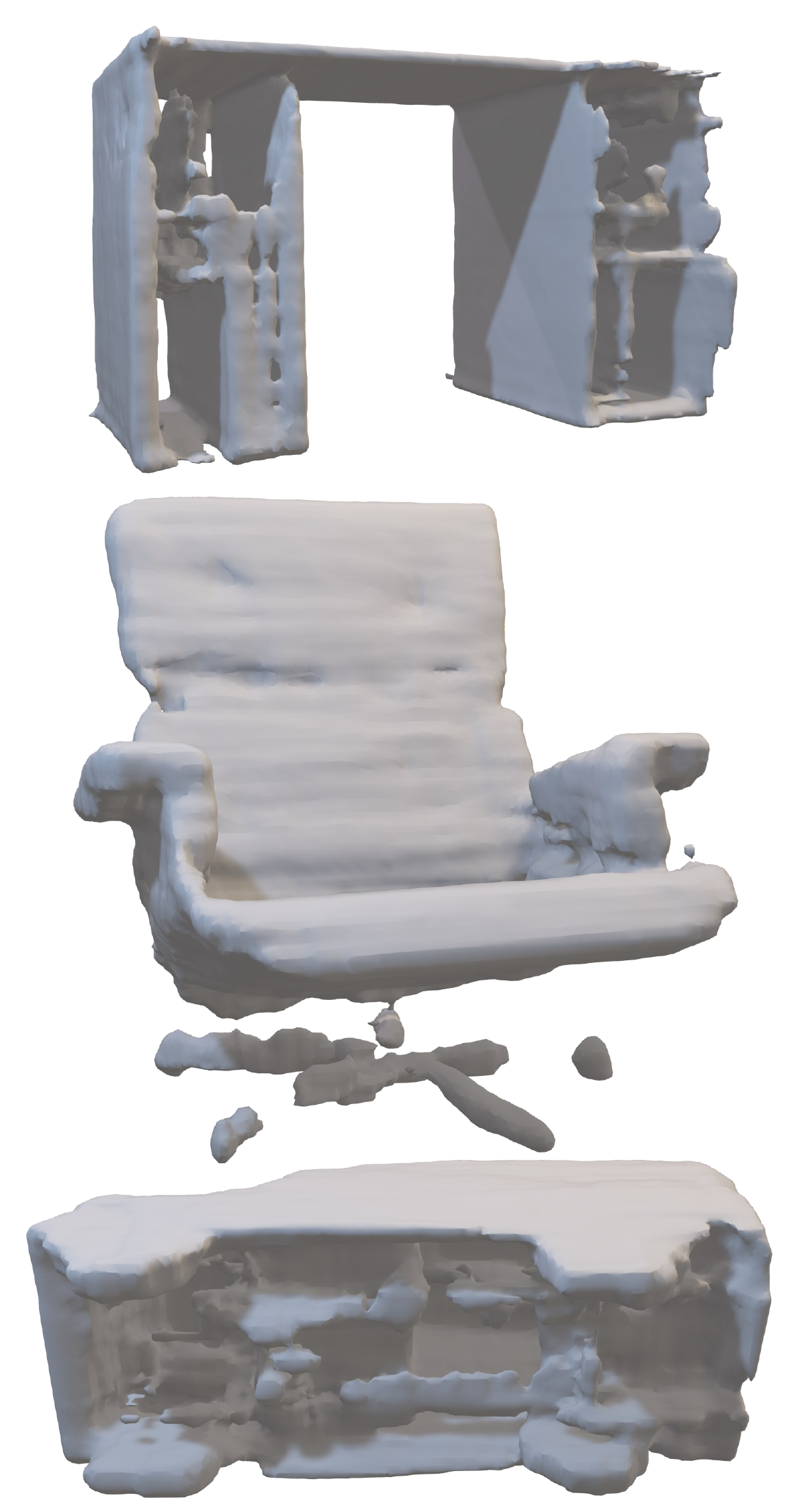}
    \caption{POCO~\cite{boulch2022poco}}
\end{subfigure}
\hspace{1mm}
\hfill
\begin{subfigure}{.30\linewidth}
    \includegraphics[width=\linewidth]{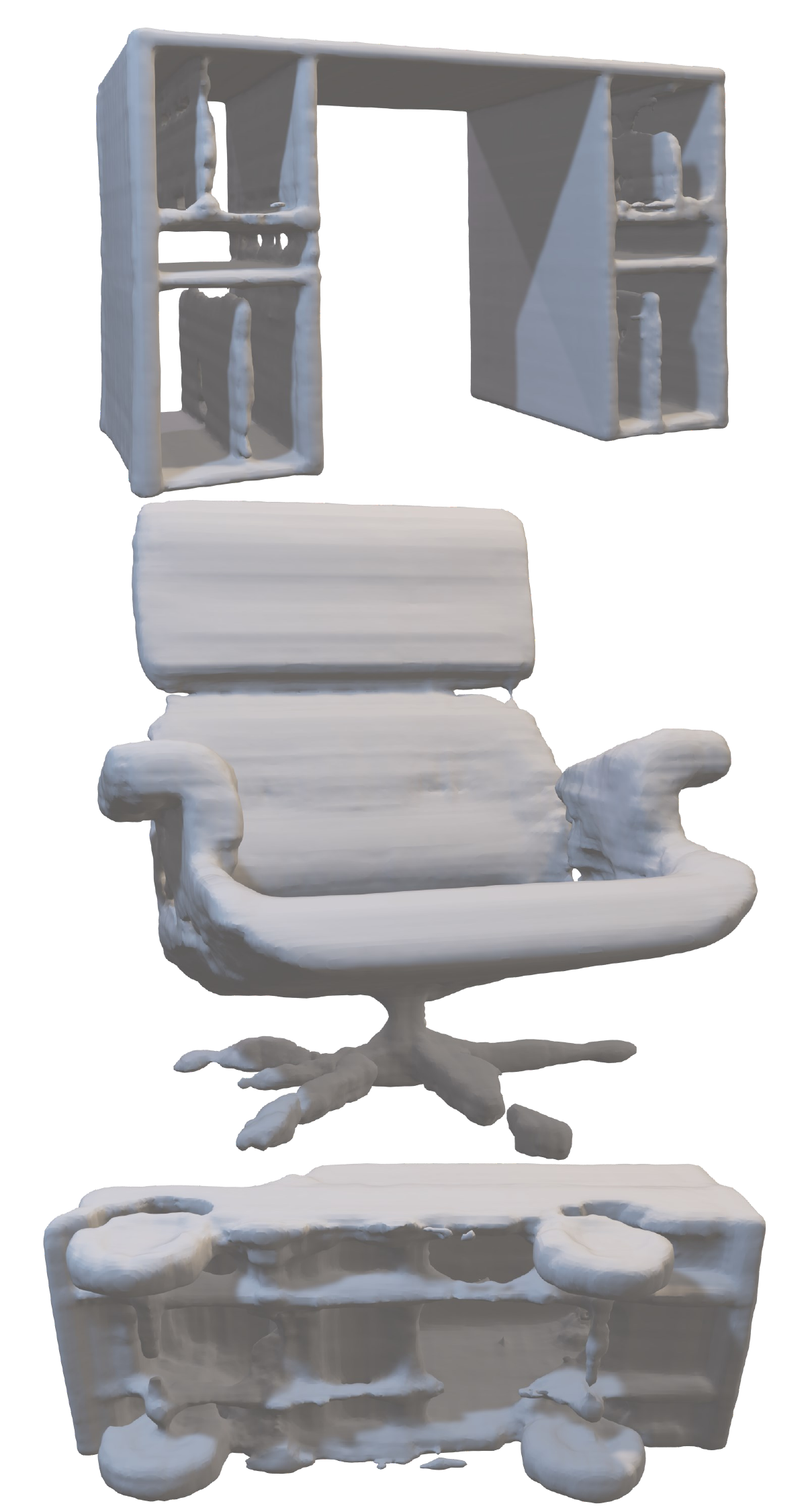}
    \caption{ALTO~\cite{wang2023alto}}
\end{subfigure}
\hspace{1mm}
\hfill
\begin{subfigure}{.30\linewidth}
    \includegraphics[width=\linewidth]{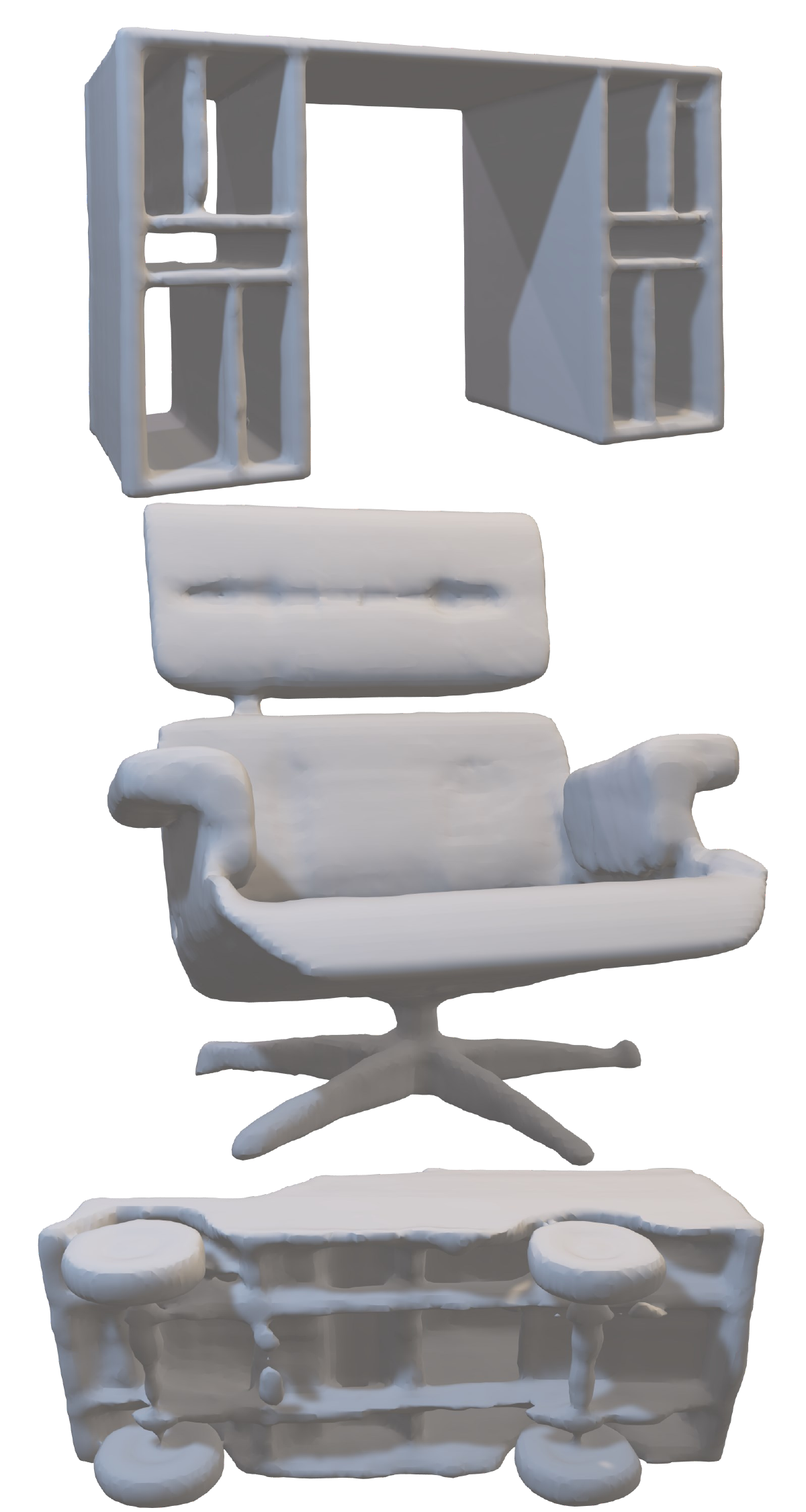}
    \caption{\texttt{DITTO} (ours)}
\end{subfigure}
\vspace{-2mm}
\caption{
    \textbf{Object-level 3D reconstruction comparison on ShapeNet~\cite{chang2015shapenet} with 1K input points and noise level 0.005}.
}
\vspace{-2mm}
\label{fig:supp:shapenet2}
\end{figure}

\begin{figure}[p]
\centering

\begin{subfigure}{.30\linewidth}
    \includegraphics[width=\linewidth]{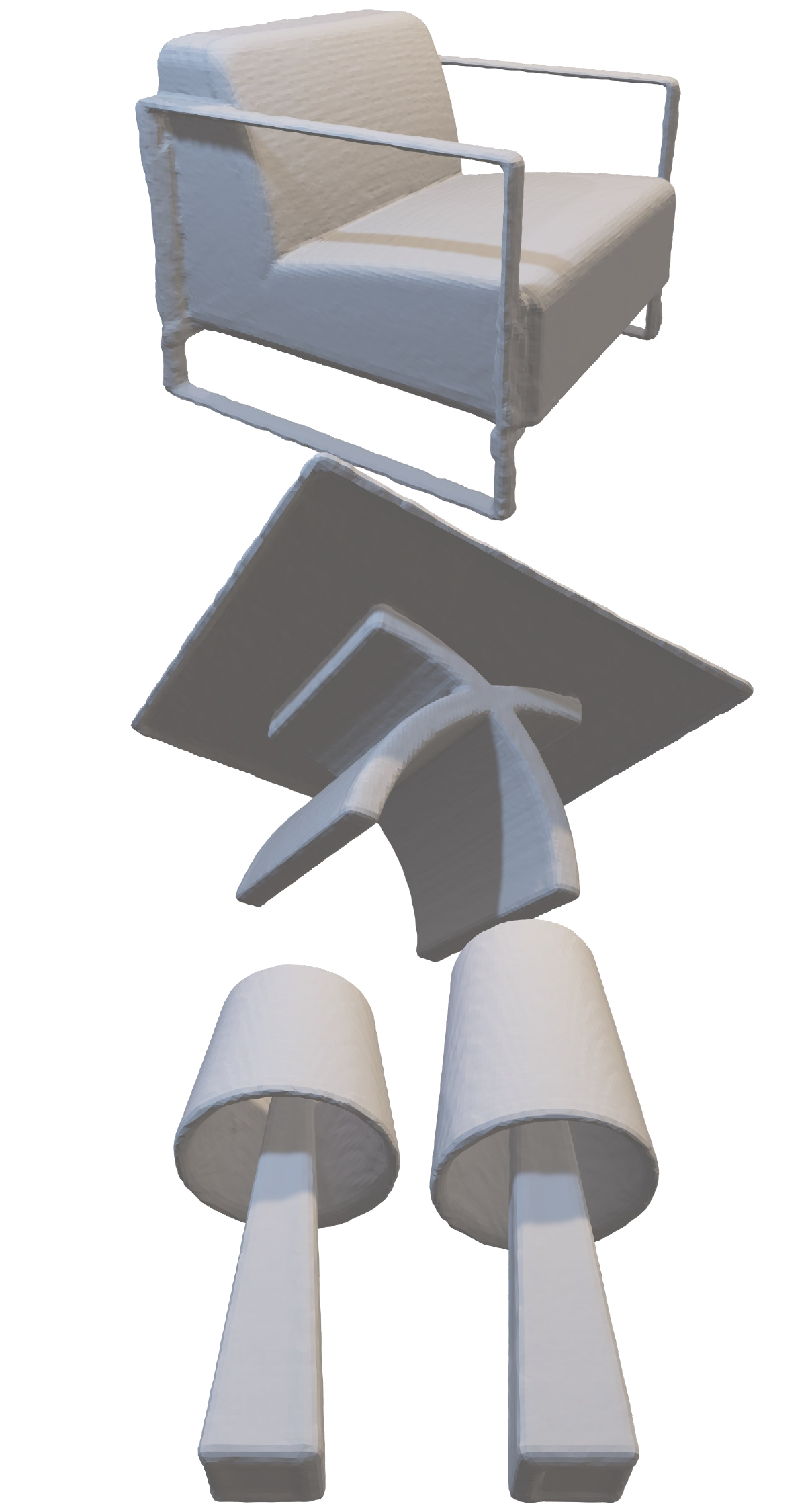}
    \caption{GT Mesh}
\end{subfigure}
\hspace{1mm}
\hfill
\begin{subfigure}{.30\linewidth}
    \includegraphics[width=\linewidth]{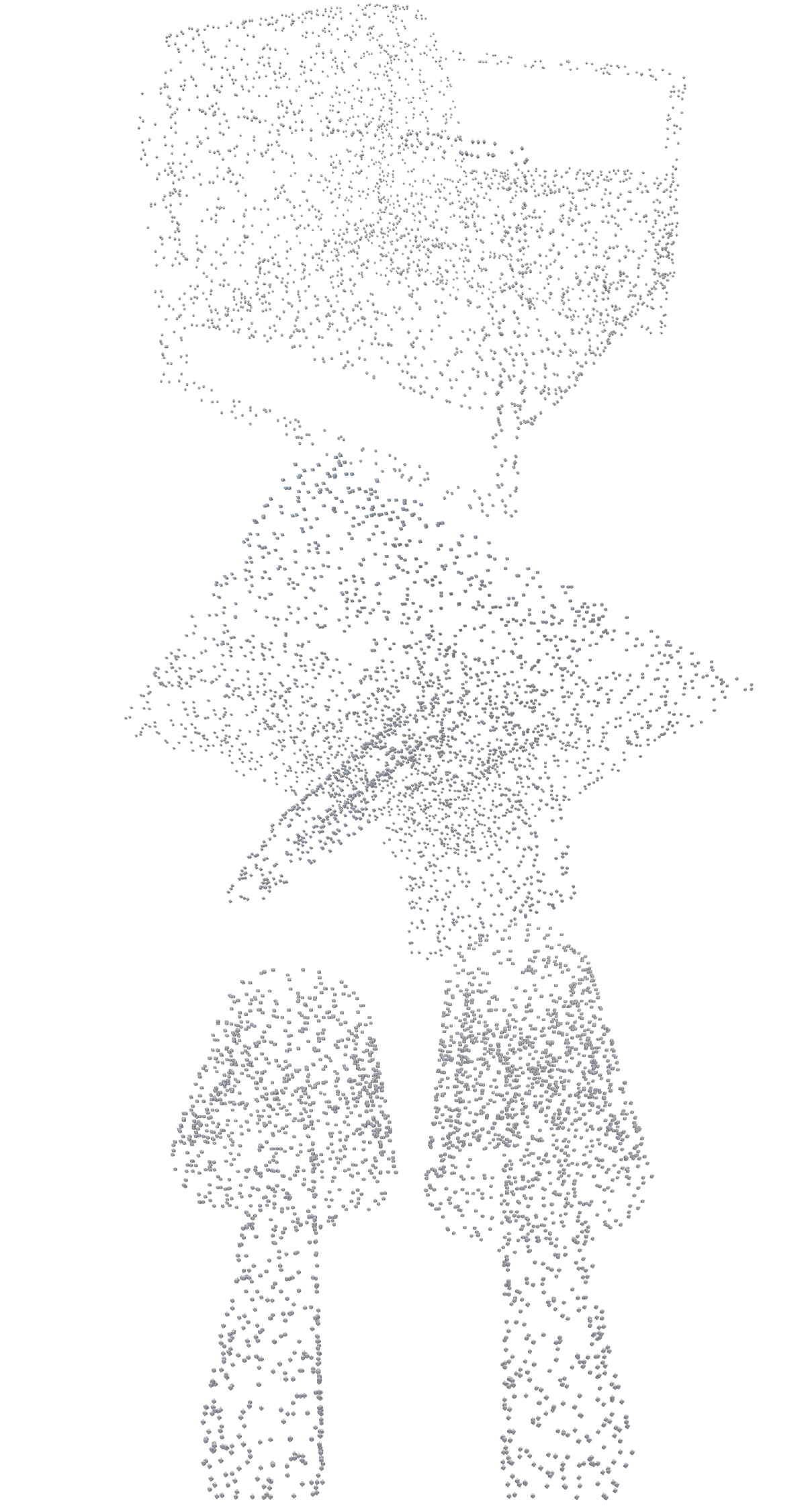}
    \caption{Input points}
\end{subfigure}
\hspace{1mm}
\hfill
\begin{subfigure}{.30\linewidth}
    \includegraphics[width=\linewidth]{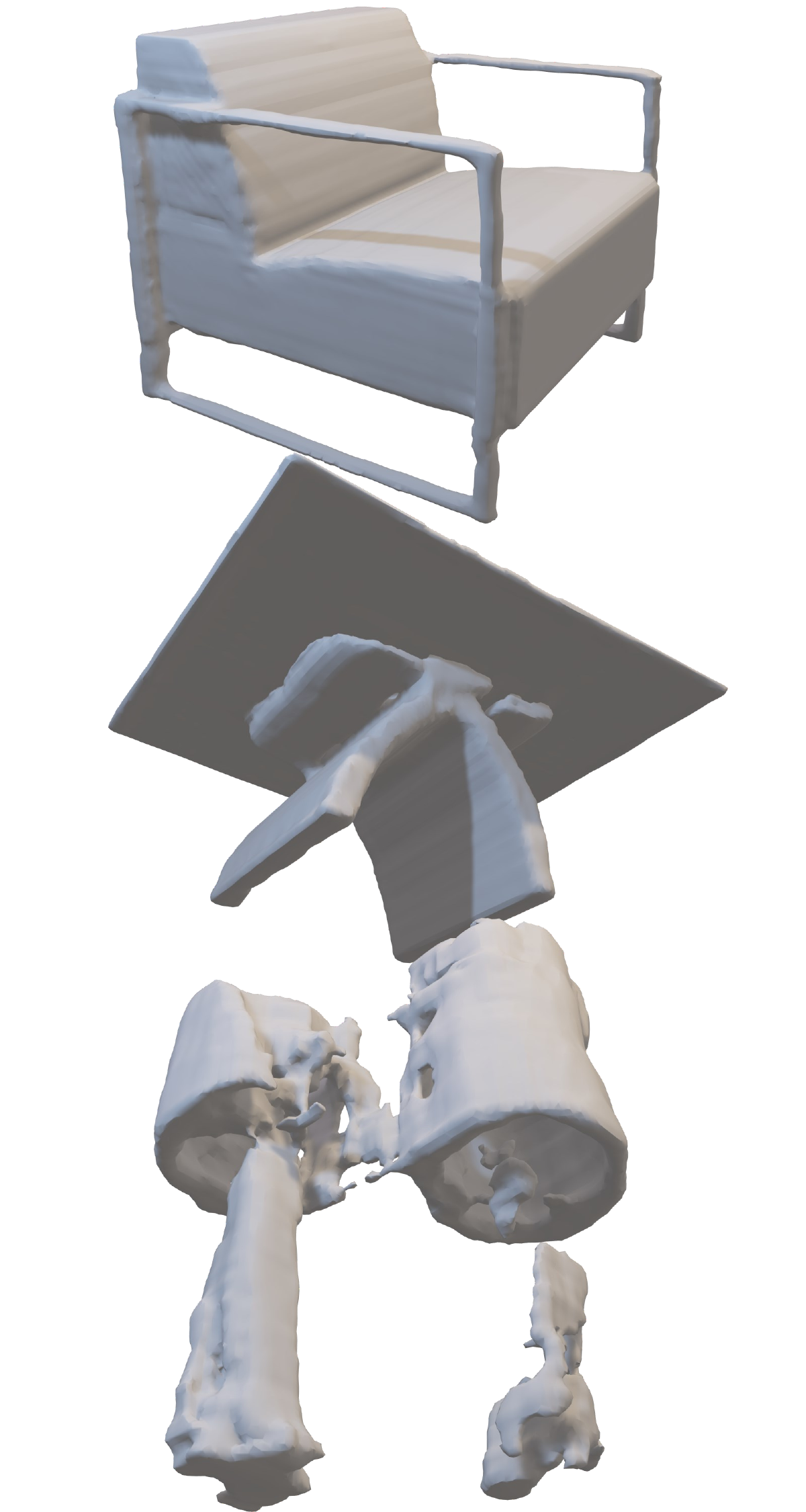}
    \caption{ConvONet~\cite{peng2020convolutional}}
\end{subfigure}
\hspace{1mm}
\hfill
\begin{subfigure}{.30\linewidth}
    \includegraphics[width=\linewidth]{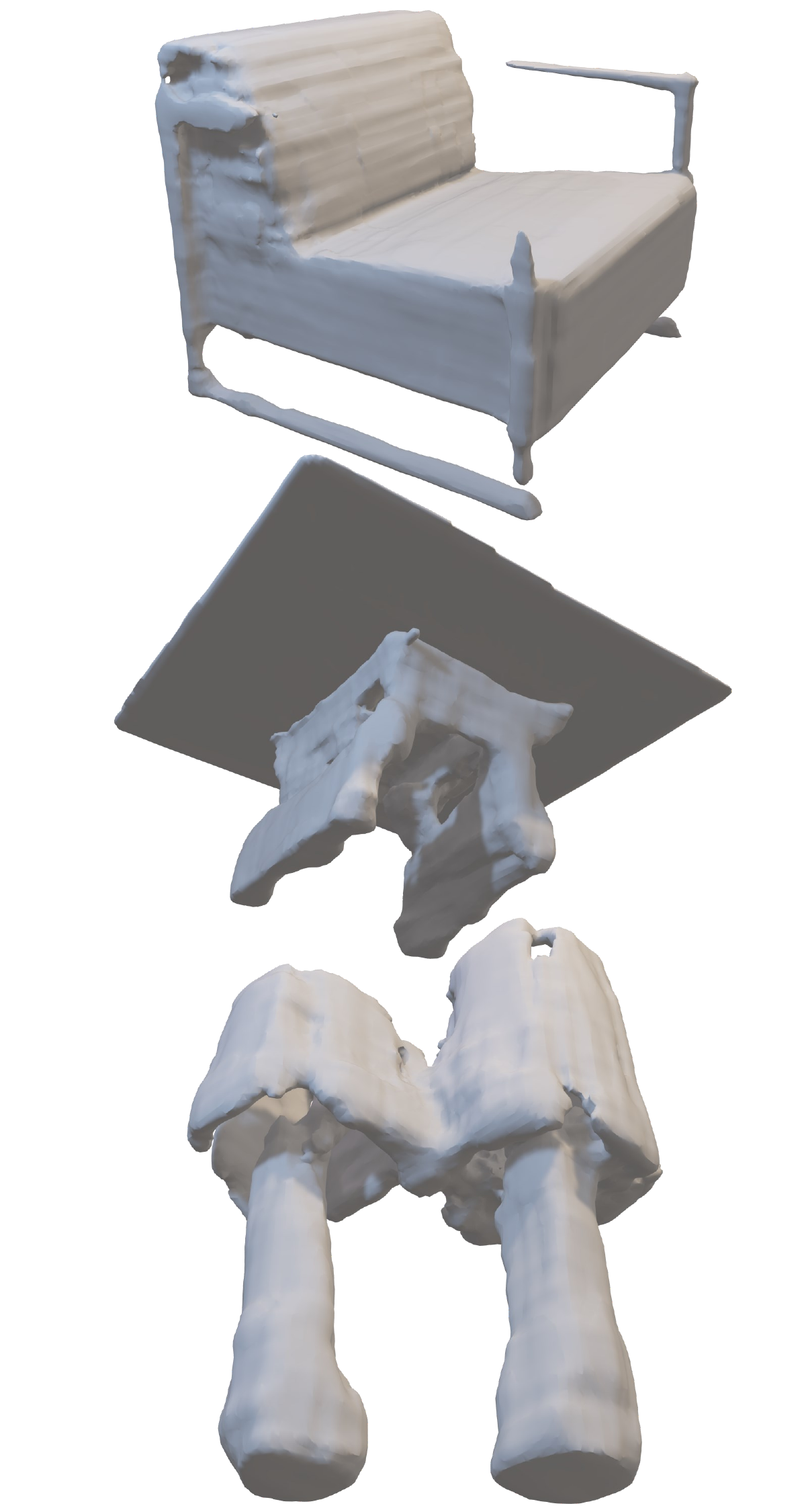}
    \caption{POCO~\cite{boulch2022poco}}
\end{subfigure}
\hspace{1mm}
\hfill
\begin{subfigure}{.30\linewidth}
    \includegraphics[width=\linewidth]{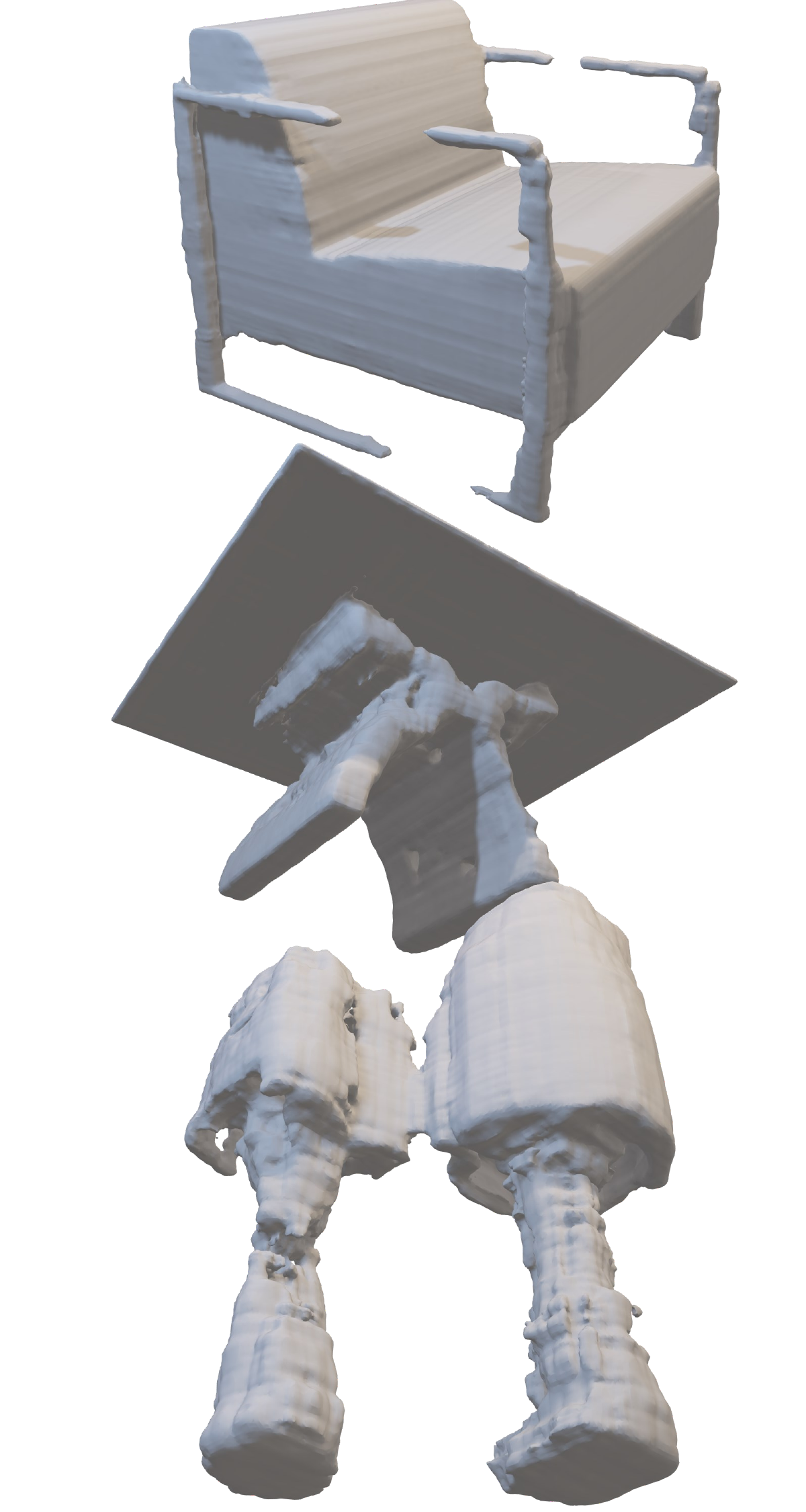}
    \caption{ALTO~\cite{wang2023alto}}
\end{subfigure}
\hspace{1mm}
\hfill
\begin{subfigure}{.30\linewidth}
    \includegraphics[width=\linewidth]{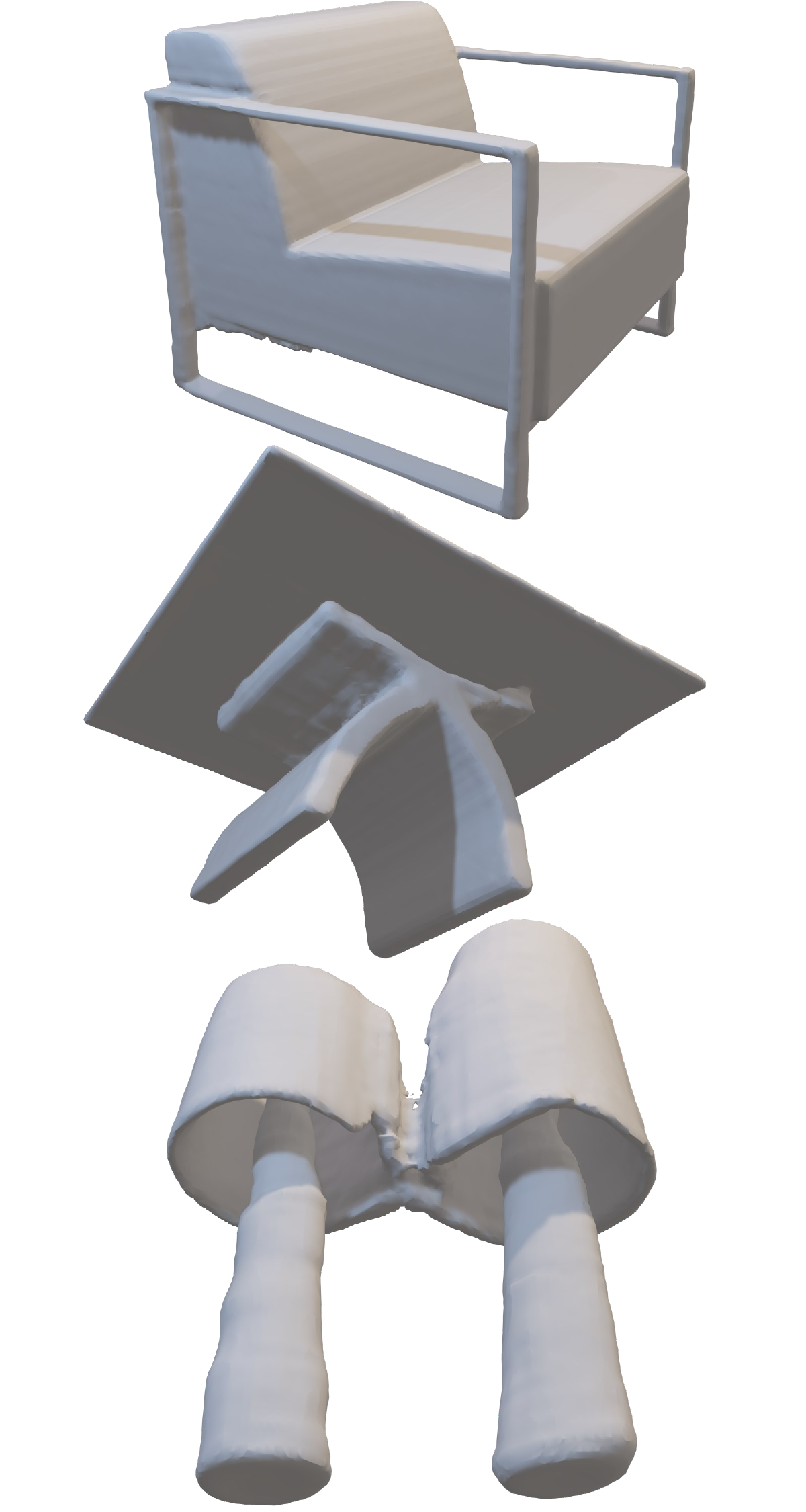}
    \caption{\texttt{DITTO} (ours)}
\end{subfigure}
\vspace{-2mm}
\caption{
    \textbf{Object-level 3D reconstruction comparison on ShapeNet~\cite{chang2015shapenet} with 0.3K input points and noise level 0.005}.
}
\vspace{-2mm}
\label{fig:supp:shapenet3}
\end{figure}

\begin{figure}[p]
\centering

\begin{subfigure}{\linewidth}
    \includegraphics[width=\linewidth]{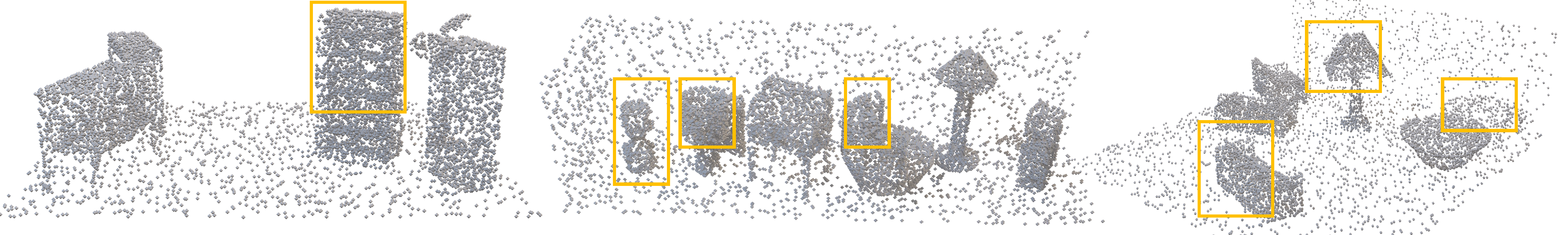}
    \caption{Input points}
\end{subfigure}
\begin{subfigure}{\linewidth}
    \includegraphics[width=\linewidth]{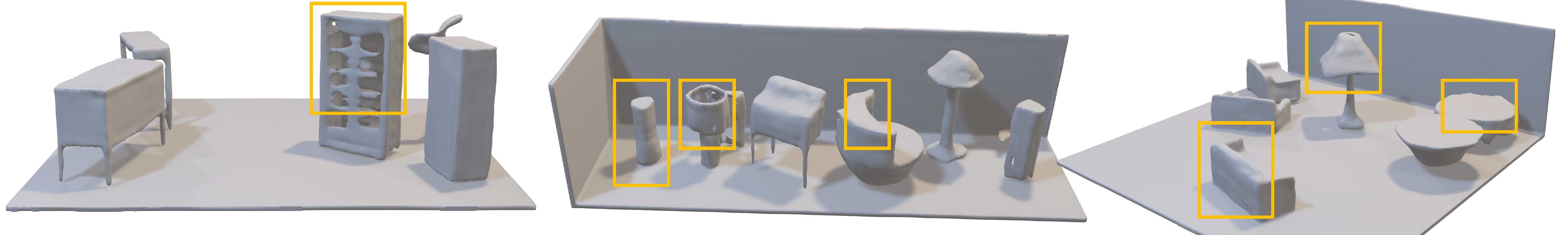}
    \caption{ConvONet~\cite{peng2020convolutional}}
\end{subfigure}
\begin{subfigure}{\linewidth}
    \includegraphics[width=\linewidth]{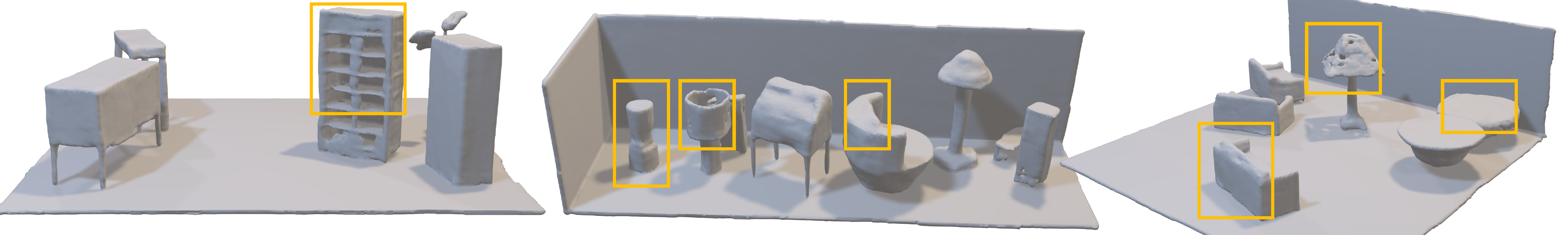}
    \caption{POCO~\cite{boulch2022poco}}
\end{subfigure}
\begin{subfigure}{\linewidth}
    \includegraphics[width=\linewidth]{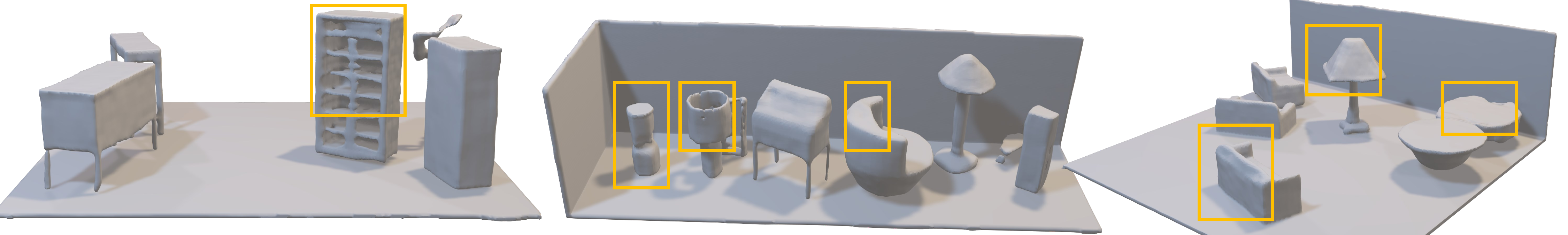}
    \caption{ALTO~\cite{wang2023alto}}
\end{subfigure}
\begin{subfigure}{\linewidth}
    \includegraphics[width=\linewidth]{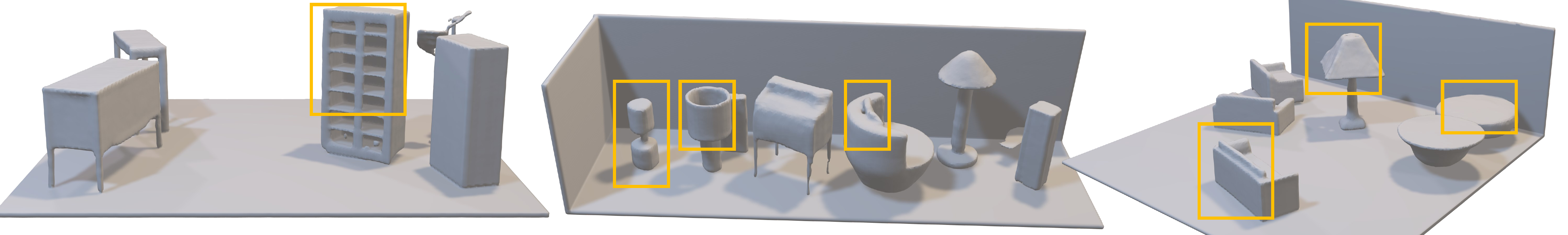}
    \caption{\texttt{DITTO} (ours)}
\end{subfigure}
\vspace{-2mm}
\caption{
    \textbf{Scene-level 3D reconstruction results on the Synthetic Rooms dataset~\cite{peng2020convolutional} with 10K input points and noise level 0.005}.
}
\vspace{-2mm}
\label{fig:supp:synthetic_rooms}
\end{figure}

\begin{figure}[p]
\centering

\begin{subfigure}{.85\linewidth}
    \includegraphics[width=\linewidth]{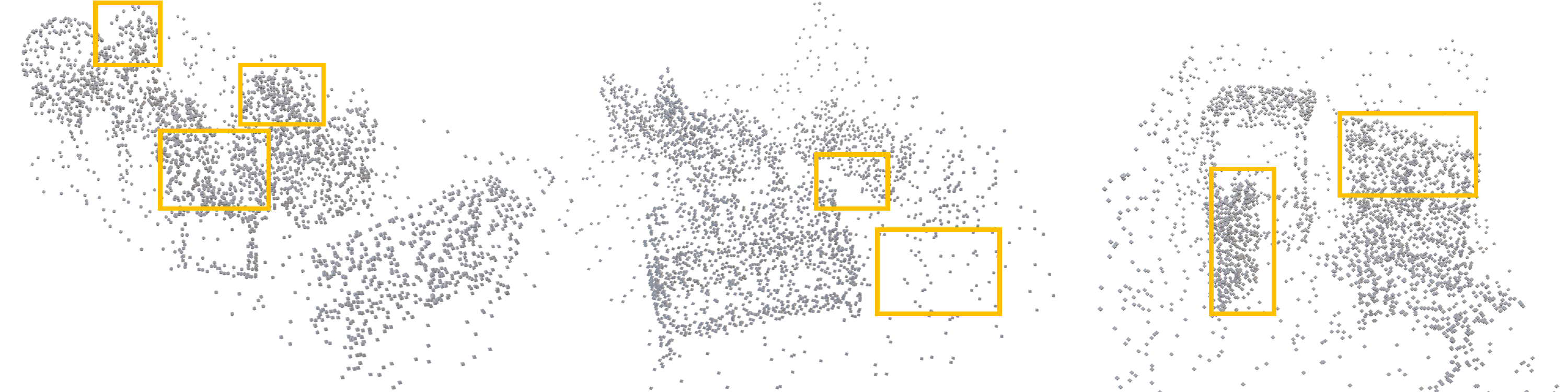}
    \caption{Input points}
\end{subfigure}
\hfill
\begin{subfigure}{.85\linewidth}
    \includegraphics[width=\linewidth]{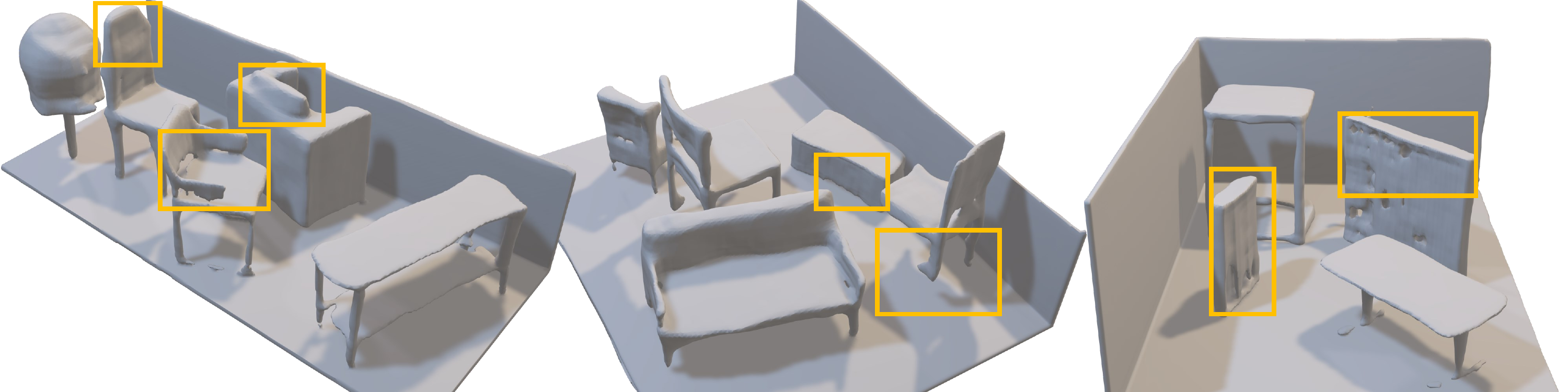}
    \caption{ConvONet~\cite{peng2020convolutional}}
\end{subfigure}
\hfill
\begin{subfigure}{.85\linewidth}
    \includegraphics[width=\linewidth]{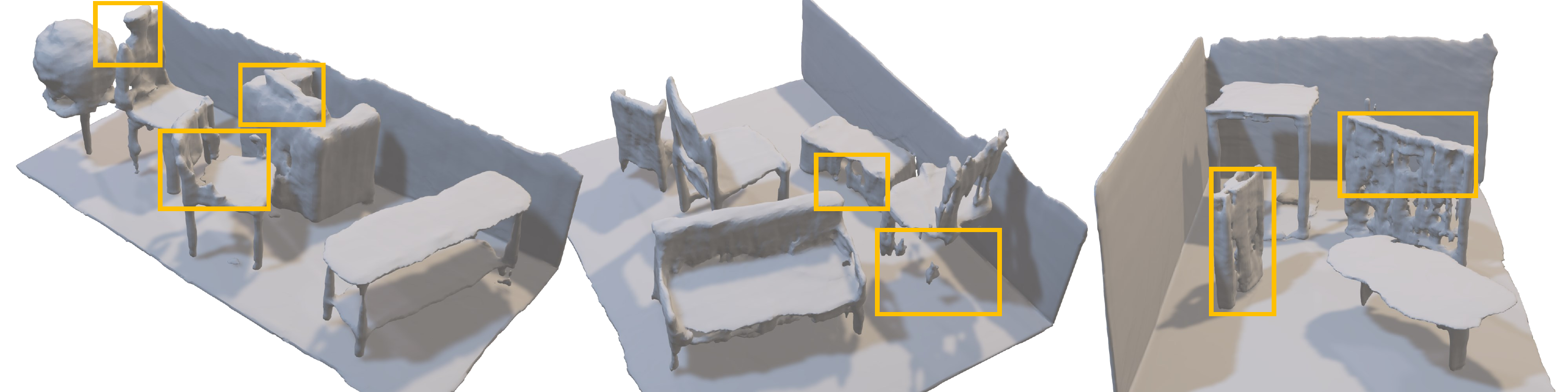}
    \caption{POCO~\cite{boulch2022poco}}
\end{subfigure}
\hfill
\begin{subfigure}{.85\linewidth}
    \includegraphics[width=\linewidth]{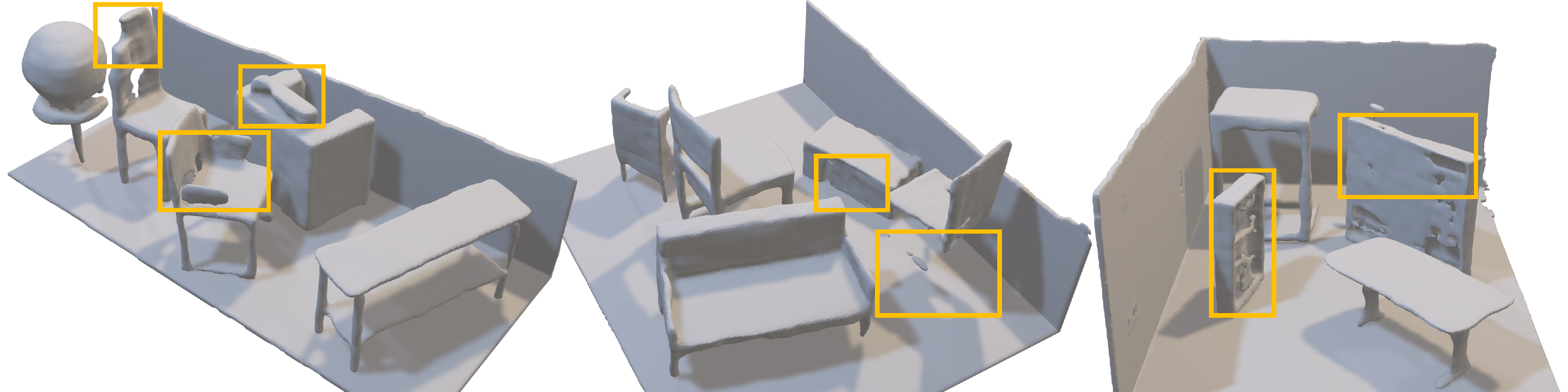}
    \caption{ALTO~\cite{wang2023alto}}
\end{subfigure}
\hfill
\begin{subfigure}{.85\linewidth}
    \includegraphics[width=\linewidth]{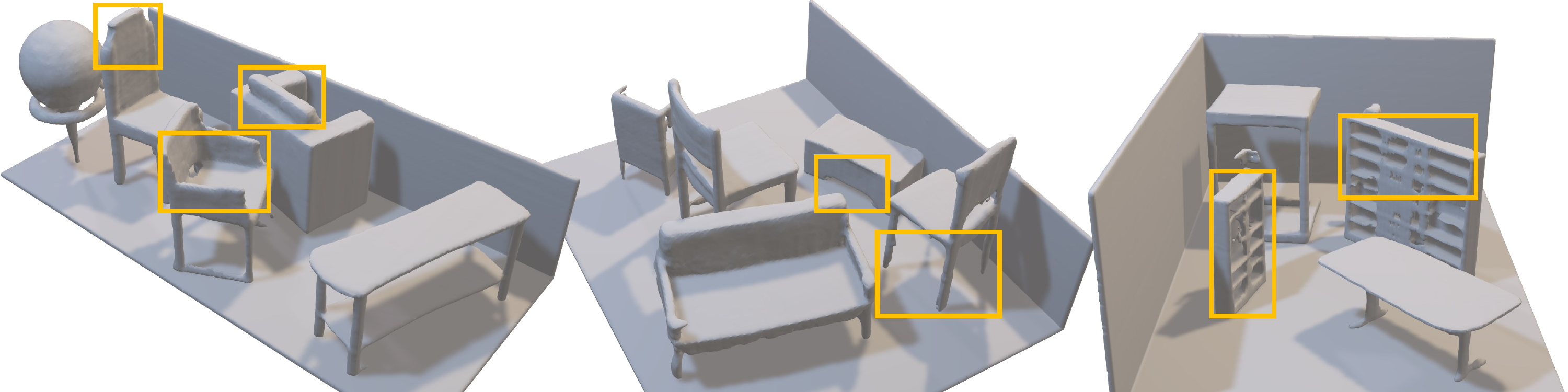}
    \caption{\texttt{DITTO} (ours)}
\end{subfigure}
\vspace{-2mm}
\caption{
    \textbf{Scene-level 3D reconstruction results on the Synthetic Rooms dataset~\cite{peng2020convolutional} with 3K input points and noise level 0.005}.
}
\vspace{-2mm}
\label{fig:supp:synthetic_rooms_sparse}
\end{figure}

\begin{figure}[p]
\centering

\begin{subfigure}{.9\linewidth}
    \includegraphics[width=\linewidth]{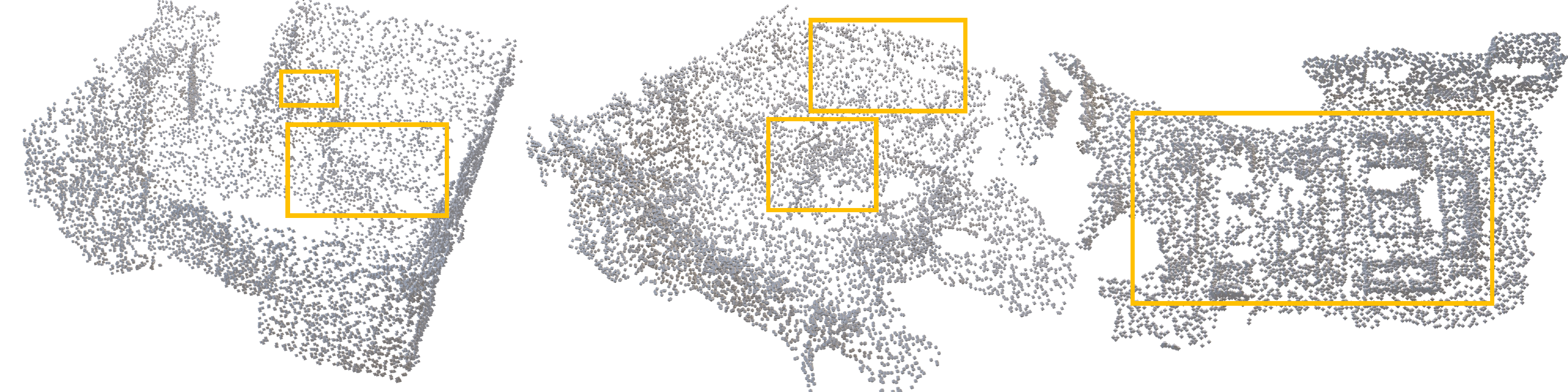}
    \caption{Input points}
\end{subfigure}
\hfill
\begin{subfigure}{.9\linewidth}
    \includegraphics[width=\linewidth]{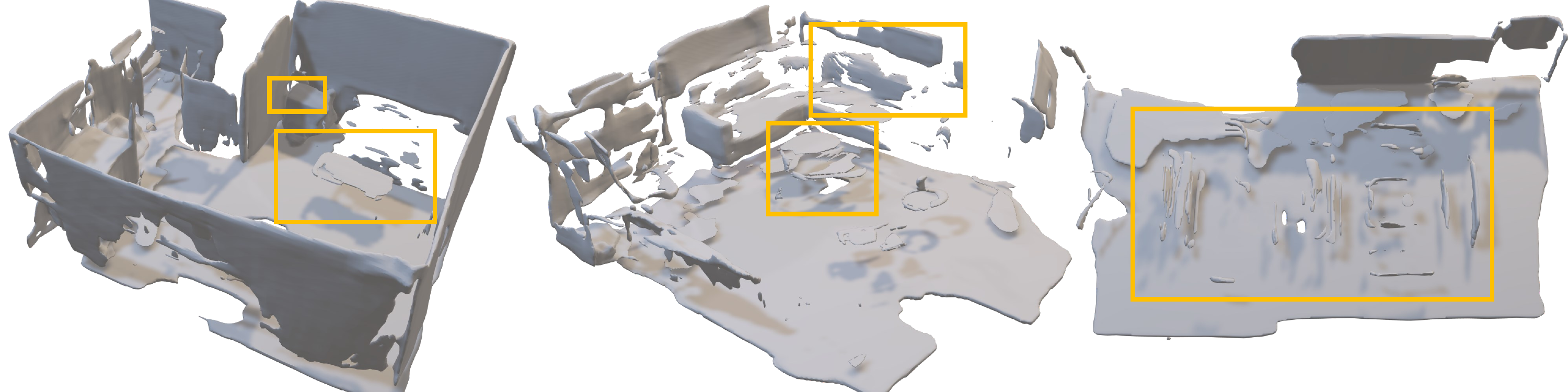}
    \caption{ConvONet~\cite{peng2020convolutional}}
\end{subfigure}
\hfill
\begin{subfigure}{.9\linewidth}
    \includegraphics[width=\linewidth]{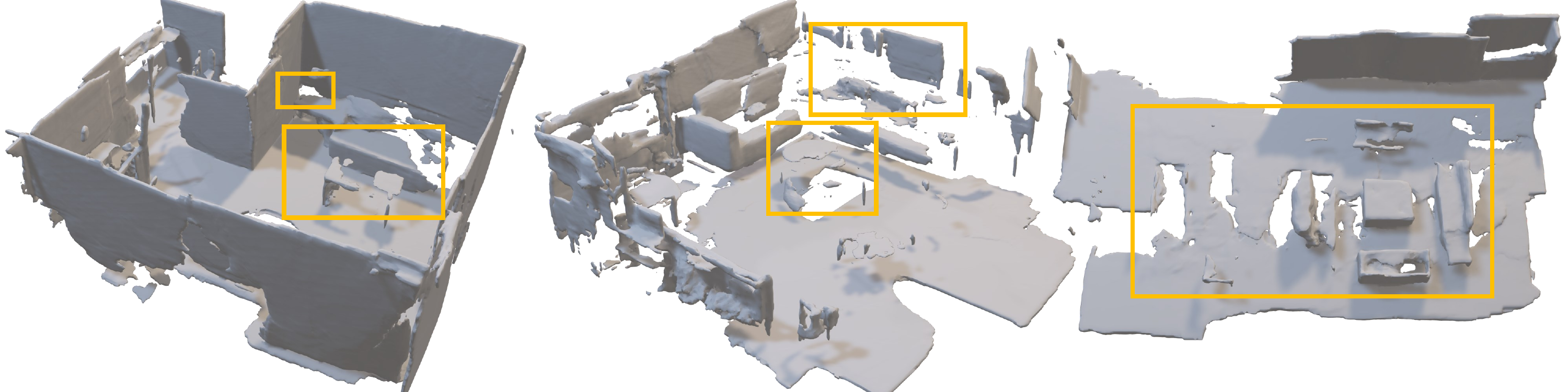}
    \caption{POCO~\cite{boulch2022poco}}
\end{subfigure}
\hfill
\begin{subfigure}{.9\linewidth}
    \includegraphics[width=\linewidth]{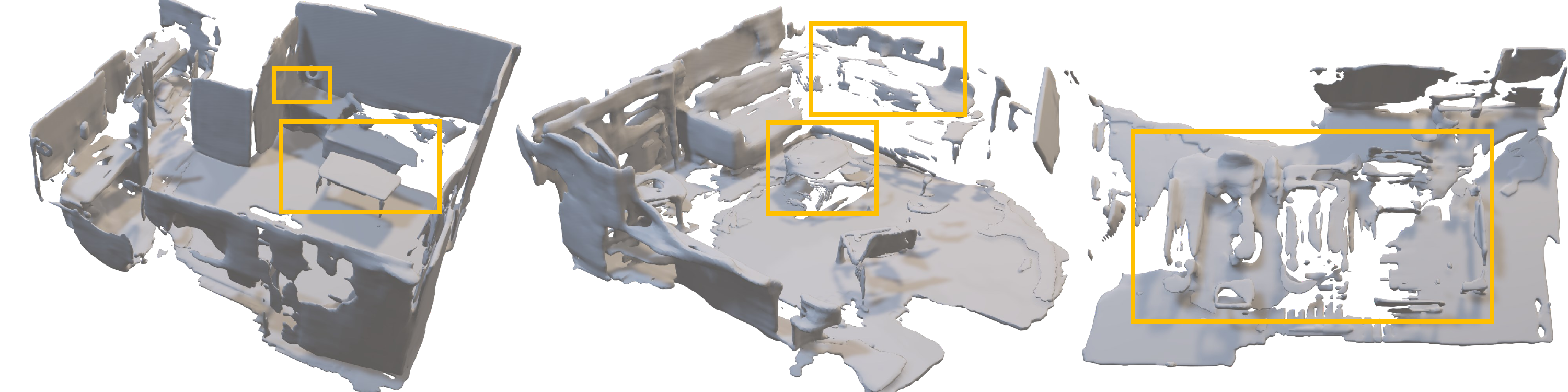}
    \caption{ALTO~\cite{wang2023alto}}
\end{subfigure}
\hfill
\begin{subfigure}{.9\linewidth}
    \includegraphics[width=\linewidth]{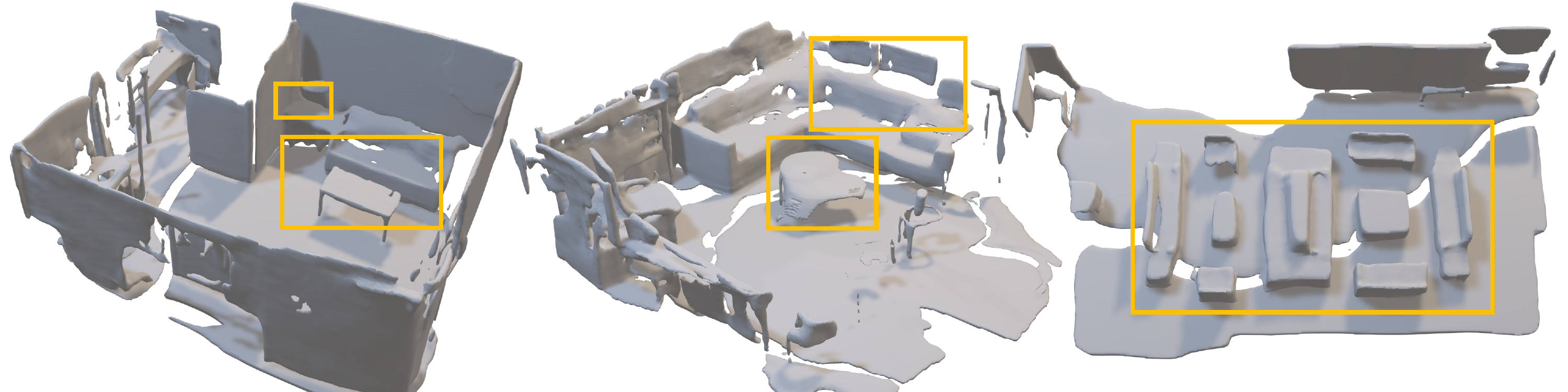}
    \caption{\texttt{DITTO} (ours)}
\end{subfigure}
\vspace{-2mm}
\caption{
    \textbf{Scene-level 3D reconstruction results on the ScanNet-v2 dataset~\cite{dai2017scannet}}.
}
\vspace{-2mm}
\label{fig:supp:scannet}
\end{figure}

}{}

\end{document}